\documentclass[letterpaper,journal]{IEEEtran}
\usepackage{amsmath}
\usepackage{amssymb}
\usepackage{mathtools}
\usepackage{amsthm}
\usepackage{algorithmic}
\usepackage{algorithm}
\usepackage{array}
\usepackage[caption=false,font=footnotesize]{subfig}
\usepackage{textcomp}
\usepackage{stfloats}
\usepackage[table]{xcolor}
\usepackage{booktabs, colortbl}
\usepackage{url}
\usepackage{verbatim}
\usepackage{graphicx}
\usepackage{cite}
\usepackage{multirow}
\usepackage{ragged2e}
\definecolor{myblue}{RGB}{0,220,0}
\usepackage[colorlinks=true,linkcolor=myblue,citecolor=myblue,urlcolor=myblue]{hyperref}
\newtheorem{theorem}{Theorem}
\newtheorem{proposition}[theorem]{Proposition}
\newtheorem{lemma}[theorem]{Lemma}
\newtheorem{corollary}[theorem]{Corollary}
\theoremstyle{definition}

\newtheorem{assumption}[theorem]{Assumption}
\theoremstyle{remark}

\allowdisplaybreaks[4]

\def \E{\mathbb{E}}

\usepackage{pifont}
\usepackage{bm}
\usepackage{tabularx}
\usepackage{placeins}

\definecolor{Gray}{gray}{0.9}
\definecolor{average}{RGB}{240, 204, 126}
\definecolor{last}{RGB}{225, 191, 192}
\definecolor{transfer}{RGB}{191, 191, 225}
\definecolor{diag}{RGB}{151, 205, 112}

\begin{document}

\title{KeepLoRA++: Continual Learning with Layer-Scaled Residual Gradient Adaptation}

\author{Mao-Lin Luo,
        Yi-Lin Zhang,
        Zi-Hao Zhou,
        Yankun Hong,
        Xialiang Tong,
        Mingxuan Yuan, \\
        Tong Wei,
        Min-Ling Zhang,~\IEEEmembership{Fellow,~IEEE
        }

\thanks{%
Mao-Lin Luo, Yi-Lin Zhang, Zi-Hao Zhou, Tong Wei and Min-Ling Zhang are with the School of Computer Science and Engineering, Southeast University, Nanjing 210096, China, and the Key Laboratory of Computer Network and Information Integration, Southeast University, Ministry of Education, China.
Yankun Hong, Xialiang Tong and Mingxuan Yuan are with Huawei Noah’s Ark Lab.

E-mail: \{ml\_luo, zhouzih, weit, 	zhangml\}@seu.edu.cn, \{hongyankun, tongxialiang, yuan.mingxuan\}@huawei.com
}%
}

\markboth{Priprint}%
{Luo \MakeLowercase{\textit{et al.}}: KeepLoRA++: Continual Learning with Layer-Scaled Residual Gradient Adaptation}

\maketitle

\begin{abstract}
\justifying\let\raggedright\justifying
Continual learning for pre-trained vision-language models requires balancing three competing objectives: retaining pre-trained knowledge, preserving knowledge from a sequence of learned tasks, and maintaining the plasticity to acquire new knowledge. %
In this paper, we study continual learning across Transformer layers and within each layer’s parameter space, and propose an effective approach called KeepLoRA++. The inter-layer perspective examines how knowledge retention is distributed across layers, while the intra-layer perspective focuses on the parameter space within each layer. Our analysis reveals a structural property: general transferable knowledge is mainly encoded in the shallow layers and the principal subspace of the parameters, while task-specific adaptations are localized in the deep layers and the residual subspace. Motivated by this insight, KeepLoRA++ introduces a layer-scaled residual gradient adaptation method. 
Specifically, at the inter-layer level, we adopt a shallow-to-deep scaling strategy that assigns smaller update magnitudes to shallow layers and larger ones to deeper layers. At the intra-layer level, we constrain parameter updates to a gradient-informed residual subspace and simplify LoRA initialization without modifying the frozen backbone weights. We further derive an optimal allocation of update magnitudes across layers based on retention curvature and prove that the simplified initialization preserves the optimization trajectory. 
Extensive experiments demonstrate that KeepLoRA++ effectively balances these three objectives, achieving strong performance across image classification, visual question answering, and video understanding tasks. The implementation code is available at \href{https://github.com/MaolinLuo/KeepLoRA}{https://github.com/MaolinLuo/KeepLoRA}.

\end{abstract}

\begin{IEEEkeywords}
continual learning, vision-language models.
\end{IEEEkeywords}

\IEEEpeerreviewmaketitle

\section{Introduction}

\begin{figure}[!t]
\centering
\subfloat[General-domain tasks]{
    \includegraphics[width=0.23\textwidth]{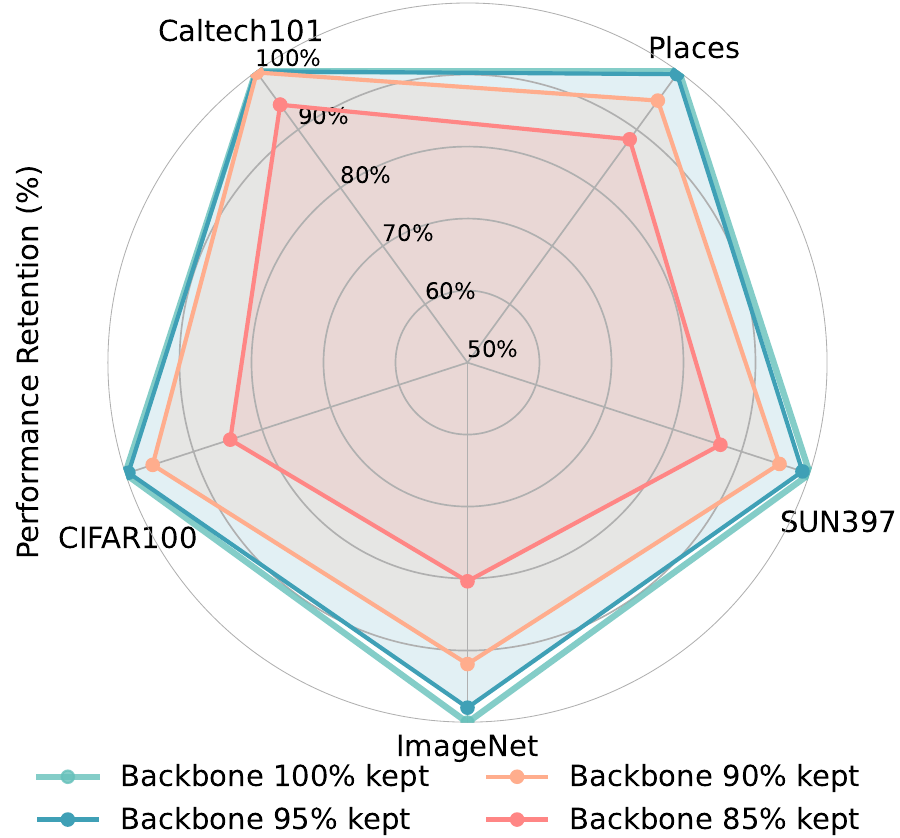}%
    \label{fig:sub_a}
}
\hfill
\subfloat[Specific-domain tasks]{
    \includegraphics[width=0.23\textwidth]{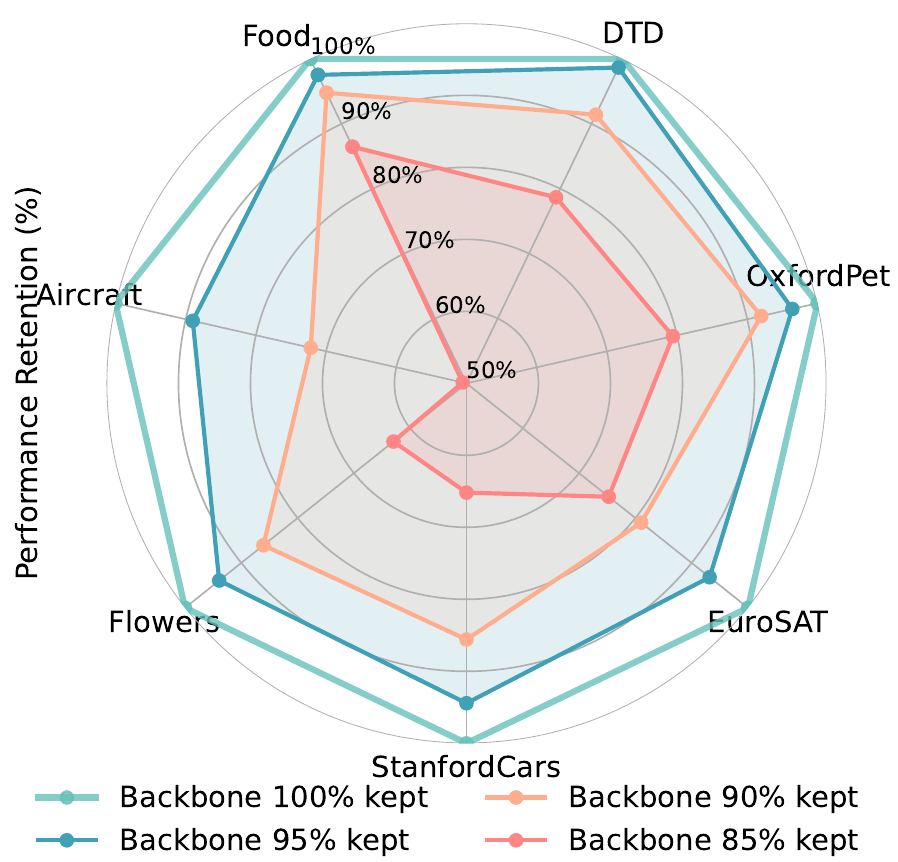}%
    \label{fig:sub_b}
}

\subfloat[Transfer accuracy]{
    \includegraphics[width=0.23\textwidth]{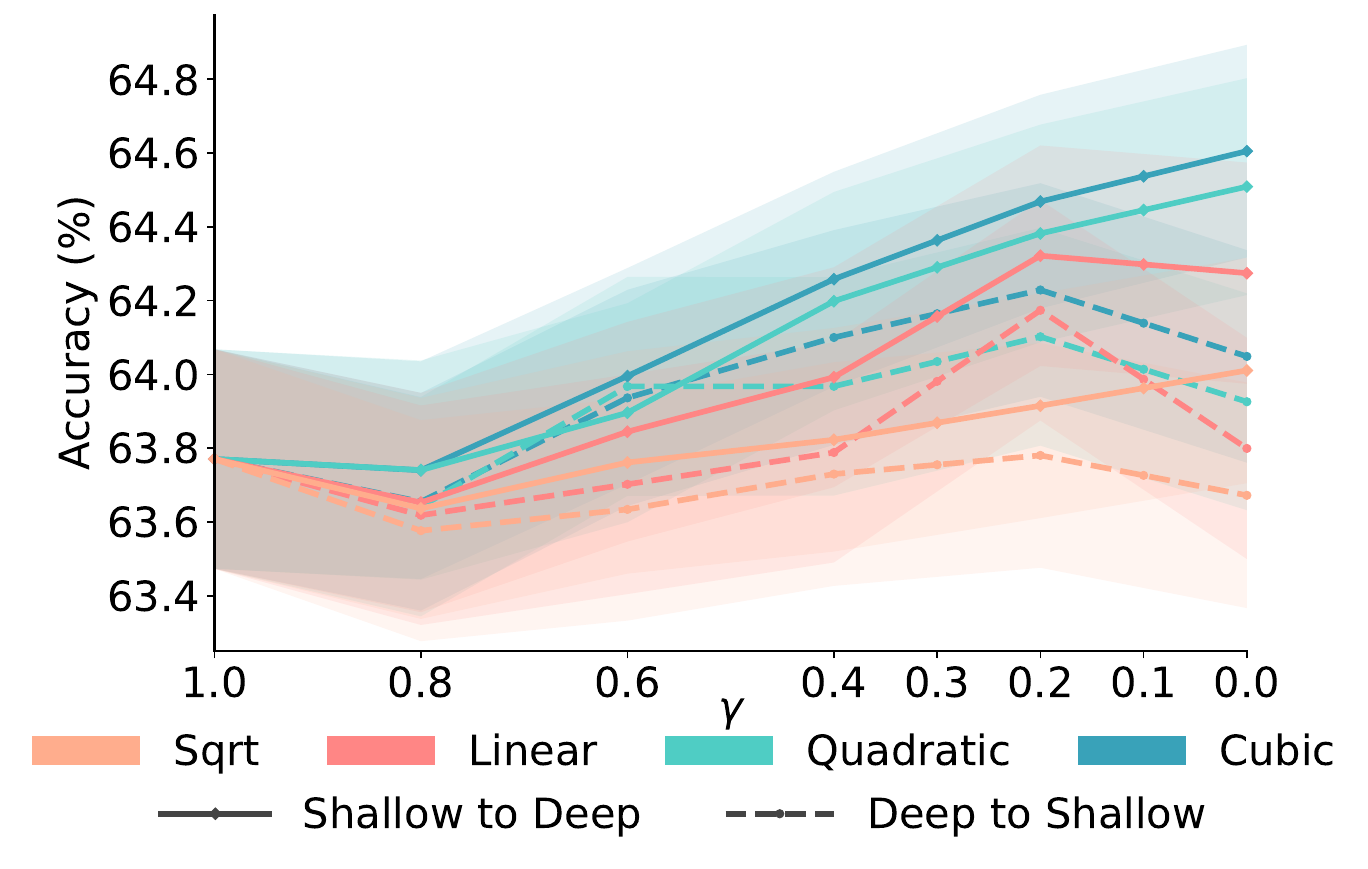}%
    \label{fig:sub_c}
}
\hfill
\subfloat[Target accuracy]{
    \includegraphics[width=0.23\textwidth]{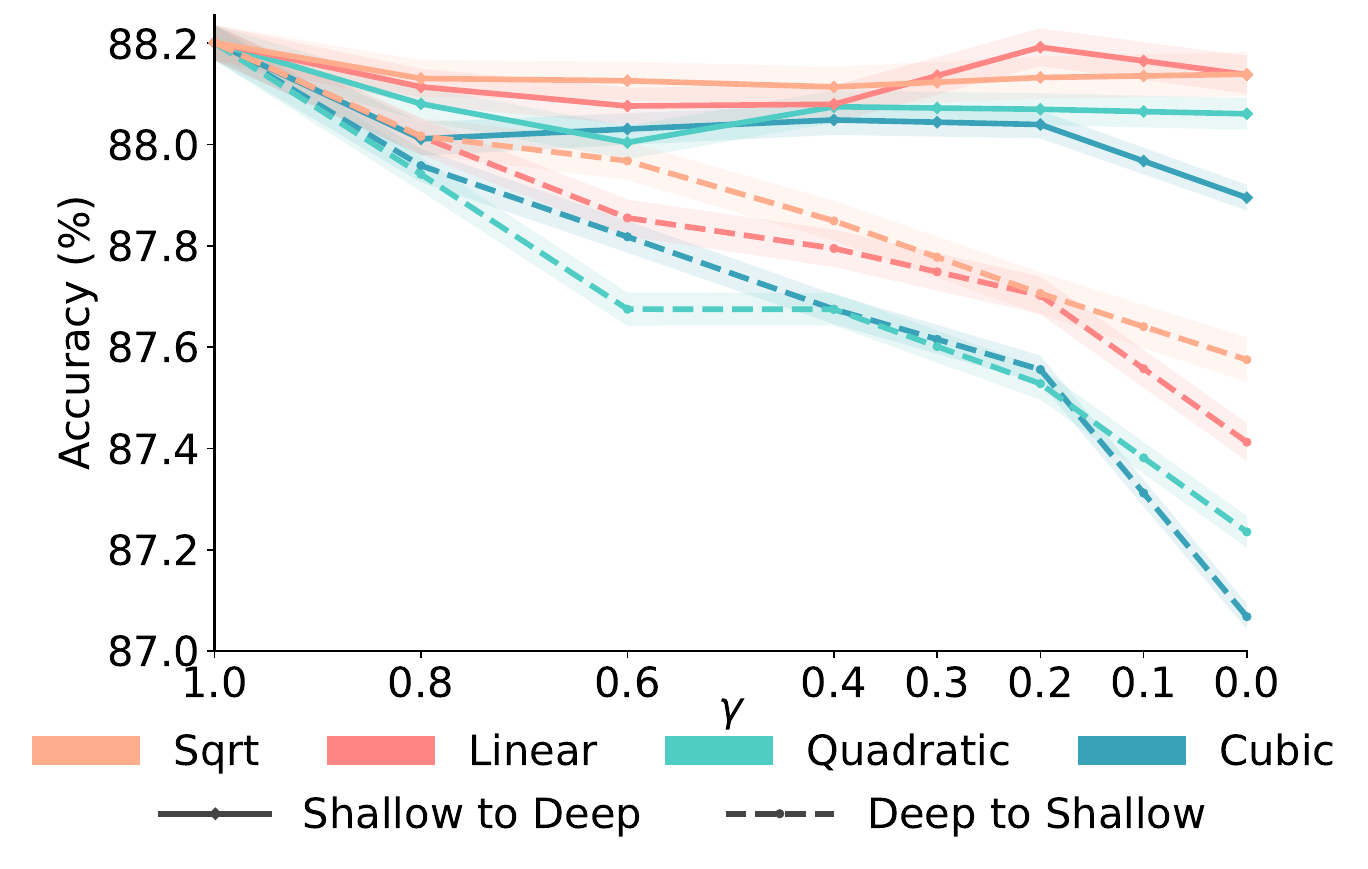}%
    \label{fig:sub_d}
}

\caption{Analysis of Transformer architecture in inter-layer structure and intra-layer parameter
space. 
In Fig.~\ref{fig:sub_a} and \ref{fig:sub_b}, we measure zero-shot performance after reconstructing attention weights using only the top principal singular components. While performance on general-domain datasets remains  robust, performance on most specific-domain datasets degrades sharply as more low-energy components are removed.
In Fig.~\ref{fig:sub_c} and \ref{fig:sub_d}, we investigate the impact of different layer-wise scaling schedules on transfer accuracy (forward stability) and target accuracy (plasticity), where $\gamma$ on the horizontal axis controls the lower bound of the scaling factor. The curves show the mean over three random seeds, and the shaded regions indicate one sample standard deviation. 
Shallow-to-deep denotes that scaling factor progressively increases from shallow to deep layers, while deep-to-shallow represents the inverse. }
\label{fig:main_figure_label}
\vspace{-0.1in}
\end{figure}

\IEEEPARstart{V}{ision-language}  models (VLMs) have demonstrated remarkable zero-shot transfer capabilities, making them cornerstones of many downstream applications \cite{comanici2025gemini, achiam2023gpt, radford2021learning}. Despite this success, their performance on certain datasets can be insufficient, motivating the need for continual learning (CL). An effective CL method requires balancing three competing objectives: maintaining the ability to learn new knowledge (\emph{plasticity}), preventing the forgetting of previously learned tasks (\emph{backward stability}), and crucially, preserving the general pre-trained knowledge that guarantees general transferability (\emph{forward stability}) \cite{gu2026spectral,mukhoti2024finetuning, zheng2023preventing}. Degradation of this pre-trained knowledge is particularly detrimental, as it erodes the core value of VLMs. Therefore, the central challenge is how to effectively learn new knowledge without undermining these critical stability constraints.

Existing continual learning methods often rely on replay or reference-data regularization to preserve previously acquired knowledge \cite{wang2024comprehensive,zhou2024survey,zheng2023preventing}. However, pre-training data are frequently unavailable, while reference-data methods introduce additional computational costs and are sensitive to the choice of reference data \cite{wu2025synthetic,zheng2023preventing}. To avoid these limitations, parameter-efficient methods seek to preserve knowledge by constraining parameter updates within specific subspaces. O-LoRA \cite{wang2023orthogonal} enforces orthogonality between LoRA subspaces of different tasks, while InfLoRA \cite{liang2024inflora} restricts updates to a subspace orthogonal to previously learned feature directions. KeepLoRA \cite{luo2026keeplora} further considers the preservation of pre-trained knowledge. As illustrated in Fig.~\ref{fig:sub_a} and Fig.~\ref{fig:sub_b}, its analysis reveals that the principal subspace of pre-trained weights primarily captures general transferable knowledge, while the residual subspace contains important information for domain-specific adaptation. Accordingly, KeepLoRA \cite{luo2026keeplora} combines the principal subspace of pre-trained weights with accumulated previous-task directions and employs gradient-informed residual subspace adaptation to balance stability and plasticity. However, these methods primarily focus on intra-layer parameter subspaces and update directions, without explicitly considering how knowledge retention sensitivity and update magnitudes vary across Transformer \cite{attention_is_all} layers.

Motivated by the insight from model merging that shallow layers should remain close to their pre-trained weights to avoid losing general representations \cite{wang2025lines}, we investigate the impact of applying different scaling factors across layers during training. Specifically, we apply a layer-wise scaling function modulated by a parameter $\gamma$, which scales the original base factor $\alpha_0$ across different layers. We compare two layer-wise scaling schedules: shallow-to-deep, in which scaling factors increase with depth, and deep-to-shallow, in which they decrease. As shown in Fig.~\ref{fig:sub_c} and Fig.~\ref{fig:sub_d}, the shallow-to-deep schedule preserves transfer performance significantly better while maintaining target task accuracy. In contrast, the deep-to-shallow schedule, which assigns larger updates to shallow layers, severely degrades both transferability and target task performance. These findings verify that deeper layers are primarily responsible for acquiring new domain-specific knowledge, whereas shallow layers must remain stable to retain general representations. Consequently, these observations form the basis of our approach. To coherently achieve \emph{forward stability}, \emph{backward stability}, and \emph{plasticity}, continual learning updates should be projected onto the residual subspace of the parameters while allocating smaller updates to shallow layers.

Building on these observations, we propose KeepLoRA++, a unified framework that jointly controls update directions within each layer and update magnitudes across Transformer \cite{attention_is_all} layers. Compared with KeepLoRA \cite{luo2026keeplora}, which focuses on intra-layer residual gradient adaptation, KeepLoRA++ further incorporates a shallow-to-deep layer-wise scaling schedule. To provide theoretical support for this design, we derive an optimal allocation of update magnitudes based on layer-wise retention curvature under a fixed first-order improvement in the current-task loss, motivating the proposed shallow-to-deep scaling strategy. Furthermore, we simplify the original LoRA initialization by setting the up-projection matrix to zero, eliminating the need for explicit backbone weight compensation. We theoretically establish its equivalence to the compensated non-zero initialization of KeepLoRA \cite{luo2026keeplora} under specified optimization conditions with a frozen down-projection matrix. This simplification facilitates integration with existing training frameworks. Beyond these methodological and theoretical extensions, we expand the empirical evaluation to video continual learning and conduct extensive experiments across image classification, visual question answering, and video understanding tasks, together with additional computational efficiency analyses.

Our contributions are summarized as follows:

\begin{itemize}
    \item We investigate knowledge retention in Transformers from both intra-layer and inter-layer perspectives, revealing that shallow layers are more sensitive to general knowledge disruption, while deeper layers offer greater flexibility for task-specific adaptation.

    \item We propose KeepLoRA++, integrating residual gradient adaptation with layer-wise scaling. We further establish theoretical results on optimal update allocation and initialization equivalence, simplifying implementation while balancing stability and plasticity.

    \item Extensive experiments on CLIP, LLaVA, and Video-LLaVA demonstrate strong performance across image classification, visual question answering, and video continual learning, with comprehensive ablation and efficiency analyses.
\end{itemize}

\section{Related Work}

\subsection{Intra-Layer Parameter Adaptation}

Intra-layer methods primarily focus on constraining parameter updates to preserve existing knowledge while maintaining adaptation capability. Gradient Projection Memory (GPM) \cite{saha2021gradient} prevents interference with previous tasks by enforcing orthogonality between new task gradients and stored feature directions, while CoSo \cite{cheng2026continuous} improves the efficiency of subspace optimization. This idea has also been extended to parameter-efficient adaptation. O-LoRA \cite{wang2023orthogonal} constrains the LoRA subspaces of new tasks to be orthogonal to those of previous tasks, while InfLoRA \cite{liang2024inflora} initializes LoRA within a subspace orthogonal to previously learned feature directions. From the perspective of parameter initialization, PiSSA \cite{meng2024pissa} employs singular value decomposition (SVD) to initialize LoRA matrices using the principal components of pre-trained weights, improving adaptation efficiency. However, these approaches do not explicitly address the simultaneous preservation of pre-trained knowledge and previously learned tasks. KeepLoRA \cite{luo2026keeplora} addresses this challenge by identifying a unified principal subspace of pre-trained parameters and previous task directions, and restricting gradient-informed updates to its orthogonal residual subspace.

\subsection{Inter-Layer Scaling}

Inter-layer methods investigate how adaptation strength and parameter capacity should be distributed across different Transformer layers. BEiT \cite{bao2022beit} and MAE \cite{he2022masked} employ layer-wise learning rate decay during fine-tuning, assigning smaller learning rates to shallow layers. AdaLoRA \cite{zhang2023adaptive} adaptively allocates low-rank parameter budgets across weight matrices based on their importance, while LISA \cite{pan2024lisa} exploits layer-wise differences to select Transformer layers for efficient fine-tuning. For LoRA-specific scaling, VeRA \cite{kopiczko2024vera} shares frozen random low-rank matrices across layers and learns lightweight scaling vectors. TLoRA \cite{islam2025tlora} introduces a tri-matrix decomposition with learnable layer-wise scaling factors, while MSLoRA \cite{luo2025mslora} employs meta-learning to optimize layer-specific scaling factors. LiNeS \cite{wang2025lines} applies depth-dependent scaling to parameter updates after fine-tuning to mitigate forgetting and improve model merging. These methods primarily focus on model adaptation efficiency or post-training parameter editing, without investigating how inter-layer scaling complements intra-layer residual gradient adaptation to balance knowledge retention and plasticity throughout sequential continual learning.

\section{Motivation and Analysis}
\label{sec:motivation}

We investigate continual learning from two complementary dimensions of the Transformer architecture. At the intra-layer level, we analyze how to select update directions that preserve existing knowledge while maintaining plasticity. At the inter-layer level, we investigate how update magnitudes should be distributed across layers with different sensitivities to knowledge disruption.

\subsection{Preliminaries}

\textbf{Problem Formulation.} We adopt a general continual learning setting for vision-language models, where the model sequentially encounters a sequence of $n$ tasks $\{\mathcal{T}^1, \mathcal{T}^2, \ldots, \mathcal{T}^n\}$. 
Each task $\mathcal{T}^i = \{(\boldsymbol{x}_j^i, \boldsymbol{y}_j^i)\}_{j=1}^{N_i}$ contains $N_i$ training examples. 
To accommodate both classification and generative tasks, we formulate the inputs and targets in a unified multimodal format.
For classification tasks, the input $\boldsymbol{x}_j^i$ denotes an image, and the target $\boldsymbol{y}_j^i$ represents a categorical class label associated with a semantic class name. 
For generative tasks, such as visual question answering and image captioning, the input $\boldsymbol{x}_j^i$ consists of an image paired with a textual query or instruction, while the target $\boldsymbol{y}_j^i$ is a sequence of text tokens representing the ground-truth response. 
During inference, given an input $\boldsymbol{x}$ from the active task, the model is required to predict the corresponding target $\boldsymbol{y}$, which is evaluated using task-appropriate metrics such as classification accuracy or language generation quality. 

\textbf{Vanilla LoRA.} Low-rank adaptation (LoRA) \cite{hu2022lora} decomposes weight updates into two low-rank matrices $\boldsymbol{A} \in \mathbb{R}^{d_{\text{in}} \times r}$ and $\boldsymbol{B} \in \mathbb{R}^{r \times d_{\text{out}}}$, where $r \ll \min(d_{\text{in}}, d_{\text{out}})$. During training, $\boldsymbol{W}$ remains frozen while only $\boldsymbol{A}$ and $\boldsymbol{B}$ are fine-tuned. The matrices are initialized with $\boldsymbol{A} \sim \mathcal{N}(0, \sigma^2)$ and $\boldsymbol{B} = \boldsymbol{0}$. For an input $\boldsymbol{x} \in \mathbb{R}^{d_{\text{in}}}$, the forward pass becomes:
\begin{equation}
\boldsymbol{h} = \boldsymbol{x}\boldsymbol{W} + \frac{\alpha_0}{r}\boldsymbol{x}\boldsymbol{AB}
\end{equation}
where $\alpha_0$ is a constant scaling factor applied uniformly across layers.

\subsection{Intra-Layer: Residual Subspace Adaptation}

We analyze the intra-layer parameter space of Transformer architecture to understand how pre-trained knowledge and task-specific information are distributed within model parameters. As revealed in Fig.~\ref{fig:sub_a} and Fig.~\ref{fig:sub_b}, the principal subspace mainly preserves general transferable knowledge, while the residual subspace contains important information for domain-specific adaptation. Specifically, for each pre-trained weight matrix $\boldsymbol{W} \in \mathbb{R}^{d_{in} \times d_{out}}$, we perform SVD as $\boldsymbol{W} = \boldsymbol{USV}^{\top}$ and define the principal subspace as $\boldsymbol{W}_p = \boldsymbol{U}_{:,1:p}$, where $p$ is determined by the energy constraint $||\boldsymbol{W}_p||_F^2 \geq \epsilon_w ||\boldsymbol{W}||_F^2$ and $\epsilon_w \in (0,1)$ controls the retained energy. 
Since no real or synthetic samples from previous tasks are available for replay, we retain their dominant gradient directions to preserve previously learned knowledge and minimize interference with subsequent tasks. Specifically, we maintain an accumulated direction matrix $\boldsymbol{M}_{t-1} \in \mathbb{R}^{d_{in} \times k}$ containing the dominant singular vectors of previous task gradients, where $k$ is determined by the energy threshold $\epsilon_f$ and $\boldsymbol{M}_0=\emptyset$. 
Since $\boldsymbol{W}_p$ and $\boldsymbol{M}_t$ consist of orthogonal direction vectors in the same $d_{in}$-dimensional feature space, they can be unified into a single principal subspace $\boldsymbol{M}_t'=[\boldsymbol{W}_p,\boldsymbol{M}_t]$. This unified representation captures the essential feature representations of pre-training data and the dominant gradient directions acquired during continual learning, consistent with the perspective that predictive models can be transformed into lossless compressors \cite{deletanglanguage} and model weights embody compressed representations of training data \cite{franceschelli2024training}. For the current task $\mathcal{T}^t$, let $\boldsymbol{G}_t=\nabla_{\boldsymbol{W}}\mathcal{L}(\boldsymbol{W};\mathcal{T}^t)$ denote the gradient of the weight matrix $\boldsymbol{W}$ at the first training step. To preserve existing knowledge while maintaining plasticity, the desired update subspace should be orthogonal to both $\boldsymbol{W}_p$ and $\boldsymbol{M}_{t-1}$, while capturing the dominant directions of $\boldsymbol{G}_t$. We first characterize the relationship between frozen LoRA down-projection and gradient projection.

\begin{proposition}{(LoRA with frozen down-projection $\boldsymbol{A}_t$ is equivalent to gradient projection update.)}
\label{pro:p1}
Let $\mathcal{L}(\boldsymbol{W}; \mathcal{D}^t)$ denote the loss function for the $t$-th task $\mathcal{T}^t$, where:
$\boldsymbol{W} = \boldsymbol{W}' + \frac{\alpha}{r}\boldsymbol{A}_{t}\boldsymbol{B}_{t}$,
$\boldsymbol{G}_{t} = \nabla_{\boldsymbol{W}}\mathcal{L}(\boldsymbol{W}; \mathcal{D}^t)$.
Optimizing only $\boldsymbol{B}_{t}$ through gradient descent with learning rate $\eta$ is equivalent to performing gradient descent on the orthogonal projection of $\boldsymbol{G}_{t}$ onto $\text{span}(\boldsymbol{A}_t)$. The weight update of $\boldsymbol{W}$ satisfies:
\begin{equation}
\Delta \boldsymbol{W} = \frac{\alpha}{r} \boldsymbol{A}_{t} \Delta \boldsymbol{B}_{t} = -c \boldsymbol{A}_{t} \boldsymbol{A}_{t}^{\top} \boldsymbol{G}_{t},
\end{equation}
where $c = \frac{\eta \alpha^{2}}{r^{2}}$ is a positive constant integrating the learning rate and LoRA scaling effects.
\end{proposition}

Proposition~\ref{pro:p1} establishes that freezing $\boldsymbol{A}_t$ confines weight updates to $\text{span}(\boldsymbol{A}_t)$, with $\boldsymbol{A}_t\boldsymbol{A}_t^\top$ acting as the corresponding orthogonal projection operator. Therefore, the choice of $\boldsymbol{A}_t$ determines the update directions and should satisfy two requirements simultaneously. It must be orthogonal to $\text{span}(\boldsymbol{W}_p)$ and $\text{span}(\boldsymbol{M}_{t-1})$ to avoid interfering with pre-trained and previously learned knowledge, while capturing the dominant directions of $\boldsymbol{G}_t$ to maintain plasticity. We formulate these requirements as a constrained optimization problem, and the following proposition establishes the optimality of the gradient-informed initialization.

\begin{proposition}
\label{pro:p2}
Let $\boldsymbol{Q}=[\boldsymbol{W}_p,\boldsymbol{M}_{t-1}]$ have orthonormal columns. If $r\leq\operatorname{rank}(\hat{\boldsymbol{G}}_t)$, the KeepLoRA++ initialization of $\boldsymbol{A}_t$ in Eq.~\ref{eq5} and Eq.~\ref{eq4} is an optimal solution to the following problem:
\begin{equation}
\begin{aligned}
\label{eq:optimization}
\min_{\boldsymbol{A}_{t}^{\top} \boldsymbol{A}_{t} = \bm{I}} \| \boldsymbol{G}_{t} - \boldsymbol{A}_{t} \boldsymbol{A}_{t}^{\top} \boldsymbol{G}_{t} \|_F^2, &
\\ \text{s.t} \quad \boldsymbol{W}_{p} ^{\top} \boldsymbol{A}_{t} =\boldsymbol{M}_{t-1}^{\top} \boldsymbol{A}_{t} = \bm{0},
\end{aligned}
\end{equation}
where $\boldsymbol{G}_t$ is the current task gradient w.r.t. the base model $\boldsymbol{W}$, $\boldsymbol{W}_p$ is the principal subspace of pre-trained parameters, and $\boldsymbol{M}_{t-1}$ is the dominant gradient directions from previous tasks.
\end{proposition}

Proposition~\ref{pro:p2} establishes that the selected update subspace satisfies both requirements. The orthogonality constraints preserve pre-trained and previously acquired knowledge, while minimizing the residual component of $\boldsymbol{G}_t$ outside $\text{span}(\boldsymbol{A}_t)$ is equivalent, by the Pythagorean theorem, to maximizing the projected gradient norm $\|\boldsymbol{A}_t\boldsymbol{A}_t^\top\boldsymbol{G}_t\|_F^2$. Consequently, the selected subspace captures the dominant directions of the current task gradient without interfering with existing knowledge, balancing stability and plasticity within each Transformer layer. However, this intra-layer analysis does not account for how update magnitudes should be distributed across different Transformer layers. %

\begin{figure}[!t]
    \centering
    \includegraphics[width=0.95\linewidth]{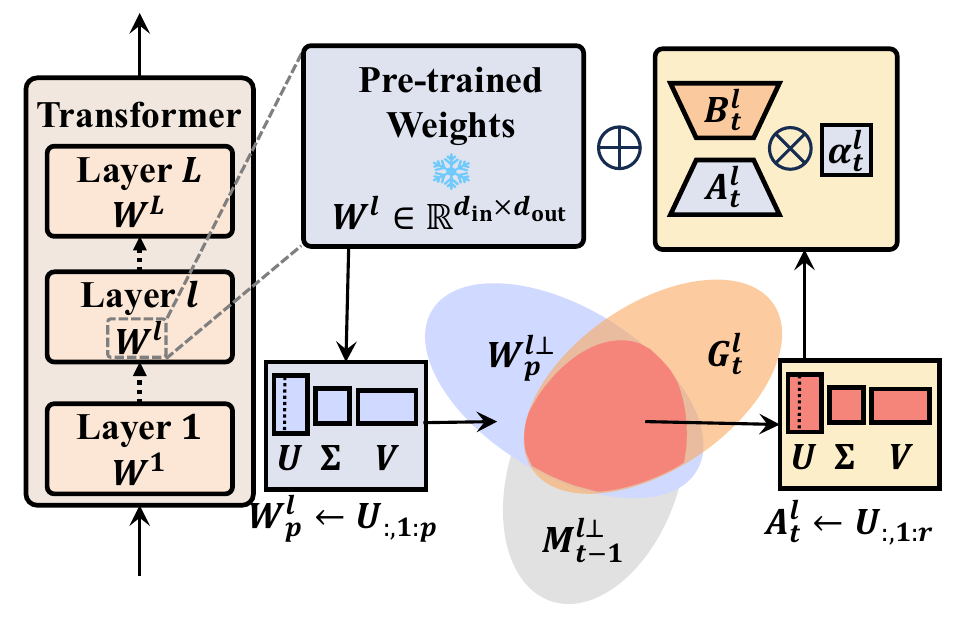}
    \caption{Illustration of the KeepLoRA++ mechanism at layer $l$. The pre-trained weights $\boldsymbol{W}^l$ are decomposed by SVD to define the principal subspace $\boldsymbol{W}_p^l$. The current task gradient $\boldsymbol{G}_t^l$ is projected onto the residual subspace orthogonal to both $\boldsymbol{W}_p^l$ and the accumulated previous task directions $\boldsymbol{M}_{t-1}^l$ to maintain forward and backward stability. The blue and gray regions represent the orthogonal complements $\boldsymbol{W}_{p}^{l \perp}$ and $\boldsymbol{M}_{t-1}^{l \perp}$, respectively, while the red overlapping region represents the retained residual gradient $\hat{\boldsymbol{G}}_t^l$. SVD on this projected gradient $\hat{\boldsymbol{G}}_t^l$ yields the initialization for the frozen down-projection matrix $\boldsymbol{A}_t^l$. Only the up-projection matrix $\boldsymbol{B}_t^l$ is learnable, and the final low-rank update is scaled by a layer-specific factor $\alpha_t^l$.}
    \label{fig:method}
\end{figure}

\subsection{Inter-Layer: Larger Updates in Deeper Layers}

We further analyze the inter-layer structure of the Transformer architecture to investigate how update magnitudes should be distributed across different layers. As shown in Fig.~\ref{fig:sub_c} and Fig.~\ref{fig:sub_d}, the shallow-to-deep layer-wise scaling schedule preserves transfer performance while maintaining target-task accuracy. In contrast, the deep-to-shallow schedule, which assigns larger updates to shallow layers, severely degrades both transferability and target task performance. Furthermore, among the various transition profiles evaluated, the linear transition exhibits an optimal and robust balance across both metrics. These observations indicate that shallow layers are more sensitive to modifications that may disrupt general representations, whereas deeper layers provide greater flexibility for task-specific adaptation. While Propositions~\ref{pro:p1} and~\ref{pro:p2} characterize the projected update and its gradient-informed residual subspace, they do not account for differences in knowledge retention sensitivity across layers. To characterize the allocation of update magnitudes, we compare the quadratic retention costs of different allocations at the same first-order decrease in the current-task loss. The following proposition formulates this comparison as a constrained optimization problem.

\begin{proposition}
\label{pro:layer_scaling}
Let $\mathcal{L}_{\mathrm{ret}}$ denote a loss measuring retention of pre-trained or previously learned knowledge, and let $\kappa^l>0$ be layer-wise upper bounds on its local curvature in the trainable subspaces. A larger $\kappa^l$ assigns a higher quadratic retention cost to the same update magnitude.

At a fixed iterate with frozen orthonormal $\boldsymbol{A}_t^l$, Proposition~\ref{pro:p1} gives the gradient-descent update $\Delta\boldsymbol{W}^l=-c^l\boldsymbol{A}_t^l(\boldsymbol{A}_t^l)^\top\boldsymbol{G}_t^l$, where $c^l=\eta(\alpha^l)^2/r^2$ and $\boldsymbol{G}_t^l$ is the current-task gradient. A local expansion of $\mathcal{L}_{\mathrm{ret}}$ bounds its second-order contribution by $\mathcal{C}(\{c^l\})=\frac12\sum_l\kappa^l\|\Delta\boldsymbol{W}^l\|_F^2$. Fix the first-order decrease in the current-task loss at $\delta=-\sum_l\langle\boldsymbol{G}_t^l,\Delta\boldsymbol{W}^l\rangle>0$. Substituting the projected update into this cost and progress constraint yields
\begin{equation}
\label{eq:ls_allocation}
\begin{aligned}
\min_{\{c^l\geq0\}}\frac12\sum_l\kappa^l(c^l)^2
\|(\boldsymbol{A}_t^l)^\top\boldsymbol{G}_t^l\|_F^2, &\\
\text{s.t.}\quad
\sum_l c^l\|(\boldsymbol{A}_t^l)^\top\boldsymbol{G}_t^l\|_F^2=\delta.
\end{aligned}
\end{equation}
All sums are over the nonempty set of layers with nonzero projected gradients. The unique minimizer and corresponding LoRA scales satisfy
\begin{equation}
\label{eq:ls_optimal}
c^{l,\star}\propto(\kappa^l)^{-1},\quad
\alpha^{l,\star}\propto(\kappa^l)^{-1/2}.
\end{equation}
Choose $\delta$ so that the updated weights remain in the neighborhood where the curvature bounds hold. The retention-risk bound and proof are provided in Appendix \ref{appendix:layer_scaling}.
\end{proposition}

Proposition~\ref{pro:layer_scaling} establishes that the optimal update allocation is inversely proportional to the layer-wise retention curvature, with $c^{l,\star}\propto(\kappa^l)^{-1}$ and $\alpha^{l,\star}\propto(\kappa^l)^{-1/2}$. When shallow layers have higher retention curvature, this allocation favors smaller updates in shallow layers and larger updates in deeper layers, consistent with the shallow-to-deep schedule. Together with Propositions~\ref{pro:p1} and~\ref{pro:p2}, this result establishes the complementary roles of intra-layer update direction selection and inter-layer update magnitude allocation in balancing stability and plasticity. The depth-correlation condition and finite-strength criterion for this schedule are derived in Appendix~\ref{appendix:layer_scaling}.

\begin{figure*}[!t]
\centering
\begin{minipage}{0.95\textwidth}
\begin{algorithm}[H]
\caption{KeepLoRA++ for Continual Learning}
\label{alg1}
\begin{algorithmic}[1]
\STATE \textbf{Input:} Pre-trained model $f_\theta$ with $L$ target layers, updatable parameters $\{\boldsymbol{B}_t^l\}$, task sequence $\{\mathcal{T}^t\}_{t=1}^n$, hyperparameters $\epsilon_w, \epsilon_f, r, \alpha_0, \gamma$
\STATE \textbf{Output:} Updated model $f_{\theta'}$ with merged KeepLoRA++ adapters

\FOR{task $t = 1$ to $n$}
    \STATE Compute layer-specific scaling factors $\{\alpha^l\}_{l=1}^L$ via Eq.~\ref{eq:scale}
    \STATE Initialize KeepLoRA++ matrices $\boldsymbol{A}_t$ and $\boldsymbol{B}_t$ in the residual subspace via Eq.~\ref{eq5} and Eq.~\ref{eq4}
    \STATE Compute the loss on $\mathcal{T}^t$ and optimize the parameters $\boldsymbol{B}_t$
    \STATE Merge KeepLoRA++ and current model by $\boldsymbol{W}_t^l = \boldsymbol{W}_{t-1}^l + \frac{\alpha^l}{r}\boldsymbol{A}_t^l\boldsymbol{B}_t^l$
    \STATE Extract and store dominant gradient directions $\boldsymbol{M}_t$ through Eq.~\ref{eq5} and Eq.~\ref{eq6}
\ENDFOR
\end{algorithmic}
\end{algorithm}
\end{minipage}
\end{figure*}

\section{KeepLoRA++ Framework}

Based on the intra-layer and inter-layer analyses in Section~\ref{sec:motivation}, we propose KeepLoRA++ to jointly control update directions and magnitudes across Transformer layers. Proposition~\ref{pro:p1} establishes that freezing the down-projection matrix $\boldsymbol{A}_t$ constrains updates to its column space, while Proposition~\ref{pro:p2} characterizes the gradient-informed residual subspace that preserves existing knowledge and current-task plasticity. Proposition~\ref{pro:layer_scaling} further establishes the optimal allocation of update magnitudes according to layer-wise retention sensitivity. Accordingly, KeepLoRA++ integrates residual gradient projection with depth-dependent layer-wise scaling, as illustrated in Fig.~\ref{fig:method} and Algorithm~\ref{alg1}.

For each task $\mathcal{T}^t$, we first project the initial task gradient $\boldsymbol{G}_t$ onto the residual subspace orthogonal to the principal subspace of pre-trained weights $\boldsymbol{W}_p$ and the accumulated dominant gradient directions of previous tasks $\boldsymbol{M}_{t-1}$. The resulting projected gradient is given by
\begin{equation}
\label{eq5}
\boldsymbol{\hat{G}}_t = \boldsymbol{G}_t - \boldsymbol{W}_p \boldsymbol{W}_p^{\top}\boldsymbol{G}_t - \boldsymbol{M}_{t-1}\boldsymbol{M}_{t-1}^{\top}\boldsymbol{G}_t
\end{equation}
The number of dominant gradient directions retained for previous tasks is dynamically determined by the energy threshold $\epsilon_f \in (0,1)$. Specifically, we retain the minimum number $k$ of dominant directions required to satisfy
\begin{equation}
\label{eq6}
||\boldsymbol{\hat{G}}_t||^2_F + || \boldsymbol{W}_p \boldsymbol{W}_p^{\top}\boldsymbol{G}_t ||^2_F + || \boldsymbol{M}_{t-1}\boldsymbol{M}_{t-1}^{\top}\boldsymbol{G}_t ||^2_F \geq \epsilon_f ||\boldsymbol{G}_t||^2_F
\end{equation}
Then, SVD is performed on the projected gradient $\boldsymbol{\hat{G}}_t = \boldsymbol{U}\boldsymbol{S}\boldsymbol{V}^{\top}$ and LoRA matrices are initialized using the top-$r$ left singular vectors:
\begin{equation}
    \label{eq4}
    \boldsymbol{A} = \boldsymbol{U}_{:,1:r}, \quad \boldsymbol{B} = \boldsymbol{0}
\end{equation}
where $\boldsymbol{U}_{:,1:r}$ denotes the first $r$ columns of $\boldsymbol{U}$, and $r$ is the rank parameter. During training, $\boldsymbol{A}_t$ remains frozen and only $\boldsymbol{B}_t$ is optimized. The zero initialization of $\boldsymbol{B}_t$ preserves the initial forward behavior without requiring compensation of the frozen backbone weights. As established in Proposition~\ref{pro:b_zero_equivalence}, it is theoretically equivalent to the compensated non-zero initialization adopted in the original KeepLoRA \cite{luo2026keeplora}. %

To further account for the differences in knowledge retention sensitivity across Transformer depths, we adopt the linear shallow-to-deep scaling schedule motivated by Proposition~\ref{pro:layer_scaling}. Specifically, for the $l$-th layer in an $L$-layer network, the layer-specific scaling factor is defined as
\begin{equation}
\label{eq:scale}
    \alpha^l = \varphi(l, L, \alpha_0, \gamma) = \alpha_0 \left( \frac{l-1}{L-1} \cdot 1 + \frac{L-l}{L-1} \cdot \gamma \right)
\end{equation}
where $\alpha_0$ is the base scaling factor and $\gamma \in (0,1]$ controls the lower bound of the layer-wise scaling factor. This formulation assigns the smallest scaling factor $\gamma\alpha_0$ to the shallowest layer and the base scaling factor $\alpha_0$ to the deepest layer, allowing smaller modifications to preserve general representations in shallow layers while retaining greater adaptation flexibility in deeper layers. The final update at each layer is obtained by combining the residual gradient-informed LoRA adaptation with its corresponding layer-specific scaling factor.

After training on task $t$, we perform SVD on the projected gradient $\boldsymbol{\hat{G}}_t = \boldsymbol{U}_t\boldsymbol{S}_t\boldsymbol{V}_t^{\top}$ and extract the top-$m$ dominant singular vectors to update the accumulated direction matrix as $\boldsymbol{M}_t = [\boldsymbol{M}_{t-1}, \boldsymbol{U}_{t(:,1:m)}]$, where $m$ is determined by the energy threshold $\epsilon_f$ in Eq.~\ref{eq6}.

We finally establish the theoretical equivalence between the zero initialization adopted in KeepLoRA++ and the compensated non-zero initialization in the original KeepLoRA. In the original KeepLoRA, the non-zero initial update requires an explicit compensation of the frozen backbone weights to preserve the initial model behavior. In contrast, KeepLoRA++ directly initializes $\boldsymbol{B}_t$ to zero. The following proposition establishes that these two parameterizations yield identical effective weights throughout training when $\boldsymbol{A}_t$ is frozen.

\begin{proposition}{(Zero and compensated non-zero initializations
yield equivalent updates under frozen $\boldsymbol{A}_t$.)}
\label{pro:b_zero_equivalence}
Let $\boldsymbol{A}_t$ be fixed and
$\boldsymbol{W}'=\boldsymbol{W}-s\boldsymbol{A}_t
\boldsymbol{B}_t^{(0)}$, where $s$ is a fixed scaling factor.
Consider gradient descent with the same loss, data, and
learning rate $\eta$, optimizing only $\boldsymbol{B}_t$
and $\widetilde{\boldsymbol{B}}_t$, initialized at
$\boldsymbol{B}_t^{(0)}$ and $\boldsymbol{0}$, respectively.
For any iteration $k\geq 0$, the two parameterizations satisfy:
\begin{equation}
\begin{aligned}
\widetilde{\boldsymbol{B}}_t^{(k)}
&=\boldsymbol{B}_t^{(k)}-\boldsymbol{B}_t^{(0)},\\
\boldsymbol{W}'+s\boldsymbol{A}_t\boldsymbol{B}_t^{(k)}
&=\boldsymbol{W}
+s\boldsymbol{A}_t\widetilde{\boldsymbol{B}}_t^{(k)}.
\end{aligned}
\end{equation}
\end{proposition}

Proposition~\ref{pro:b_zero_equivalence} shows that zero initialization eliminates explicit weight compensation without altering the optimization trajectory under the stated conditions. This equivalence also holds for gradient-based optimizers such as SGD with momentum, Adam, and RMSProp, given identical optimizer states and hyperparameters. However, the equivalence may not hold under parameter-dependent regularization or non-zero weight decay, as in AdamW. By avoiding explicit modifications to the frozen backbone weights, the zero initialization also simplifies implementation and facilitates integration with distributed training frameworks.

\section{Experiments}
We conducted extensive experiments across multiple benchmarks including image classification tasks, visual question answering (VQA), and video understanding to validate the effectiveness of KeepLoRA++ in balancing three core objectives of continual learning: forward stability, backward stability, and plasticity. 
Specifically, our experimental analysis is structured as follows:
(i) To evaluate \textbf{forward stability}, we utilize the \emph{Transfer} metric (presented in Table~\ref{tab:MTIL}, \ref{tab:DCL_methods}, \ref{tab:UCIT_methods}, and \ref{tab:CL_VISTA_methods}), which quantifies the preservation of zero-shot transferability on unseen tasks. Fig.~\ref{fig:Interference} further visually demonstrates how KeepLoRA++ minimizes structural disruption to maintain this capability.
(ii) To assess \textbf{backward stability}, we rely on the \emph{PostAverage} alongside the \emph{Last} metric, which isolate the  capacity to retain previously acquired knowledge. This is also corroborated by Fig.~\ref{fig:Interference}, showing minimal inter-task interference.
(iii) To analyze \textbf{plasticity}, Fig.~\ref{fig:plasticity} compares the learning capacity of KeepLoRA++ against the LoRA baseline, confirming that KeepLoRA++ preserves stringent stability constraints with minimal sacrifice to its adaptive capability.

\subsection{Benchmarks}
\label{sec:benchmark}

\textbf{MTIL }benchmark~\cite{zheng2023preventing} consists of 11 image classification datasets: Aircraft~\cite{maji2013fine}, Caltech101~\cite{fei2004learning}, CIFAR100~\cite{krizhevsky2009learning}, DTD~\cite{cimpoi2014describing}, EuroSAT~\cite{helber2019eurosat}, Flowers~\cite{nilsback2008automated}, Food~\cite{bossard2014food}, MNIST~\cite{deng2012mnist}, OxfordPet~\cite{parkhi2012cats}, StanfordCars~\cite{krause20133d}, and SUN397~\cite{xiao2010sun}. Each dataset is treated as a task.

\textbf{MLLM-DCL }benchmark~\cite{zhao2025mllmcl} consists of multiple downstream VQA datasets: RSVQA~\cite{lobry2020rsvqa}, PathVQA~\cite{he2020pathvqa}, DriveLM~\cite{sima2024drivelm}, FinVis~\cite{wang2023finvisgpt}, AI2D~\cite{kembhavi2016diagram}, Sciverse~\cite{guo2025sciverse}, MapQA ~\cite{chang2022mapqa}, and TQA~\cite{kembhavi2017sixthgrader}. It covers 5 specialized areas: Remote Sensing, Medical, Driving, Finance, and Science. Each area is treated as a task.

\textbf{UCIT} benchmark~\cite{guo2025hide} consists of 6 VQA datasets: ArxivQA~\cite{li2024multimodal}, CLEVR-Math~\cite{lindstrom2022clevrmath}, IconQA~\cite{lu2021iconqa}, ImageNet-R~\cite{hendrycks2021manyfaces}, VizWiz-Caption~\cite{gurari2018vizwiz}, and Flickr30k~\cite{plummer2015flickr30k}. Each dataset is treated as a task.

\textbf{CL-VISTA }benchmark~\cite{guo2026cl} consists of 8 diverse video question answering datasets spanning perception, understanding, and reasoning: Counting (based on Molmo2-Pointing~\cite{clark2026molmo2}), Spatial (based on ScanNet~\cite{dai2017scannet} and Spatial-MLLM~\cite{wu2026spatial}), Traffic (based on TUMTraffic~\cite{zhou2025tumtraf}, RoadSocial~\cite{parikh2025roadsocial}, and LingoQA~\cite{marcu2024lingoqa}), Movie (based on CinePile~\cite{rawal2024cinepile}), GUI (based on GUI-World~\cite{chen2024gui}), Science (based on FineVideo~\cite{farre2024finevideo}), Sports (based on Sports-QA~\cite{li2026sports} and SoccerChat~\cite{gautam2025soccerchat}), and Reasoning (based on STAR~\cite{wu2024benchmark}). Each dataset is treated as a task.

\begin{table*}[t!]
\centering
\caption{\textnormal{\textbf{Comparison of different methods on MTIL for each classification task} in terms of \emph{Transfer}, \emph{Average}, and \emph{Last} scores (\%). The best results are in \textbf{bold}, and the second-best results are \underline{underlined}.}}
\label{tab:MTIL}
\begin{tabular}
{b{11em}@{\hspace{1pt}}b{1.0em}b{1.0em} *{11}{b{2.0em}}>{\centering\arraybackslash}m{1.8em}@{\hspace{0.8em}}}
\toprule
        \textbf{Method} & \rotatebox{45}{Arch. Kept} & \rotatebox{45}{w/o Extra Data} & \rotatebox{45}{Aircraft} &  \rotatebox{45}{Caltech101} & \rotatebox{45}{CIFAR100} & \rotatebox{45}{DTD} & \rotatebox{45}{EuroSAT} & \rotatebox{45}{Flowers} & \rotatebox{45}{Food} & \rotatebox{45}{MNIST} & \rotatebox{45}{OxfordPet} & \rotatebox{45}{Cars} & \rotatebox{45}{Sun397} & \textbf{Avg.}~~ \\
\midrule
\quad Zero-shot & \checkmark & \checkmark & 24.8 & 88.4 & 68.2 & 44.6 & 54.9 & 71.0 & 88.5 & 59.4 & 89.0 & 64.7 & 65.4 & \\
\hline
\multicolumn{15}{l}{\textbf{Transfer}}\\
\quad LwF {\cite{li2017learning}} & \checkmark & \ding{55} & ~~-- & 74.5 & 56.9 & 39.1 & \underline{51.1} & 52.6 & 72.8 & 60.6 & 75.1 & 30.3 & 55.9 & 56.9 \\
\quad iCaRL {\cite{rebuffi2017icarl}} & \checkmark & \ding{55} & ~~-- & 56.6 & 44.6 & 32.7 & 39.3 & 46.6 & 68.0 & 46.0 & 77.4 & 31.9 & 60.5 & 50.4 \\
\quad LwF-VR {\cite{ding2022don}} & \checkmark & \ding{55} & ~~-- & 77.1 & 61.0 & 40.5 & 45.3 & 54.4 & 74.6 & 47.9 & 76.7 & 36.3 & 58.6 & 57.2 \\
\quad WiSE-FT {\cite{wortsman2022robust}} & \checkmark & \ding{55} & ~~-- & 73.5 & 55.6 & 35.6 & 41.5 & 47.0 & 68.3 & 53.9 & 69.3 & 26.8 & 51.9 & 52.3 \\
\quad ZSCL {\cite{zheng2023preventing}} & \checkmark & \ding{55} & ~~-- & \textbf{86.0} & 67.4 & \underline{45.4} & 50.4 & \underline{69.1} & \underline{87.6} & 61.8 & 86.8 & \textbf{60.1} & \textbf{66.8} & \underline{68.1} \\
\quad O-LoRA {\cite{wang2023orthogonal}} & \checkmark & \checkmark & ~~-- & 80.8 & \underline{68.0} & 44.5 & 49.8 & 67.5 & 86.7 & 59.3 & \underline{88.7} & 56.1 & 63.6 & 66.5 \\
\quad InfLoRA {\cite{liang2024inflora}} & \checkmark & \checkmark & ~~-- & 84.3 & 67.4 & 44.3 & 50.6 & 68.2 & 87.1 & \underline{62.7} & \underline{88.7} & 57.8 & 62.8 & 67.4 \\
\quad SD-LoRA {\cite{wu2025sd}} & \checkmark & \checkmark & ~~-- & 82.3 & 67.5 & 44.4 & 51.0 & 67.9 & 87.2 & 61.1 & 88.4 & 58.2 & 63.4 & 67.1 \\
\rowcolor{gray!15}
\quad KeepLoRA++ & \checkmark & \checkmark & ~~-- & \underline{85.5} & \textbf{69.5} & \textbf{45.8} & \textbf{55.1} & \textbf{70.7} & \textbf{88.1} & \textbf{64.9} & \textbf{90.6} & \underline{60.0} & \underline{65.1} & \textbf{69.5} \\
\cmidrule{2-15}
\quad L2P {\cite{wang2022learning}} & \ding{55} & \checkmark & ~~-- & 65.6 & 50.9 & 30.4 & 41.4 & 49.3 & 71.8 & 36.3 & 77.5 & 55.3 & 53.4 & 53.2 \\
\quad DualPrompt {\cite{wang2022dualprompt}} & \ding{55} & \checkmark & ~~-- & 56.7 & 51.4 & 28.7 & 33.7 & 45.6 & 70.9 & 59.5 & 77.7 & 49.5 & 50.4 & 52.4 \\
\quad S-Prompts {\cite{wang2022s}} & \ding{55} & \checkmark & ~~-- & 67.3 & 49.4 & 26.7 & 39.7 & 47.1 & 70.2 & 34.3 & 78.9 & 56.7 & 52.2 & 52.2 \\
\quad DIKI {\cite{tang2024mind}} & \ding{55} & \checkmark & ~~-- & \underline{92.9} & \underline{69.1} & 43.2 & 43.9 & 65.4 & 85.3 & 56.0 & 88.4 & 64.0 & \underline{65.6} & 67.4 \\
\quad MoE-Adapters {\cite{yu2024boosting}} & \ding{55} & \ding{55} & ~~-- & 87.9 & 68.2 & \textbf{44.4} & \underline{49.9} & \underline{70.7} & \underline{88.7} & 59.7 & 89.1 & \underline{64.5} & 65.5 & 68.9 \\
\quad IAP {\cite{fu2026iap}} & \ding{55} & \checkmark & ~~-- & \textbf{93.0} & 68.7 & \underline{44.0} & 47.0 & 70.4 & 85.9 & \underline{63.5} & \underline{89.7} & \textbf{66.2} & 63.3 & \underline{69.2} \\
\rowcolor{gray!15}
\quad KeepLoRA++$_\text{cls}$ & \ding{55} & \checkmark & ~~-- & 86.1 & \textbf{69.9} & \textbf{44.4} & \textbf{54.5} & \textbf{71.3} & \textbf{89.1} & \textbf{64.8} & \textbf{90.8} & 63.6 & \textbf{66.4} & \textbf{70.1} \\
\hline
\multicolumn{15}{l}{\textbf{Average}}\\
\quad LwF {\cite{li2017learning}} & \checkmark & \ding{55} & 36.3 & 86.9 & 72.0 & 59.0 & 73.7 & 60.0 & 73.6 & 74.8 & 80.0 & 37.3 & 58.1 & 64.7 \\
\quad iCaRL {\cite{rebuffi2017icarl}} & \checkmark & \ding{55} & 35.5 & 89.2 & 72.2 & 60.6 & 68.8 & 70.0 & 78.2 & 62.3 & 81.8 & 41.2 & 62.5 & 65.7 \\
\quad LwF-VR {\cite{ding2022don}} & \checkmark & \ding{55} & 29.6 & 87.7 & 74.4 & 59.5 & 72.4 & 63.6 & 77.0 & 66.7 & 81.2 & 43.7 & 60.7 & 65.1 \\
\quad WiSE-FT {\cite{wortsman2022robust}} & \checkmark & \ding{55} & 26.7 & 86.5 & 64.3 & 57.1 & 65.7 & 58.7 & 71.1 & 70.5 & 75.8 & 36.9 & 54.6 & 60.7 \\
\quad ZSCL {\cite{zheng2023preventing}} & \checkmark & \ding{55} & 45.1 & 92.0 & 80.1 & 64.3 & 79.5 & \underline{81.6} & \underline{89.6} & 75.2 & 88.9 & \textbf{64.7} & \textbf{68.0} & 75.4 \\
\quad O-LoRA {\cite{wang2023orthogonal}} & \checkmark & \checkmark & 39.8 & 93.2 & 78.3 & 61.7 & 78.9 & 76.3 & 88.5 & 73.9 & 90.1 & 60.2 & 65.2 & 73.3 \\
\quad InfLoRA {\cite{liang2024inflora}} & \checkmark & \checkmark & \underline{53.6} & \underline{95.6} & \underline{82.8} & \underline{65.0} & \underline{80.9} & 79.6 & 89.1 & \underline{76.1} & \underline{90.2} & 62.3 & 64.5 & \underline{76.3} \\
\quad SD-LoRA {\cite{wu2025sd}} & \checkmark & \checkmark & 36.7 & 92.2 & 80.2 & 55.9 & 77.5 & 73.2 & 89.2 & 74.9 & 89.8 & 62.5 & 65.0 & 72.5 \\
\rowcolor{gray!15}
\quad KeepLoRA++ & \checkmark & \checkmark & \textbf{56.1} & \textbf{95.8} & \textbf{83.9} & \textbf{66.3} & \textbf{82.6} & \textbf{82.6} & \textbf{90.0} & \textbf{77.5} & \textbf{91.7} & \underline{64.6} & \underline{66.7} & \textbf{78.0} \\
\cmidrule{2-15}
\quad L2P {\cite{wang2022learning}} & \ding{55} & \checkmark & 38.0 & 85.2 & 78.2 & 61.3 & 72.9 & 74.9 & 79.7 & 59.1 & 82.0 & 59.7 & 55.4 & 67.9 \\
\quad DualPrompt {\cite{wang2022dualprompt}} & \ding{55} & \checkmark & 37.8 & 84.3 & 78.6 & 60.1 & 71.1 & 73.2 & 79.1 & 73.9 & 82.3 & 55.1 & 52.8 & 68.0 \\
\quad S-Prompts {\cite{wang2022s}} & \ding{55} & \checkmark & 37.5 & 92.5 & 77.5 & 58.2 & 76.4 & 74.1 & 78.8 & 57.9 & 83.0 & 60.8 & 54.4 & 68.3 \\
\quad DIKI {\cite{tang2024mind}} & \ding{55} & \checkmark & 45.4 & 95.7 & 83.0 & 65.0 & 78.2 & 82.5 & 87.1 & 71.7 & 90.0 & 67.2 & 66.6 & 75.7 \\
\quad MoE-Adapters {\cite{yu2024boosting}} & \ding{55} & \ding{55} & \underline{50.2} & 91.9 & 83.1 & \textbf{69.4} & 78.9 & 84.0 & \underline{89.1} & 73.7 & 89.3 & 67.7 & \underline{66.9} & 76.7 \\
\quad IAP {\cite{fu2026iap}} & \ding{55} & \checkmark & 45.9 & \underline{95.8} & \underline{83.3} & 66.5 & \underline{79.5} & \textbf{84.8} & 87.5 & \underline{76.6} & \underline{91.0} & \textbf{69.2} & 64.5 & \underline{76.8} \\
\rowcolor{gray!15}
\quad KeepLoRA++$_\text{cls}$ & \ding{55} & \checkmark & \textbf{58.4} & \textbf{96.6} & \textbf{84.5} & \underline{67.5} & \textbf{82.4} & \underline{84.7} & \textbf{90.9} & \textbf{77.4} & \textbf{92.0} & \underline{68.0} & \textbf{67.9} & \textbf{79.1} \\
\hline
\multicolumn{15}{l}{\textbf{Last}}\\
\quad LwF {\cite{li2017learning}} & \checkmark & \ding{55} & 26.3 & 87.5 & 71.9 & 66.6 & 79.9 & 66.9 & 83.8 & \textbf{99.6} & 92.1 & 66.1 & 80.4 & 74.6 \\
\quad iCaRL {\cite{rebuffi2017icarl}} & \checkmark & \ding{55} & 35.8 & 93.0 & 77.0 & 70.2 & 83.3 & 88.5 & 90.4 & 86.7 & 93.2 & 81.2 & \underline{81.9} & 80.1 \\
\quad LwF-VR {\cite{ding2022don}} & \checkmark & \ding{55} & 20.5 & 89.8 & 72.3 & 67.6 & 85.5 & 73.8 & 85.7 & \textbf{99.6} & 93.1 & 73.3 & 80.9 & 76.6 \\
\quad WiSE-FT {\cite{wortsman2022robust}} & \checkmark & \ding{55} & 27.2 & 90.8 & 68.0 & 68.9 & 86.9 & 74.0 & 87.6 & \textbf{99.6} & 92.6 & 77.8 & 81.3 & 77.7 \\
\quad ZSCL {\cite{zheng2023preventing}} & \checkmark & \ding{55} & 40.6 & 92.2 & 81.3 & 70.5 & 94.8 & \underline{90.5} & \underline{91.9} & 98.7 & 93.9 & \textbf{85.3} & 80.2 & 83.6 \\
\quad O-LoRA {\cite{wang2023orthogonal}} & \checkmark & \checkmark & 31.4 & 91.8 & 75.7 & 61.1 & 89.0 & 76.0 & 88.9 & 99.1 & 92.3 & 74.8 & 81.3 & 78.3 \\
\quad InfLoRA {\cite{liang2024inflora}} & \checkmark & \checkmark & \underline{51.1} & \underline{96.5} & \underline{85.1} & \underline{70.7} & \underline{98.1} & 87.7 & 91.3 & 99.4 & \underline{94.2} & 82.0 & 81.4 & \underline{85.2} \\
\quad SD-LoRA {\cite{wu2025sd}} & \checkmark & \checkmark & 31.1 & 92.3 & 79.8 & 57.4 & 88.7 & 76.1 & 90.6 & 99.0 & 92.9 & 81.3 & 81.6 & 79.2 \\
\rowcolor{gray!15}
\quad KeepLoRA++ & \checkmark & \checkmark & \textbf{54.0} & \textbf{96.6} & \textbf{86.5} & \textbf{72.5} & \textbf{98.3} & \textbf{91.4} & \textbf{92.1} & \underline{99.5} & \textbf{94.5} & \underline{84.6} & \textbf{82.2} & \textbf{86.6} \\
\cmidrule{2-15}
\quad L2P {\cite{wang2022learning}} & \ding{55} & \checkmark & 38.0 & 87.1 & 84.2 & 72.9 & 86.0 & 96.1 & 89.2 & 99.0 & 94.1 & 79.6 & 76.0 & 82.0 \\
\quad DualPrompt {\cite{wang2022dualprompt}} & \ding{55} & \checkmark & 37.8 & 87.1 & 84.6 & 71.8 & 89.2 & 96.3 & 89.1 & 99.1 & 94.5 & 79.9 & 76.5 & 82.3 \\
\quad S-Prompts {\cite{wang2022s}} & \ding{55} & \checkmark & 37.5 & 95.1 & 83.7 & 70.2 & 97.5 & 96.5 & 89.0 & 99.1 & 94.0 & 79.5 & 75.8 & 83.4 \\
\quad DIKI {\cite{tang2024mind}} & \ding{55} & \checkmark & 45.4 & 95.9 & 86.0 & 73.0 & 97.8 & \underline{96.8} & 89.3 & 99.3 & 94.4 & 81.8 & 76.4 & 85.1 \\
\quad MoE-Adapters {\cite{yu2024boosting}} & \ding{55} & \ding{55} & \underline{49.8} & 92.2 & 86.1 & \textbf{78.1} & 95.7 & 94.3 & 89.5 & 98.1 & 89.9 & 81.6 & \underline{80.0} & 85.0 \\
\quad IAP {\cite{fu2026iap}} & \ding{55} & \checkmark & 46.8 & \underline{96.1} & \underline{86.7} & 75.2 & \underline{98.1} & \textbf{97.0} & \underline{89.6} & \underline{99.4} & \underline{94.7} & \underline{82.8} & 76.7 & \underline{85.7} \\
\rowcolor{gray!15}
\quad KeepLoRA++$_\text{cls}$ & \ding{55} & \checkmark & \textbf{57.8} & \textbf{97.6} & \textbf{87.6} & \underline{76.3} & \textbf{98.2} & 95.8 & \textbf{92.9} & \textbf{99.5} & \textbf{95.0} & \textbf{87.7} & \textbf{83.1} & \textbf{88.3} \\
\bottomrule
\end{tabular}
\vspace{-0.1in}
\end{table*}

\subsection{Evaluation Metrics}

We define the \emph{Transfer}, \emph{Average}, \emph{PostAverage}, and \emph{Last} metrics to evaluate model performance under continual learning scenarios. Let $a_{t}^{(i)}$ represent the accuracy of the model on task $t$ after training on task $i$ with a total of $n$ tasks. The metrics for task $t$ are computed as follows:
\begin{equation}
\text{Transfer}_{t} = \frac{1}{t - 1} \sum_{i=1}^{t-1} a_{t}^{(i)}, \quad t = 2, 3, \dots, n,
\end{equation}
\begin{equation}
\text{Average}_{t} = \frac{1}{n} \sum_{i=1}^{n} a_{t}^{(i)}, \quad t = 1, 2, \dots, n,
\end{equation}
\begin{equation}
\text{PostAverage}_{t} = \frac{1}{n-t+1} \sum_{i=t}^{n} a_{t}^{(i)}, \quad t = 1, 2, \dots, n,
\end{equation}
\begin{equation}
\text{Last}_{t} = a_{t}^{(n)}, \quad t = 1, 2, \dots, n.
\end{equation}
The \emph{Transfer} metric evaluates forward stability by measuring the performance of unseen tasks throughout ($i+1, i+2, \dots, n$) after training on the task $i$. 
The \emph{PostAverage} metric captures the performance of a task from the moment it is learned until the end of the sequence, reflecting both learning plasticity and backward stability. 
The \emph{Last} metric measures the final performance on each task after completing all training steps, which primarily quantifies backward stability. 
Finally, the \emph{Average} metric represents the mean accuracy across all time steps to offer a holistic measure of overall stability and plasticity.

\subsection{Prototype-Based Classification Extension}
We further enhance classification performance with KeepLoRA++$_\text{cls}$\(_{\mathrm{cls}}\), a variant that adds a learnable prototype vector for each class. We clarify that the KeepLoRA+ variant \cite{luo2026keeplora} is identical to KeepLoRA$_{\mathrm{cls}}$. KeepLoRA++$_{\mathrm{cls}}$ combines KeepLoRA++ with the prototype-based classification extension used in KeepLoRA+.

Let $\boldsymbol{v}$ denote the visual feature extracted by the vision encoder, and $\boldsymbol{t}_c$ denote the textual feature extracted by the text encoder for the $c$-th class name, where $c \in \{1, 2, \dots, C\}$. We introduce a set of class-specific prototype vectors $\boldsymbol{P} = \{\boldsymbol{p}_1, \boldsymbol{p}_2, \dots, \boldsymbol{p}_C\}$. Prior to training on a new task, each prototype $\boldsymbol{p}_c$ is initialized using the mean visual feature of its corresponding training samples:
$
    \boldsymbol{p}_c = \frac{1}{N_c} \sum_{j=1}^{N_c} \boldsymbol{v}_{j}^{(c)},
$
where $N_c$ is the number of training images belonging to class $c$, and $\boldsymbol{v}_{j}^{(c)}$ is the visual feature of the $j$-th sample. During the training stage, these prototype vectors $\boldsymbol{P}$ are jointly optimized alongside the KeepLoRA++ adapter parameters. 

During inference, the prediction is determined by combining the standard vision-text contrastive logits and the vision-prototype similarity. Specifically, the combined logit $s_c$ for class $c$ is computed as the average of the two similarities:
$
    s_c = \frac{1}{2} \Big( \text{sim}(\boldsymbol{v}, \boldsymbol{t}_c) + \text{sim}(\boldsymbol{v}, \boldsymbol{p}_c) \Big),
$
where $\text{sim}(\cdot, \cdot)$ denotes the cosine similarity. %

\subsection{Main Results}
We evaluate our method on the dual-encoder model CLIP~\cite{radford2021learning} and encoder-decoder models LLaVA~\cite{liu2023visual} and Video-LLaVA~\cite{lin2024video}.

We evaluate CLIP on MTIL~\cite{zheng2023preventing} under alphabetical and randomly permuted task orders, reported in Table~\ref{tab:MTIL} and Table~\ref{tab:MTIL_order2}, respectively. Both orders are evaluated using KeepLoRA++ and its prototype-enhanced variant, KeepLoRA++$_\text{cls}$. 
In Table~\ref{tab:DCL_methods} and \ref{tab:UCIT_methods}, we evaluate LLaVA on MLLM-DCL~\cite{guo2025mcitlib} and UCIT~\cite{guo2025mcitlib} benchmarks, respectively. %
In Table~\ref{tab:CL_VISTA_methods}, we evaluate Video-LLaVA on the CL-VISTA~\cite{guo2026cl} benchmark. %
Detailed experimental settings are provided in Appendix~\ref{appendix:experiment details}. 
KeepLoRA++ and its variant achieve state-of-the-art performance on the \emph{Transfer}, \emph{PostAverage}, \emph{Last}, and \emph{Average} metrics across all these settings, demonstrating that our approach consistently addresses the challenges of forward stability, backward stability, and plasticity in continual learning.

\begin{table}[t!]
\centering
\footnotesize
\caption{\textnormal{\textbf{Comparison of different continual learning methods on MLLM-DCL benchmark for VQA tasks} in terms of \emph{Transfer}, \emph{Average}, and \emph{Last} scores (\%). The best results are in \textbf{bold}, and the second-best results are \underline{underlined}.}}
\label{tab:DCL_methods}
\begin{tabular}{b{7em} *{5}{b{2.1em}} >{\centering\arraybackslash}m{2.2em} @{\hspace{0.8em}}}
\toprule
\textbf{Method} & \rotatebox{45}{Sensing} & \rotatebox{45}{Medical~~} & \rotatebox{45}{Driving} & \rotatebox{45}{Science} & \rotatebox{45}{Finance} & \textbf{Avg.} \\
\midrule
\quad Zero-shot & 32.29 & 28.28 & 15.59 & 35.55 & 62.56 &  \\
\hline
\multicolumn{7}{l}{\textbf{Transfer}}\\
\quad LoRA-FT {\cite{hu2022lora}} & ~~-- & 28.10 & 17.44 & 34.03 & 50.19 & 32.44 \\
\quad O-LoRA {\cite{wang2023orthogonal}} & ~~-- & \underline{28.37} & 18.37 & 33.72 & \underline{52.53} & 33.25\\
\quad CL-MoE {\cite{huai2025clmoe}} & ~~-- & 28.25 & \underline{19.38} & \underline{34.08} & 48.56 & 32.57 \\
\quad RegLoRA {\cite{chen2025sefe}} & ~~-- & 28.10 & \textbf{19.63} & 33.85 & 52.36 & \underline{33.49} \\
\rowcolor{gray!15}
\quad KeepLoRA++ & ~~-- & \textbf{29.10} & 15.57 & \textbf{34.99} & \textbf{58.42} & \textbf{34.52} \\
\hline
\multicolumn{7}{l}{\textbf{Average}}\\
\quad LoRA-FT {\cite{hu2022lora}} & 73.34 & 44.94 & 31.38 & 38.79 & 57.84 & 49.26 \\
\quad O-LoRA {\cite{wang2023orthogonal}} & 75.04 & 45.71 & 32.62 & 38.54 & \underline{59.64} & 50.31 \\
\quad CL-MoE {\cite{huai2025clmoe}} & 74.19 & 45.60 & 32.08 & 38.88 & 56.68 & 49.49 \\
\quad RegLoRA {\cite{chen2025sefe}} & \underline{77.71} & \underline{47.69} & \underline{35.35} & \underline{38.99} & 59.57 & \underline{51.86} \\
\rowcolor{gray!15}
\quad KeepLoRA++ & \textbf{79.31} & \textbf{52.77} & \textbf{37.88} & \textbf{41.10} & \textbf{64.63} & \textbf{55.14} \\
\hline
\multicolumn{7}{l}{\textbf{Last}}\\
\quad LoRA-FT {\cite{hu2022lora}} & 69.34 & 44.30 & 29.10 & 41.44 & 88.43 & 54.52 \\
\quad O-LoRA {\cite{wang2023orthogonal}} & 72.30 & 46.89 & 31.59 & 41.50 & 88.06 & 56.07\\
\quad CL-MoE {\cite{huai2025clmoe}} & 71.83 & 47.36 & 29.49 & 41.48 & \underline{89.16} & 55.86 \\
\quad RegLoRA {\cite{chen2025sefe}} & \underline{77.05} & \underline{50.86} & \underline{40.27} & \underline{42.98} & 88.40 & \underline{59.91} \\
\rowcolor{gray!15}
\quad KeepLoRA++ & \textbf{79.76} & \textbf{57.65} & \textbf{51.86} & \textbf{50.01} & \textbf{89.49} & \textbf{65.75} \\
\bottomrule
\end{tabular}
\vspace{-0.1in}
\end{table}

\begin{table}[t!]
\centering
\caption{\textnormal{\textbf{Comparison of different continual learning methods on UCIT benchmark for VQA tasks} in terms of \emph{Transfer}, \emph{Average}, and \emph{Last} scores (\%). The best results are in \textbf{bold}, and the second-best results are \underline{underlined}.}}
\label{tab:UCIT_methods}
\begin{tabular}{b{7.0em} *{6}{b{1.6em}} >{\centering\arraybackslash}m{2.2em} @{\hspace{0.8em}}}
\toprule
\textbf{Method} & \rotatebox{45}{ImgNet-R} & \rotatebox{45}{ArxivQA} & \rotatebox{45}{VizWiz} & \rotatebox{45}{IconQA} & \rotatebox{45}{CLEVR} & \rotatebox{45}{Flickr30k} & \textbf{Avg.} \\
\midrule
\quad Zero-shot & 16.27 & 53.73 & 38.39 & 19.20 & 20.63 & 41.88 &  \\
\hline
\multicolumn{8}{l}{\textbf{Transfer}}\\
\quad LoRA-FT {\cite{hu2022lora}} & ~~-- & 52.63 & 18.30 & 6.02 & 16.97 & 40.29 & 26.84 \\
\quad O-LoRA {\cite{wang2023orthogonal}} & ~~-- & 52.87 & \textbf{19.57} & 4.42 & 16.85 & \underline{41.04} & 26.95\\
\quad CL-MoE {\cite{huai2025clmoe}} & ~~-- & 52.00 & 19.32 & 7.37 & \textbf{17.81} & \textbf{41.28} & \underline{27.56} \\
\quad RegLoRA {\cite{chen2025sefe}} & ~~-- & \textbf{53.33} & 18.68 & \underline{7.48} & \underline{17.03} & 40.90 & 27.48 \\
\rowcolor{gray!15}
\quad KeepLoRA++ & ~~-- & \underline{53.27} & \underline{19.50} & \textbf{12.15} & 16.89 & 40.20 & \textbf{28.40} \\
\hline
\multicolumn{8}{l}{\textbf{Average}}\\
\quad LoRA-FT {\cite{hu2022lora}} & 75.98 & 77.78 & 41.56 & 38.83 & 34.56 & 43.25 & 51.99 \\
\quad O-LoRA {\cite{wang2023orthogonal}} & 82.43 & \underline{80.06} & 41.73 & 35.87 & 33.94 & \underline{43.74} & 52.96 \\
\quad CL-MoE {\cite{huai2025clmoe}} & 80.16 & 77.10 & 40.43 & 30.33 & 33.10 & \textbf{43.95} & 50.85 \\
\quad RegLoRA {\cite{chen2025sefe}} & \underline{85.49} & 78.55 & \textbf{42.92} & \underline{40.33} & \underline{34.80} & 43.64 & \underline{54.29} \\
\rowcolor{gray!15}
\quad KeepLoRA++ & \textbf{86.07} & \textbf{86.64} & \underline{42.02} & \textbf{41.26} & \textbf{35.37} & 42.95 & \textbf{55.72} \\
\hline
\multicolumn{8}{l}{\textbf{Last}}\\
\quad LoRA-FT {\cite{hu2022lora}} & 58.60 & 76.73 & 45.72 & \underline{67.43} & 61.57 & \textbf{58.03} & 61.35 \\
\quad O-LoRA {\cite{wang2023orthogonal}} & 74.17 & \underline{80.93} & 45.30 & 62.87 & 63.83 & 57.24 & 64.06 \\
\quad CL-MoE {\cite{huai2025clmoe}} & 67.17 & 75.77 & 44.38 & 52.63 & 54.40 & 57.28 & 58.61 \\
\quad RegLoRA {\cite{chen2025sefe}} & \underline{80.23} & 79.13 & \textbf{47.11} & \textbf{69.40} & \underline{65.70} & \underline{57.33} & \underline{66.48} \\
\rowcolor{gray!15}
\quad KeepLoRA++ & \textbf{81.20} & \textbf{94.40} & \underline{46.14} & 66.50 & \textbf{69.63} & 56.62 & \textbf{69.08} \\
\bottomrule
\end{tabular}
\vspace{-0.1in}
\end{table}

\begin{table*}[t!]
\centering
\caption{\textnormal{\textbf{Comparison of different continual learning methods on CL-VISTA benchmark for video understanding tasks} in terms of \emph{Transfer}, \emph{PostAverage}, and \emph{Last} scores (\%). The best results are in \textbf{bold}, and the second-best results are \underline{underlined}.}}
\label{tab:CL_VISTA_methods}
\begin{tabular}{b{8em}b{0.8em}b{0.8em} *{8}{b{2.5em}} >{\centering\arraybackslash}m{2.5em} @{\hspace{0.8em}}}
\toprule
\textbf{Method} & \rotatebox{45}{Arch. Kept} & \rotatebox{45}{w/o Extra Data} & \rotatebox{45}{Count.} & \rotatebox{45}{Space} & \rotatebox{45}{Traffic} & \rotatebox{45}{Movie} & \rotatebox{45}{GUI} & \rotatebox{45}{Science} & \rotatebox{45}{Sports} & \rotatebox{45}{Reason.} & \textbf{Avg.} \\
\midrule
\quad Zero-shot  & \checkmark & \checkmark  & 32.24 & 37.24 & 41.42 & 22.78 & 60.17 & 56.91 & 34.40 & 37.52 &  \\
\hline
\multicolumn{8}{l}{\textbf{Transfer}}\\
\quad LoRA-FT {\cite{hu2022lora}} & \checkmark & \checkmark & --& 41.75 &	50.42 &	36.10 &	63.28 &	61.20 &	\textbf{52.37} &	52.67 &	51.11  \\
\rowcolor{gray!15}
\quad KeepLoRA++  & \checkmark & \checkmark & --& \textbf{42.32} &	\textbf{51.46} &	\textbf{37.56} &	\textbf{64.97} &	\textbf{61.60} &	52.14 &	\textbf{55.97} &	\textbf{52.29}  \\
\hline
\multicolumn{8}{l}{\textbf{PostAverage}}\\
\quad LoRA-FT {\cite{hu2022lora}} & \checkmark & \checkmark & 46.36 &	57.68 &	59.02 &	80.79 &	72.74 &	\underline{81.22} &	\underline{89.33} &	\underline{86.86} &	71.75 \\
\quad O-LoRA {\cite{wang2023orthogonal}} & \checkmark & \checkmark & 51.80 &	60.46 &	48.58 &	64.56 &	59.51 &	63.56 &	67.44 	& 46.97 &	57.86  \\
\quad RegLoRA {\cite{chen2025sefe}} & \checkmark & \checkmark & 41.12 &	46.64 &	50.34 &	45.23 &	65.02 &	68.62 &	69.54 & 	86.51 &	59.13 \\
\quad HiDe{\cite{guo2025hide}}  & \ding{55} & \checkmark & 52.51 &	57.43 &	48.01 &	72.59 &	64.46 &	67.34 &	68.71 &	51.26 &	60.29  \\
\quad CL-MoE {\cite{huai2025clmoe}}  & \ding{55} & \checkmark & 47.67 &	58.37 &	57.05 &	79.47 &	72.36 &	80.76 &	88.75 &	86.58 &	71.38  \\
\quad SMoLoRA{\cite{wang2025smolora}}  & \ding{55} & \checkmark & \underline{55.57} &	\underline{62.46} &	\underline{62.03} &	\underline{83.88} &	\underline{74.94} &	\textbf{82.66} & 	88.13 &	\textbf{86.92} &	\underline{74.57} 
 \\
\rowcolor{gray!15}
\quad KeepLoRA++  & \checkmark & \checkmark & \textbf{57.09} &	\textbf{65.01} &	\textbf{67.76} &	\textbf{84.66} &	\textbf{76.84} &	81.19 &	\textbf{89.40} &	86.75 &	\textbf{76.09}  \\
\hline
\multicolumn{8}{l}{\textbf{Last}}\\
\quad LoRA-FT{\cite{hu2022lora}} & \checkmark & \checkmark  & 41.13 &	54.81 &	55.84 &	74.61 &	68.80 &	78.20 	& \underline{88.90} &	\underline{86.86} &	68.64   \\
\quad O-LoRA {\cite{wang2023orthogonal}} & \checkmark & \checkmark &  41.05 &	52.53 &	41.00 &	58.82 &	57.94 &	62.37 &	67.46 &	46.97 &	53.52  \\
\quad RegLoRA {\cite{chen2025sefe}}  & \checkmark & \checkmark  & 45.20 &	43.47 &	46.89 &	37.23 &	59.17 &	60.36 &	52.92 &	86.51 &	53.97  \\
\quad HiDe{\cite{guo2025hide}}  & \ding{55} & \checkmark & 41.35 &	47.32 &	39.01 &	64.46 &	60.29 &	64.36 &	67.46 &	51.26 &	54.44   \\
\quad CL-MoE {\cite{huai2025clmoe}}  & \ding{55} & \checkmark & 42.79 &	55.77 &	54.15 &	74.28 &	67.65 &	78.06 &	87.79 &	86.58 &	68.38   \\
\quad SMoLoRA{\cite{wang2025smolora}}  & \ding{55} & \checkmark &  \underline{51.66} &	\underline{59.45} &	\underline{57.70} &	\underline{80.64} &	\underline{71.17} &	\textbf{81.87} &	87.31 &	\textbf{86.92} &	\underline{72.09} \\
\rowcolor{gray!15}
\quad KeepLoRA++  & \checkmark & \checkmark & \textbf{56.83} &	\textbf{66.34} &	\textbf{65.71} &	\textbf{82.18} &	\textbf{75.35} &	\underline{79.93} &	\textbf{89.13} &	86.75 &	\textbf{75.28}  \\
\bottomrule
\end{tabular}
\vspace{-0.1in}
\end{table*}

\subsection{Analysis of Model Plasticity}

\begin{figure}[!t]
    \centering
    \subfloat[\# Param. 0.49 M]{
        \includegraphics[width=0.46\linewidth]{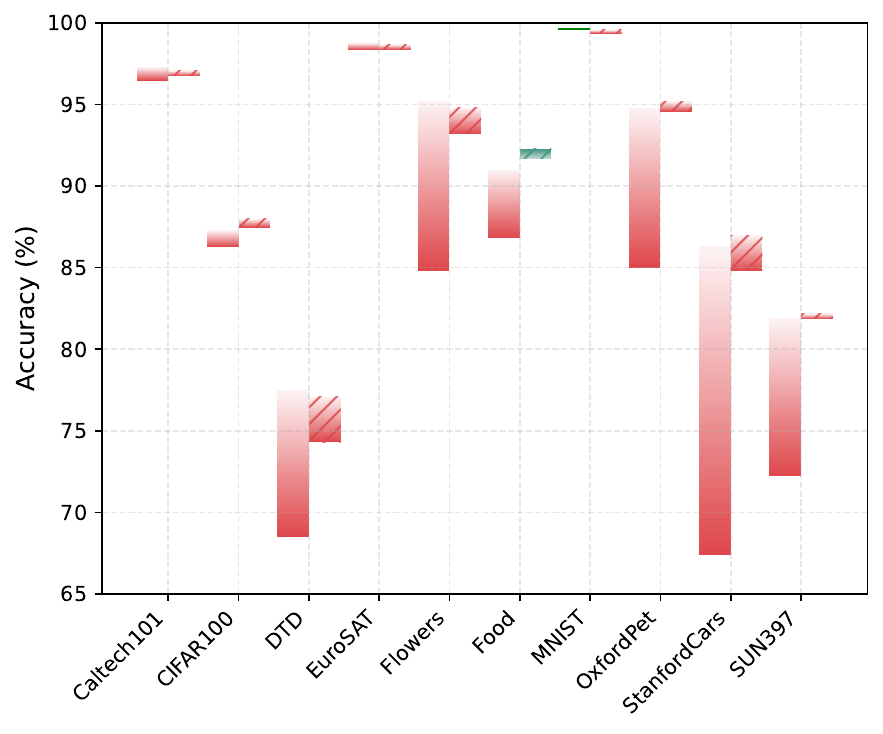}%
        \label{fig:sub_aa}
    }
    \hfill
    \subfloat[\# Param. 0.98 M]{
        \includegraphics[width=0.46\linewidth]{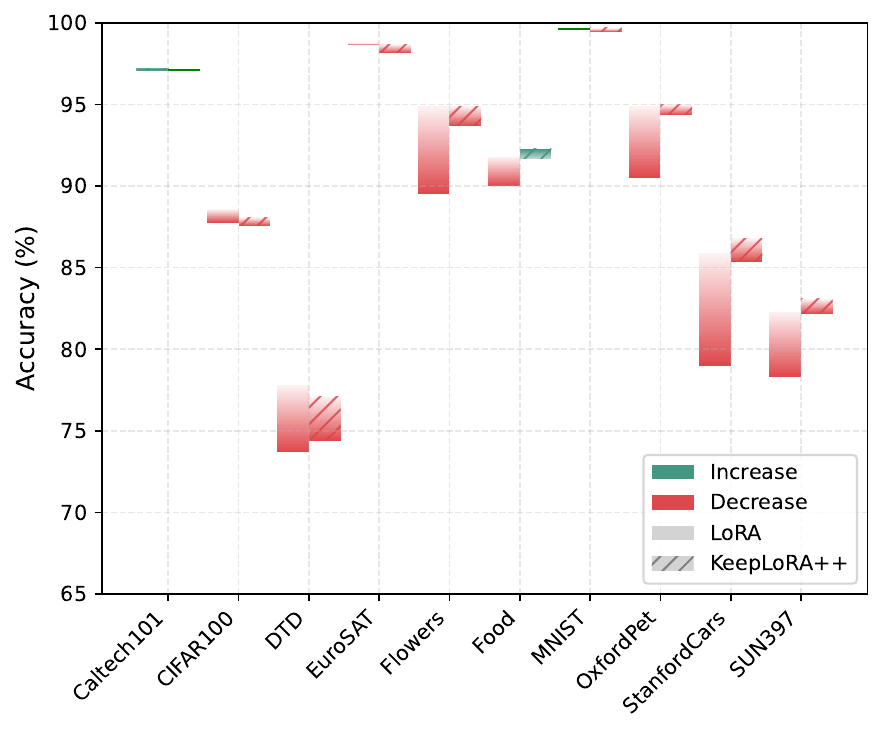}%
        \label{fig:sub_bb}
    }
    \caption{Comparison of plasticity between KeepLoRA++ and LoRA baseline under the same learnable parameter budgets: Fig.~\ref{fig:sub_aa} 0.49 million parameters and Fig.~\ref{fig:sub_bb} 0.98 million parameters. Each bar represents the performance drop for a task, measured as the difference between accuracy from isolated training and accuracy after sequential learning and immediate testing.}
    \label{fig:plasticity}
\end{figure}

Plasticity assesses the ability to effectively acquire new knowledge following a sequence of continual learning tasks. We evaluate two performance metrics for each task: (i) the accuracy achieved by training on the task in isolation, serving as an upper bound, and (ii) the accuracy measured immediately after the task is learned within the continual sequence. Our analysis in Fig.~\ref{fig:plasticity} compares KeepLoRA++ with a standard LoRA baseline. In the isolation-task setting, KeepLoRA++ performs comparably to LoRA, as gradient-informed initialization of the frozen down-projection matrix $\boldsymbol{A}$ effectively captures the essential learning direction, maintaining high learning capacity. Furthermore, when switching to the continual learning scenario, KeepLoRA++ exhibits a consistently smaller performance drop on new tasks compared to LoRA. %

\subsection{Analysis of Model Stability}
\begin{figure*}[!t]
    \centering
    \subfloat[LoRA]{
        \includegraphics[width=0.22\textwidth]{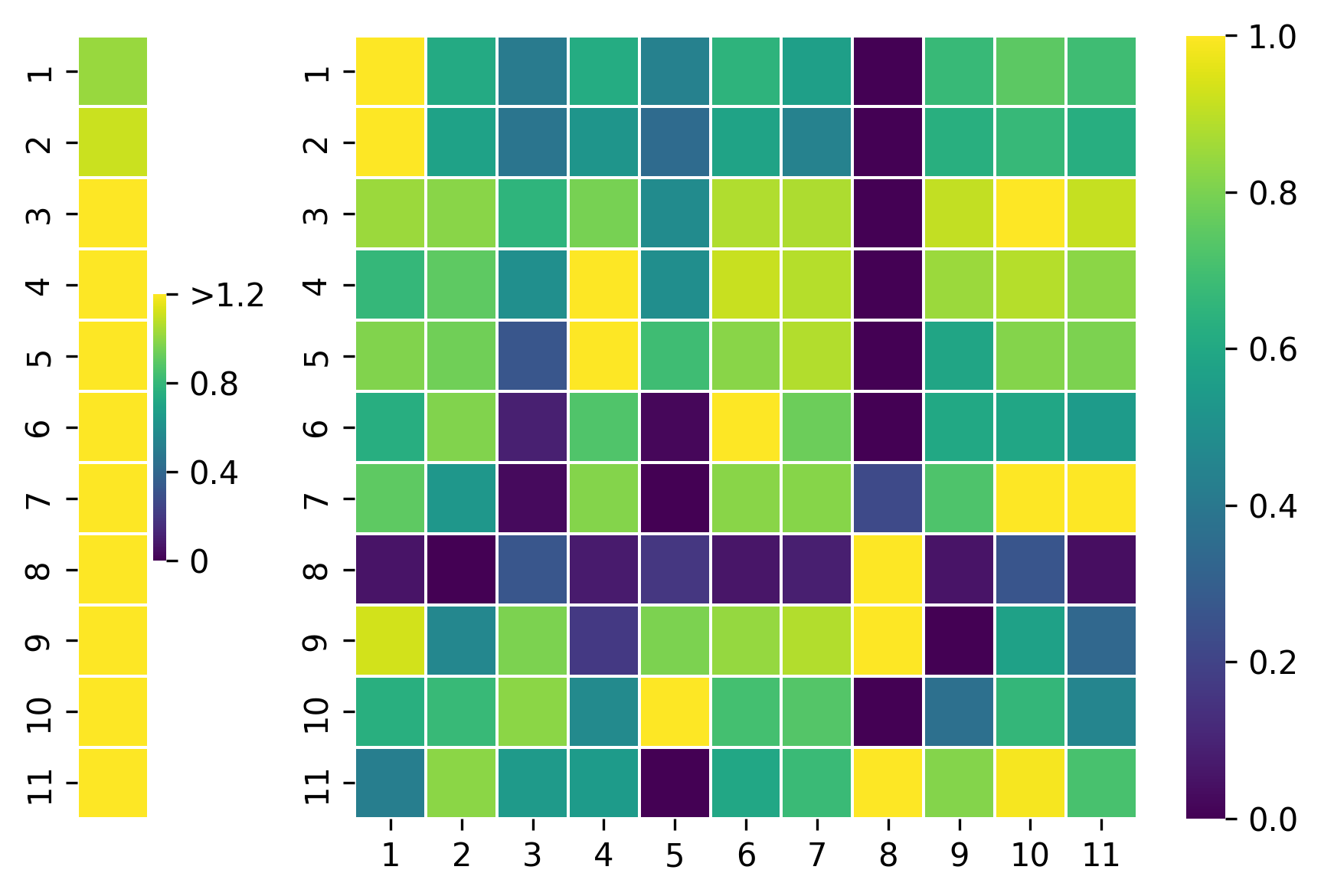}%
        \label{fig:sub_3_1}
    }
    \hfill
    \subfloat[LoRA frozen $\boldsymbol{A}$]{
        \includegraphics[width=0.22\textwidth]{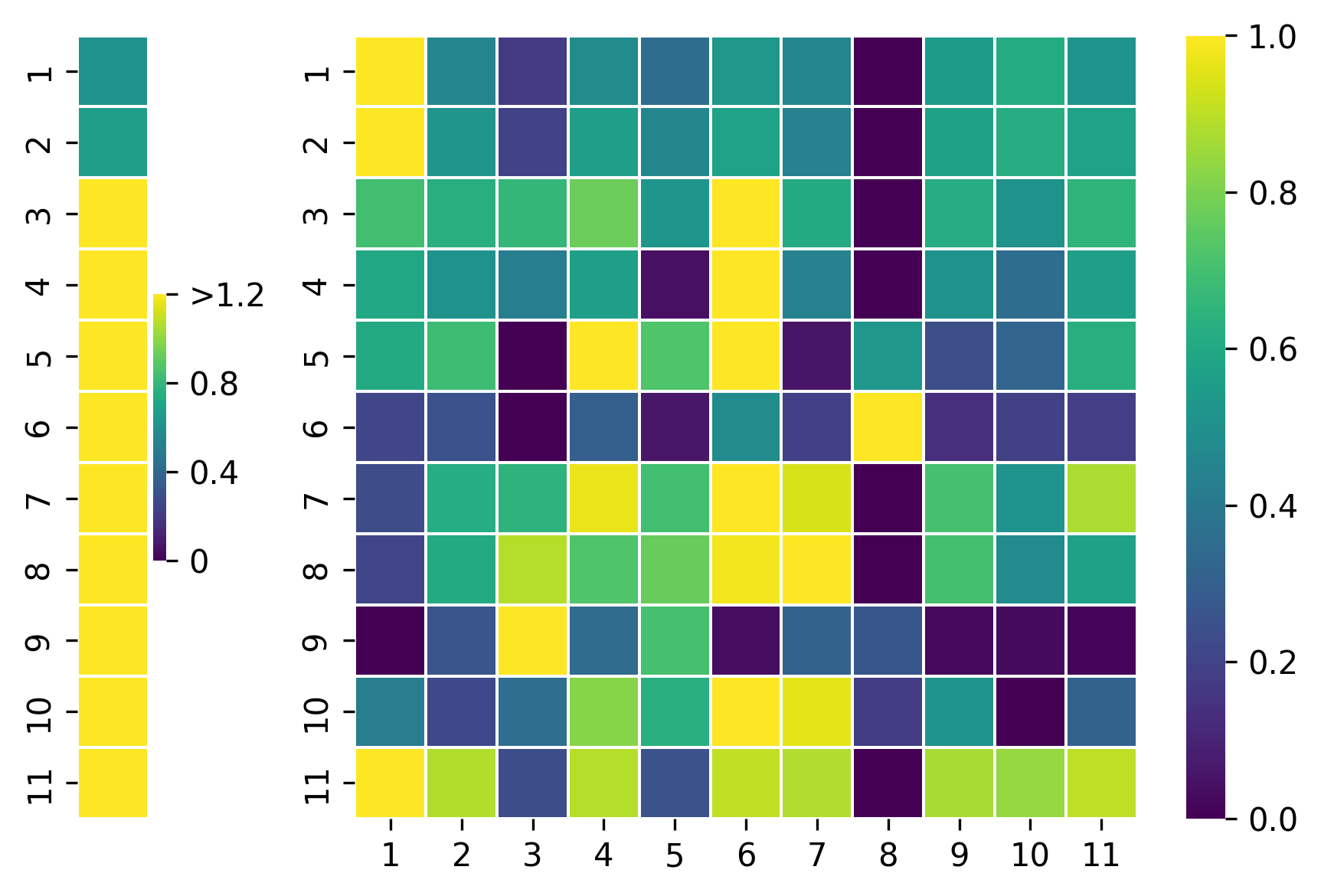}%
        \label{fig:sub_3_2}
    }
    \hfill
    \subfloat[LoRA frozen grad $\boldsymbol{A}$]{
        \includegraphics[width=0.22\textwidth]{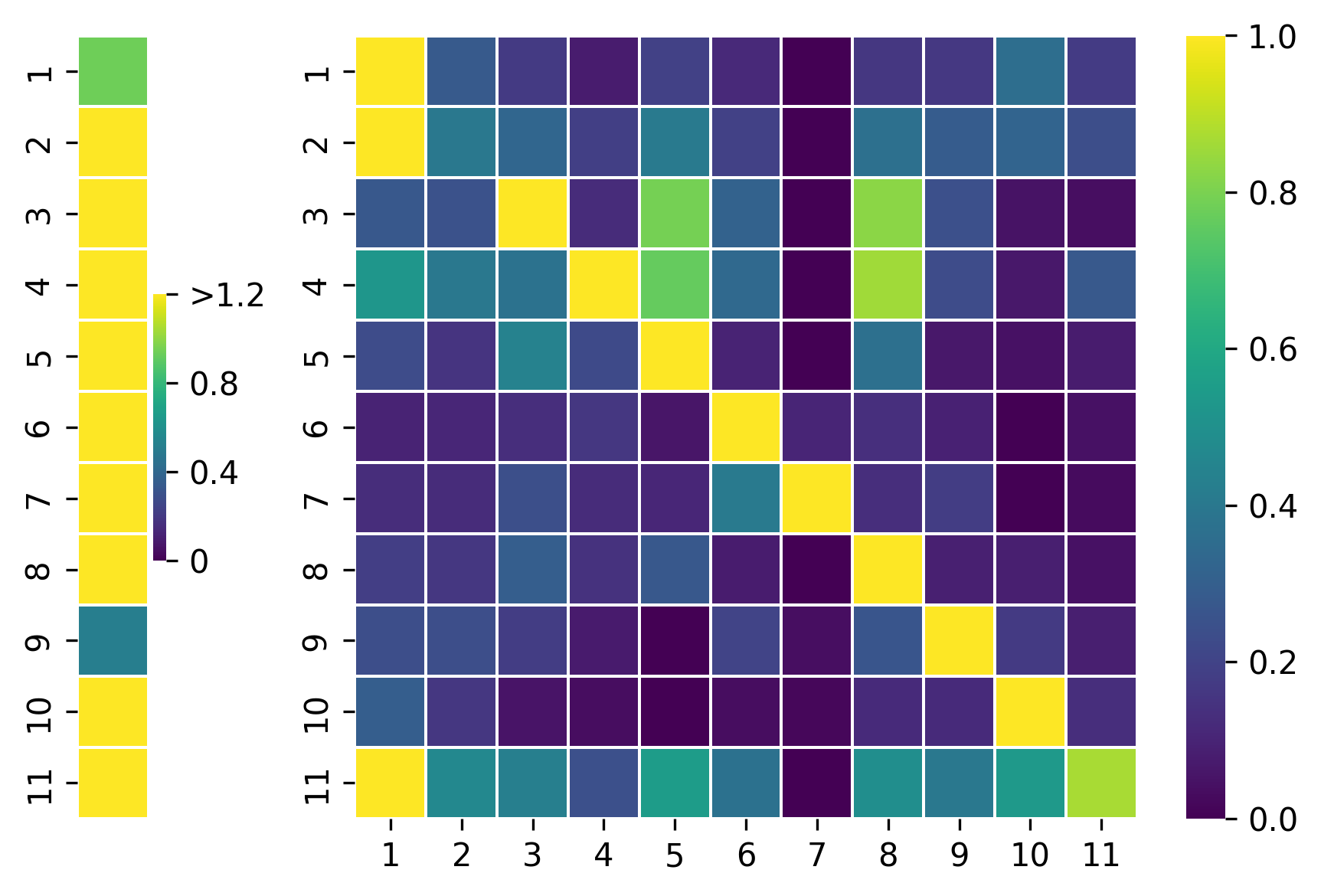}%
        \label{fig:sub_3_3}
    }
    \hfill
    \subfloat[KeepLoRA++]{
        \includegraphics[width=0.22\textwidth]{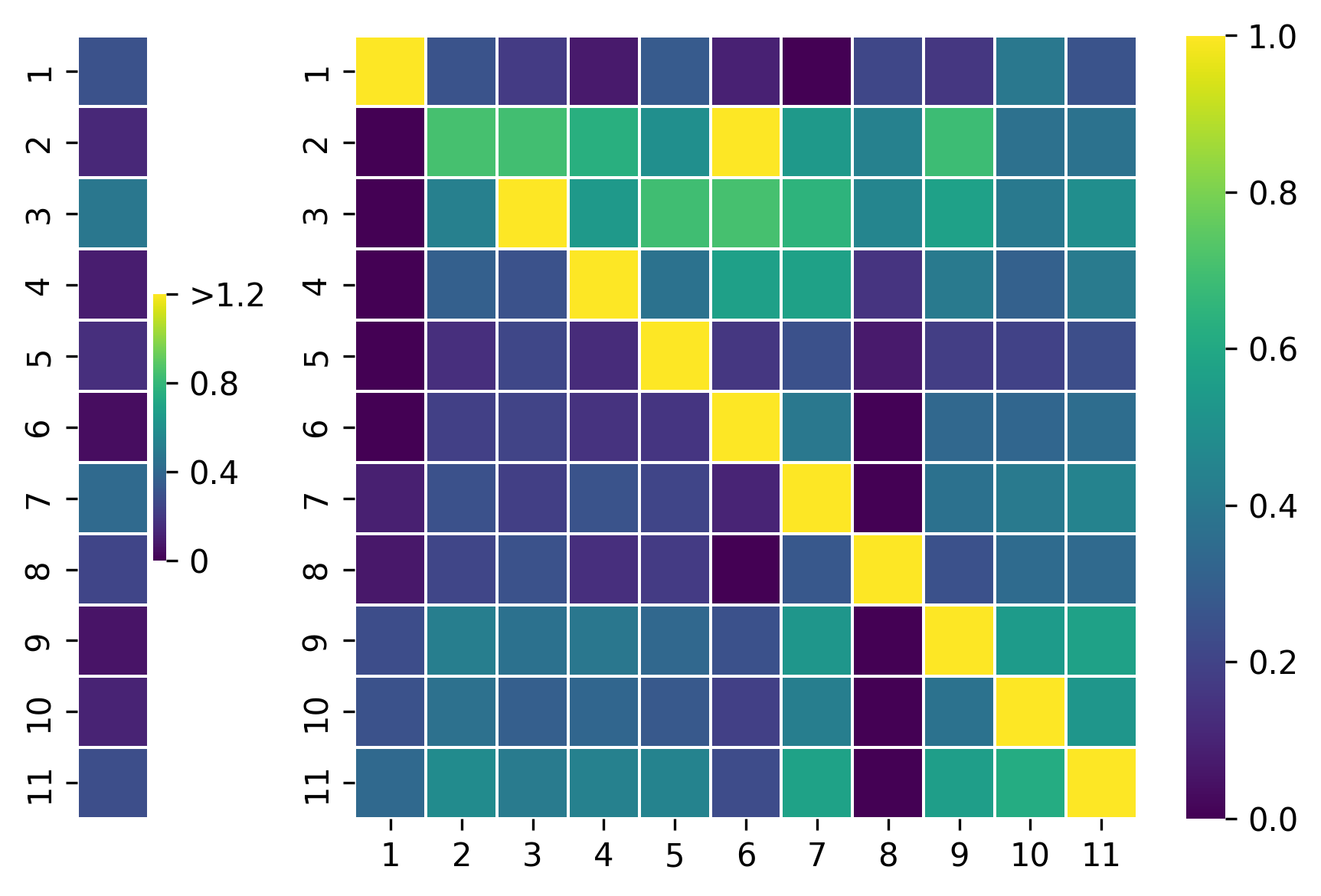}%
        \label{fig:sub_3_4}
    }
    \caption{Visualization of the average L2 norm of the output magnitude from the learned LoRA across multiple tasks. Each heatmap cell at row $i$ and column $j$ displays the normalized average L2 norm of the LoRA’s output when the model, trained up to task $i$, is tested on task $j$’s data. The vertical bar to the left of each heatmap indicates the mean output norm across all test tasks after each training stage, with darker colors signifying a lower norm. }%
    \label{fig:Interference}
\end{figure*}

We analyze stability by visualizing the interference of the LoRA module between multiple tasks in Fig.~\ref{fig:Interference}. In these heatmaps, the off-diagonal cells represent inter-task interference, while the vertical bar on the left indicates the overall impact on the backbone.
The standard LoRA (Fig.~\ref{fig:sub_3_1}) and LoRA with a frozen matrix $\boldsymbol{A}$ (Fig.~\ref{fig:sub_3_2}) both exhibit significant interference. The bright patterns in their heatmaps and vertical bars show that training on a current task heavily interferes with the representations of other tasks, leading to poor stability. Although gradient-informed initialization (Fig.~\ref{fig:sub_3_3}) reduces off-diagonal interference, the overall impact on the backbone remains high, as shown by its bright vertical bar.
In contrast, KeepLoRA++ (Fig. \ref{fig:sub_3_4}) shows a desirable pattern: a bright diagonal with dark off-diagonal cells. It indicates that the updates of the model focus on the current task, causing minimal interference with others. The dark vertical bar further confirms that the overall impact on the backbone is consistently low. By minimizing interference with previously learned tasks, KeepLoRA++ ensures backward stability. Furthermore, its minimal interference with unseen tasks, as indicated by the low norm, is critical to preserve forward stability.

\subsection{Ablation Study}

To analyze the contribution of each component, we conduct an ablation study starting from a standard LoRA baseline. Due to the high computational costs of training generative vision-language models, we perform the step-by-step analysis on the MTIL benchmark and evaluate key milestones on MLLM-DCL and UCIT. 

As shown in Table~\ref{tab:Ablation}, KeepLoRA++ achieves substantial improvements over the standard LoRA baseline across all settings. On the MTIL benchmark, KeepLoRA++ boosts the \emph{Transfer} metric by $11.2\%$, the \emph{Average} metric by $16.5\%$, and the \emph{Last} metric by $27.2\%$. These gains are driven by our dual-dimensional design. At the intra-layer parameter space level, constraining parameter updates to the residual subspace via orthogonal projection onto both the principal subspace $\boldsymbol{W}_p$ and previous task directions $\boldsymbol{M}_{t-1}$ protects transferable knowledge and prevents task interference. At the inter-layer level, applying the shallow-to-deep scaling schedule to transition from KeepLoRA \cite{luo2026keeplora} to KeepLoRA++ yields an additional $0.5\%$ improvement across all three metrics. The inter-layer scaling prevents disruption to the representations in shallow layers while exploiting the plasticity of deeper layers.

The benefits of this design also extend to generative multimodal tasks. On the MLLM-DCL benchmark, KeepLoRA++ improves \emph{Transfer}, \emph{Average}, and \emph{Last} over standard LoRA by $2.1$, $5.9$, and $11.2$ percentage points, respectively. Similarly, on the UCIT benchmark, KeepLoRA++ improves \emph{Transfer} and \emph{Last} by $1.6\%$ and $7.7\%$, respectively. Compared with KeepLoRA, KeepLoRA++ further improves \emph{Average} and \emph{Last} on both benchmarks, while achieving the same \emph{Transfer} score on UCIT at the reported precision. Although the improvements are not uniform across individual tasks, as detailed in Table~\ref{tab:methods_ucit}, the overall results demonstrate that jointly restricting parameter updates to the residual subspace and scaling their magnitudes across Transformer depths effectively balances plasticity, backward stability, and forward stability.

\begin{table*}[!t]
    \centering
    \caption{\textnormal{\textbf{Ablation Study of KeepLoRA++ on MTIL, MLLM-DCL and UCIT (\%).} For MTIL ablations, standard LoRA uses $r=8$, whereas frozen-$\boldsymbol{A}$ methods use $r=16$ to maintain the same number of trainable parameters. The best results are in \textbf{bold}, and the second-best results are \underline{underlined}.}}
    \label{tab:Ablation}
    \begin{tabular}{l|ccc|ccc|ccc} 
    \toprule
        \multirow{2}{*}{\textbf{Training Strategy}} & \multicolumn{3}{c|}{\textbf{MTIL}} & \multicolumn{3}{c|}{\textbf{MLLM-DCL}} & \multicolumn{3}{c}{\textbf{UCIT}} \\
        \cmidrule(lr){2-4} \cmidrule(lr){5-7} \cmidrule(lr){8-10}
        & Transfer & Average & Last & Transfer & Average & Last & Transfer & Average & Last \\
        \midrule
        LoRA  & 58.3 & 61.5 & 59.4 & 32.4 & 49.3 & 54.5 & \underline{26.8} & 52.0 & 61.4\\
        LoRA frozen $\boldsymbol{A}$  & 63.9 & 68.2  & 69.5 & -- & -- & -- & -- & -- & -- \\
        \quad (i)~~ Replace Eq. \ref{eq5} by $\hat{\boldsymbol{G}}_{t} = \boldsymbol{G}_{t}$ & 65.0  & 70.2  & 75.4 & -- & -- & -- & -- & -- & -- \\
        \quad (ii)~ Replace Eq. \ref{eq5} by $\hat{\boldsymbol{G}}_{t} = \boldsymbol{G}_{t} - \boldsymbol{W}_p \boldsymbol{W}_p^{\top}\boldsymbol{G}_{t}$ & 65.9  & 71.5  & 76.5  & -- & -- & -- & -- & -- & --\\
        \quad (iii) Replace Eq. \ref{eq5} by $\hat{\boldsymbol{G}}_{t} = \boldsymbol{G}_{t} - \boldsymbol{M}_{t-1}\boldsymbol{M}_{t-1}^{\top}\boldsymbol{G}_{t}$ & 68.1  & 77.2  & \underline{86.1}  & -- & -- & -- & -- & -- & --\\
        KeepLoRA (Eq. \ref{eq5}) & \underline{69.0}  & \underline{77.5}  & \underline{86.1} & \underline{33.7} & \underline{54.2} & \underline{64.4} & \textbf{28.4} & \underline{55.4} & \underline{67.8} \\
        KeepLoRA++ (Eq. \ref{eq5} and Eq.  \ref{eq:scale}) & \textbf{69.5} & \textbf{78.0} & \textbf{86.6} & \textbf{34.5} & \textbf{55.1} & \textbf{65.8} & \textbf{28.4} & \textbf{55.7} & \textbf{69.1} \\
    \bottomrule
    \end{tabular}
\end{table*}

\subsection{Computational Efficiency}
\begin{figure}[!t]
    \centering
    \subfloat[Training Efficiency]{
        \includegraphics[width=0.46\linewidth]{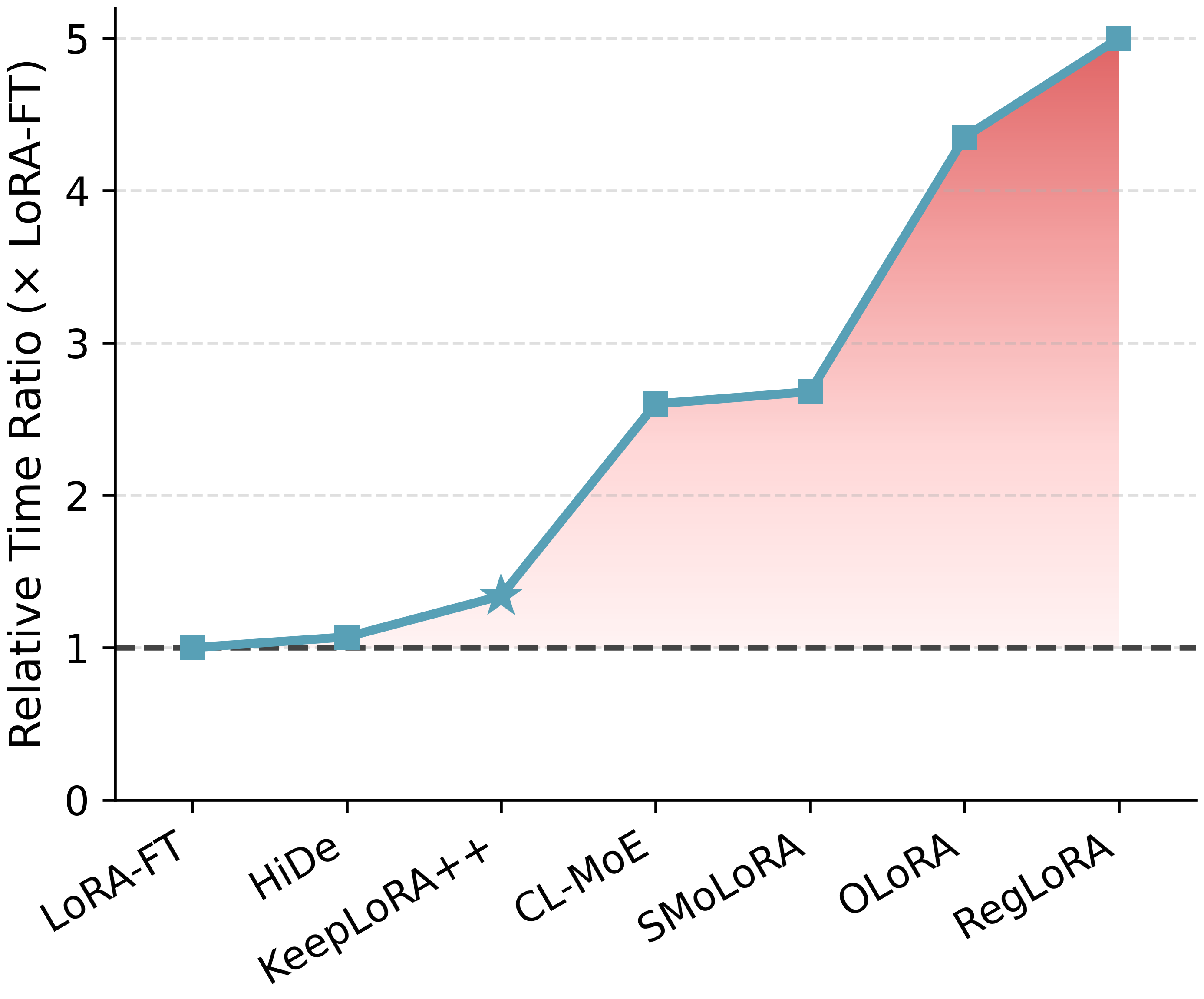}%
        \label{fig:sub_5a}
    }
    \hfill
    \subfloat[Inference Efficiency]{
        \includegraphics[width=0.46\linewidth]{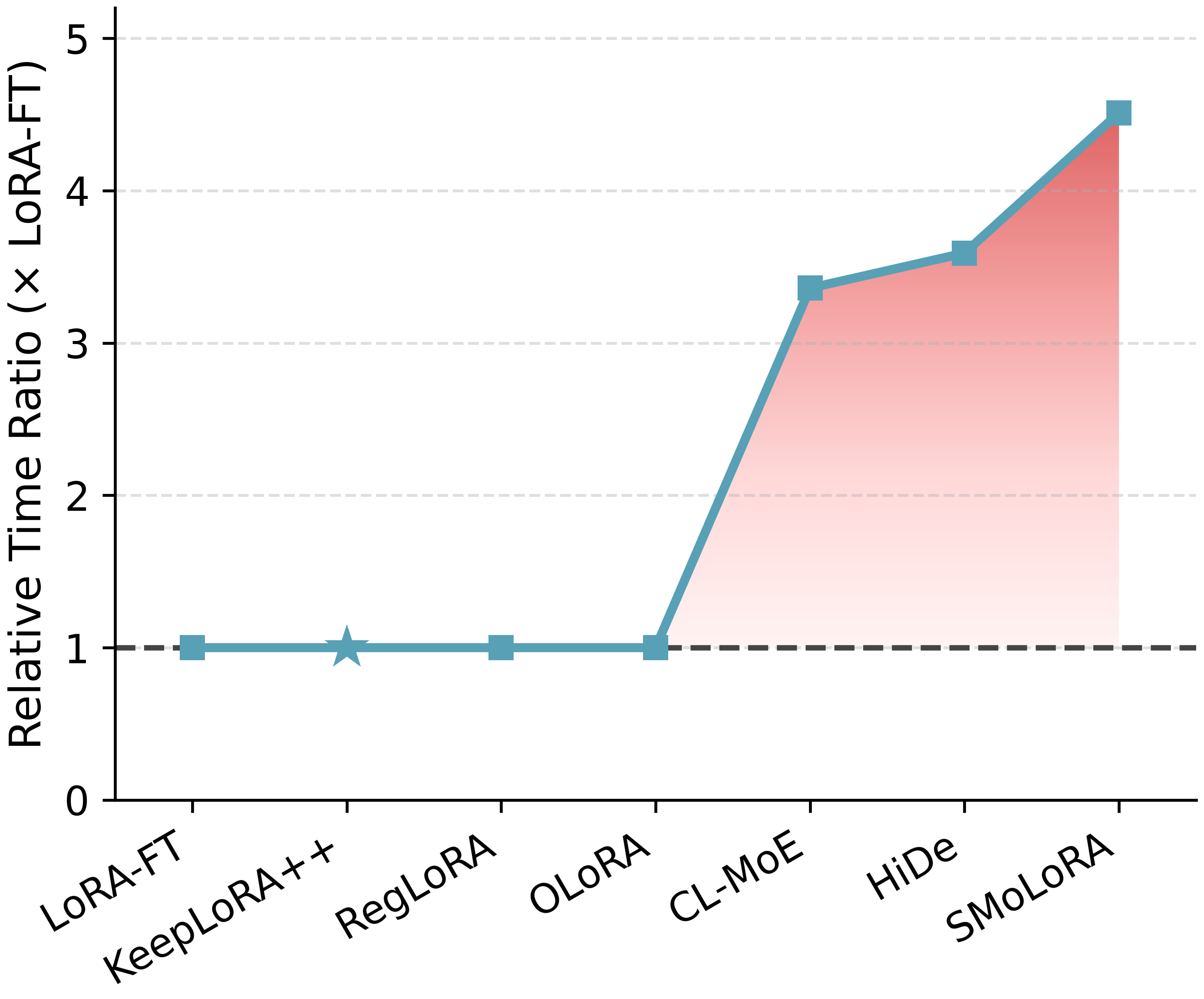}%
        \label{fig:sub_5b}
    }
    \caption {Comparison of computational cost on CL-VISTA benchmark. The vertical axis is the relative time cost normalized to the standard LoRA baseline indicated by the dashed line. KeepLoRA++ marked with a star introduces a marginal training overhead of 1.34 times the baseline while maintaining zero inference overhead.}
    \label{fig:efficiency}
\end{figure}

Beyond predictive accuracy and stability, computational efficiency is a critical factor for practical deployment. We benchmark the relative time costs of KeepLoRA++ on the CL-VISTA benchmark normalized to the standard LoRA baseline in Fig.~\ref{fig:efficiency}. During the training phase, the only additional computational overhead of our method stems from using a subset of data to extract the gradient for initializing matrix A before training a new task, and extracting the dominant gradient directions to update the stored basis after training. These operations are computationally lightweight, resulting in a marginal training time increase to 1.34 times that of the standard LoRA. During the inference phase, since KeepLoRA++ is an architecture kept method, the learned low-rank modules can be directly merged into the pre-trained backbone post-training. This structural property guarantees that our method introduces strictly zero additional inference latency. Consequently, KeepLoRA++ achieves optimal continual learning performance with minimal training time increase and no inference overhead.

\section{Conclusion}
In this paper, we investigate continual learning for pre-trained vision-language models from both intra-layer and inter-layer perspectives. Our analyses reveal that general transferable knowledge is mainly encoded in the principal subspaces and shallow layers, whereas residual subspaces and deeper layers provide greater flexibility for task-specific adaptation. Based on these findings, we propose KeepLoRA++, which integrates residual gradient adaptation with shallow-to-deep scaling. Theoretical analyses provide support for the proposed scaling strategy and simplified initialization, while extensive experiments across image classification, visual question answering, and video understanding tasks demonstrate the effectiveness of KeepLoRA++ in balancing forward stability, backward stability, and plasticity. We hope that this work will encourage further exploration of the structural properties of Transformers in the design of continual learning methods.

\bibliography{reference}
\bibliographystyle{IEEEtran}

\vfill

\newpage
\clearpage
\appendix

\subsection{Experiment Details}
\label{appendix:experiment details}

\subsubsection{Image Classification with CLIP}
We adopt the CLIP~\cite{radford2021learning} model with a ViT-B/16~\cite{dosovitskiy2021an} image encoder. The training process is carried out using the AdamW~\cite{loshchilov2018decoupled} optimizer, with a learning rate of $10^{-3}$ and a batch size of $64$ across all tasks with no more than $10$ epochs. For the primary experiments, we set the hyperparameters to $\epsilon_{w(\text{vision})}=0.85$ and $\epsilon_{w(\text{text})}=0.2$ for the vision and text encoders, respectively. The subspace constraint and layer scaling hyperparameters are set to $\epsilon_f=0.99$ and $\gamma=0.2$. We run the MTIL experiments on a single NVIDIA 4090 GPU. For the reproduced methods, we performed careful hyperparameter tuning. For O-LoRA~\cite{wang2023orthogonal}, the learning rate is $5\times 10^{-4}$ with a regularization coefficient of $0.1$. For InfLoRA~\cite{liang2024inflora}, the learning rate is $10^{-3}$, with $\epsilon_f=0.99$. The learning rate for SD-LoRA~\cite{wu2025sd} is set to $5\times 10^{-3}$. 

\subsubsection{Visual Question Answering with LLaVA}
We adopt the LLaVA-1.5-7b~\cite{liu2023visual} model for multimodal continual instruction tuning experiments. Training is conducted on $4 \times$ NVIDIA H100 GPUs using the AdamW optimizer. For the MLLM-DCL benchmark, we set the learning rate to $2\times 10^{-5}$ and train for no more than $3$ epochs per task. For the UCIT benchmark, the learning rate is set to $2\times 10^{-4}$ for all tasks except Flickr30k, which uses $1\times 10^{-4}$ and train for $1$ epoch per task. The hyperparameters for the subspace constraints and layer scaling are configured as $\epsilon_w=0.6$, $\epsilon_f=0.99$, and $\gamma=0.2$

\subsubsection{Video Understanding with Video-LLaVA}
We adopt the Video-LLaVA-7b~\cite{lin2024video} model to evaluate continual video understanding capabilities on the CL-VISTA benchmark. Training is conducted on $2 \times$ NVIDIA H100 GPUs using the AdamW optimizer. Following the standard benchmark protocols, we uniformly sample $8$ frames from each video and optimize each task for $1$ epoch with a global batch size of $64$. The learning rate for the adaptation modules is set to $1\times 10^{-4}$. The hyperparameters for the subspace constraints and layer-wise scaling are configured as $\epsilon_w=0.6$, $\epsilon_f=0.99$, and $\gamma=0.2$.

\subsection{Theoretical Analysis of Subspace Adaptation}
\label{appendix:subspace_adaptation}

\subsubsection{Proof of Proposition~\ref{pro:p1}}

Suppose for loss function $\mathcal{L}$ for task $\mathcal{T}^{t}$ and a linear layer with $\boldsymbol{h} = \boldsymbol{xW}$, where $\boldsymbol{h}$ is the output of the layer and $\boldsymbol{x}$ is the input. We can compute the gradient of $\boldsymbol{B}_{t}$ directly as follows:
\begin{equation}
\frac{\partial \mathcal{L}}{\partial \boldsymbol{B}_{t}} = \frac{\partial \boldsymbol{W}}{\partial \boldsymbol{B}_{t}} \cdot \frac{\partial \mathcal{L}}{\partial \boldsymbol{W}} = \frac{\alpha}{r} \boldsymbol{A}_{t}^{\top} \boldsymbol{G}_{t}.
\end{equation}
In a gradient descent iteration, the change of $\boldsymbol{B}_{t}$ is represented by a negative gradient:
\begin{equation}
\Delta \boldsymbol{B}_{t} = -\frac{\eta\alpha}{r} \boldsymbol{A}_{t}^{\top} \boldsymbol{G}_{t}.
\end{equation}
Therefore, when $\boldsymbol{A}_{t}$ is frozen to only update $\boldsymbol{B}_{t}$ in each iteration, we can obtain the variation of $\boldsymbol{W}$ in one iteration to complete the proof:
\begin{equation}
\Delta \boldsymbol{W} = \frac{\alpha}{r} \boldsymbol{A}_{t} \Delta \boldsymbol{B}_{t} = -\frac{\eta \alpha^{2}}{r^{2}}\boldsymbol{A}_{t} \boldsymbol{A}_{t}^{\top} \boldsymbol{G}_{t}.
\end{equation}

\subsubsection{Proof of Proposition~\ref{pro:p2}}

We proceed by transforming the constrained optimization problem, leveraging subspace properties, and applying the Eckart--Young--Mirsky Theorem \cite{EYM} to confirm the optimal solution.

\noindent\textit{Step 1: Equivalent Transformation of the Objective Function.}
For an orthonormal matrix $\boldsymbol{A}_t$ satisfying $\boldsymbol{A}_t^\top \boldsymbol{A}_t = \boldsymbol{I}$, the orthogonal projection operator $\boldsymbol{P}_{\boldsymbol{A}_t} = \boldsymbol{A}_t\boldsymbol{A}_t^\top$ satisfies the Pythagorean theorem for the Frobenius norm:
\begin{equation}
\|\boldsymbol{G}_t\|_F^2 = \|\boldsymbol{P}_{\boldsymbol{A}_t} \boldsymbol{G}_t\|_F^2 + \|\boldsymbol{G}_t - \boldsymbol{P}_{\boldsymbol{A}_t} \boldsymbol{G}_t\|_F^2.
\end{equation}
Since $\|\boldsymbol{G}_t\|_F^2$ is a constant independent of $\boldsymbol{A}_t$, minimizing the original objective $\|\boldsymbol{G}_t - \boldsymbol{P}_{\boldsymbol{A}_t} \boldsymbol{G}_t\|_F^2$ is \textit{equivalent} to maximizing the projected norm $\|\boldsymbol{P}_{\boldsymbol{A}_t} \boldsymbol{G}_t\|_F^2$. The optimization problem thus can be rewritten as:
\begin{equation}
\begin{aligned}
\max_{\boldsymbol{A}_t^\top \boldsymbol{A}_t = \boldsymbol{I}} \quad & \|\boldsymbol{A}_t\boldsymbol{A}_t^\top \boldsymbol{G}_t\|_F^2, \\
\text{s.t.} \quad & \boldsymbol{W}_p^\top \boldsymbol{A}_t = \boldsymbol{0}, \\
& \boldsymbol{M}_{t-1}^\top \boldsymbol{A}_t = \boldsymbol{0}.
\end{aligned}
\end{equation}

\noindent\textit{Step 2: Substitute $\hat{\boldsymbol{G}}_t$ and Simplify Using Constraints.}
Recall from Eq.~\ref{eq5} that the projected gradient $\hat{\boldsymbol{G}}_t$ is defined as:
\begin{equation}
\hat{\boldsymbol{G}}_t = \boldsymbol{G}_t - \boldsymbol{W}_p\boldsymbol{W}_p^\top\boldsymbol{G}_t - \boldsymbol{M}_{t-1}\boldsymbol{M}_{t-1}^\top\boldsymbol{G}_t.
\end{equation}
Rearranging gives
\begin{equation}
\boldsymbol{G}_t = \hat{\boldsymbol{G}}_t + \boldsymbol{W}_p\boldsymbol{W}_p^\top\boldsymbol{G}_t + \boldsymbol{M}_{t-1}\boldsymbol{M}_{t-1}^\top\boldsymbol{G}_t.
\end{equation}
Substitute this into the objective:
\begin{equation}
\|\boldsymbol{A}_t\boldsymbol{A}_t^\top \left( \hat{\boldsymbol{G}}_t + \boldsymbol{W}_p\boldsymbol{W}_p^\top\boldsymbol{G}_t + \boldsymbol{M}_{t-1}\boldsymbol{M}_{t-1}^\top\boldsymbol{G}_t \right)\|_F^2.
\end{equation}
For any feasible $\boldsymbol{A}_t$, we use $\boldsymbol{W}_p^\top \boldsymbol{A}_t = \boldsymbol{M}_{t-1}^\top \boldsymbol{A}_t = \boldsymbol{0}$ to simplify:
\begin{equation}
\boldsymbol{A}_t^\top (\boldsymbol{W}_p\boldsymbol{W}_p^\top\boldsymbol{G}_t)
= (\boldsymbol{W}_p^\top \boldsymbol{A}_t)^\top (\boldsymbol{W}_p^\top\boldsymbol{G}_t)
= \boldsymbol{0},
\end{equation}
and similarly,
\begin{equation}
\boldsymbol{A}_t^\top (\boldsymbol{M}_{t-1}\boldsymbol{M}_{t-1}^\top\boldsymbol{G}_t) = \boldsymbol{0}.
\end{equation}
Thus,
\begin{equation}
\boldsymbol{A}_t\boldsymbol{A}_t^\top (\boldsymbol{W}_p\boldsymbol{W}_p^\top\boldsymbol{G}_t + \boldsymbol{M}_{t-1}\boldsymbol{M}_{t-1}^\top\boldsymbol{G}_t) = \boldsymbol{0},
\end{equation}
and the objective reduces to maximizing $\|\boldsymbol{A}_t\boldsymbol{A}_t^\top \hat{\boldsymbol{G}}_t\|_F^2$. The optimization problem simplifies to:
\begin{equation}
\label{eq:reduced_optim}
\begin{aligned}
\max_{\boldsymbol{A}_t^\top \boldsymbol{A}_t = \boldsymbol{I}} \quad & \|\boldsymbol{A}_t\boldsymbol{A}_t^\top \hat{\boldsymbol{G}}_t\|_F^2, \\
\text{s.t.} \quad & \boldsymbol{W}_p^\top \boldsymbol{A}_t = \boldsymbol{0}, \\
& \boldsymbol{M}_{t-1}^\top \boldsymbol{A}_t = \boldsymbol{0}.
\end{aligned}
\end{equation}

\noindent\textit{Step 3: Optimal $\boldsymbol{A}_t$ via Eckart--Young--Mirsky Theorem.}
The Eckart--Young--Mirsky Theorem \cite{EYM} states that for any matrix $\boldsymbol{X} \in \mathbb{R}^{m \times n}$ and integer $r \leq \min(m,n)$, the $r$-dimensional subspace that maximizes $\|\boldsymbol{P} \boldsymbol{X}\|_F^2$, where $\boldsymbol{P}$ is the orthogonal projection onto the subspace, is spanned by the top-$r$ left singular vectors of $\boldsymbol{X}$.

Ignoring the orthogonality constraints involving $\boldsymbol{W}_p$ and $\boldsymbol{M}_{t-1}$ gives an upper bound on the objective in Eq.~\ref{eq:reduced_optim}. By the theorem, the top-$r$ left singular vectors of $\hat{\boldsymbol{G}}_t$ attain this bound.

\noindent\textit{Step 4: Verify Feasibility of the Optimal $\boldsymbol{A}_t$.}
We confirm the optimal $\boldsymbol{A}_t$ satisfies the constraints $\boldsymbol{W}_p^\top \boldsymbol{A}_t = \boldsymbol{0}$ and $\boldsymbol{M}_{t-1}^\top \boldsymbol{A}_t = \boldsymbol{0}$.

With $\boldsymbol{Q}=[\boldsymbol{W}_p,\boldsymbol{M}_{t-1}]$ and $\boldsymbol{Q}^\top\boldsymbol{Q}=\boldsymbol{I}$, Eq.~\ref{eq5} gives $\hat{\boldsymbol{G}}_t=(\boldsymbol{I}-\boldsymbol{Q}\boldsymbol{Q}^\top)\boldsymbol{G}_t$. Hence:
\begin{equation}
\label{eq13}
\boldsymbol{W}_p^\top \hat{\boldsymbol{G}}_t = \boldsymbol{0}, \quad \boldsymbol{M}_{t-1}^\top \hat{\boldsymbol{G}}_t = \boldsymbol{0}.
\end{equation}
Let $\hat{\boldsymbol{G}}_t=\boldsymbol{U}\boldsymbol{S}\boldsymbol{V}^\top$ be its SVD. Since $r\leq\operatorname{rank}(\hat{\boldsymbol{G}}_t)$, each of its first $r$ singular values satisfies $\sigma_j>0$. For the corresponding singular vectors $\boldsymbol{u}_j$ and $\boldsymbol{v}_j$, Eq.~\ref{eq13} gives
\[
\boldsymbol{Q}^\top \boldsymbol{u}_j
= \sigma_j^{-1}\boldsymbol{Q}^\top\hat{\boldsymbol{G}}_t\boldsymbol{v}_j
= \boldsymbol{0},\qquad j=1,\ldots,r.
\]
Thus $\boldsymbol{A}_t=\boldsymbol{U}_{:,1:r}$ is feasible and attains the upper bound from Step 3, so it is an optimal solution to Eq.~\ref{eq:optimization}.

\subsection{Theoretical Analysis of Layer-Wise Scaling}
\label{appendix:layer_scaling}

Building on the projected-update geometry of Proposition~\ref{pro:p1} and the subspace initialization of Proposition~\ref{pro:p2}, we analyze how layer scaling distributes adaptation across fixed trainable spaces. For task $\mathcal{T}^t$ and target layer $l\in\{1,\ldots,L\}$, the scale is $\alpha^l=\varphi(l,L,\alpha_0,\gamma)$ as in Eq.~\ref{eq:scale}. We quantify retention through a reference loss and derive the allocation that minimizes its quadratic risk budget at matched first-order adaptation progress.

\subsubsection{Fixed Subspaces and Effective Updates}
At a fixed iterate on task $\mathcal{T}^t$, the effective weights are
\begin{equation}
\label{eq:ls_parameterization}
\boldsymbol{W}^l=\boldsymbol{W}^{\prime l}
+\frac{\alpha^l}{r}\boldsymbol{A}_t^l\boldsymbol{B}_t^l,
\end{equation}
where $\boldsymbol{W}^{\prime l}$ and $\boldsymbol{A}_t^l$ are frozen, and only $\boldsymbol{B}_t^l$ is optimized. The factors satisfy the layer-indexed constraints of Proposition~\ref{pro:p2}:
\begin{equation}
\label{eq:ls_subspace}
\begin{aligned}
(\boldsymbol{A}_t^l)^\top\boldsymbol{A}_t^l&=\boldsymbol{I},\\
(\boldsymbol{W}_p^l)^\top\boldsymbol{A}_t^l
&=(\boldsymbol{M}_{t-1}^l)^\top\boldsymbol{A}_t^l=\boldsymbol{0}.
\end{aligned}
\end{equation}
Following the projector notation in the proof of Proposition~\ref{pro:p2}, define
\begin{equation}
\label{eq:ls_projector}
\boldsymbol{P}_{\boldsymbol{A}_t^l}
=\boldsymbol{A}_t^l(\boldsymbol{A}_t^l)^\top.
\end{equation}
Eq.~\ref{eq:ls_projector} projects onto $\operatorname{span}(\boldsymbol{A}_t^l)$, a rank-$r$ subspace of the admissible residual space $\operatorname{span}([\boldsymbol{W}_p^l,\boldsymbol{M}_{t-1}^l])^\perp$. Within this space, $\boldsymbol{P}_{\boldsymbol{A}_t^l}\boldsymbol{G}_t^l$ is the projection of $\hat{\boldsymbol{G}}_t^l$ onto the selected trainable directions. The constraints in Eq.~\ref{eq:ls_subspace} specify admissibility through orthogonality to the stored knowledge directions.

Here $\boldsymbol{G}_t^l=\nabla_{\boldsymbol{W}^l}\mathcal{L}(\boldsymbol{W};\mathcal{T}^t)$ denotes the current-task gradient at the iterate under study, and $\boldsymbol{A}_t^l$ is fixed at its task-initialization value. We use $\boldsymbol{W}=(\boldsymbol{W}^1,\ldots,\boldsymbol{W}^L)$ for the collection of target weights, with the sum Frobenius inner product; all other parameters are held fixed. Matrices at the same depth that share a scale can be grouped into one block. Orthonormality gives
\begin{equation}
\label{eq:ls_gradient_energy}
\begin{aligned}
\langle\boldsymbol{G}_t^l,
\boldsymbol{P}_{\boldsymbol{A}_t^l}\boldsymbol{G}_t^l\rangle
&=\|\boldsymbol{P}_{\boldsymbol{A}_t^l}\boldsymbol{G}_t^l\|_F^2\\
&=\|(\boldsymbol{A}_t^l)^\top\boldsymbol{G}_t^l\|_F^2.
\end{aligned}
\end{equation}
By Proposition~\ref{pro:p1}, one ordinary gradient-descent step with learning rate $\eta>0$ satisfies
\begin{equation}
\label{eq:ls_effective_step}
\begin{aligned}
\nabla_{\boldsymbol{B}_t^l}\mathcal{L}
&=\frac{\alpha^l}{r}(\boldsymbol{A}_t^l)^\top\boldsymbol{G}_t^l,\\
\Delta\boldsymbol{W}^l
&=-c^l\boldsymbol{P}_{\boldsymbol{A}_t^l}\boldsymbol{G}_t^l,
\end{aligned}
\end{equation}
where $c^l=\eta(\alpha^l)^2/r^2$. Thus, the LoRA scale enters the effective gradient-descent step quadratically, with $c^l$ controlling the magnitude of the projected update.

\subsubsection{Local Retention-Risk Bound}
Let $\mathcal{L}_{\mathrm{ret}}(\boldsymbol{W})$ quantify retention through a previous-task loss, a pre-training reference loss, a nonnegative weighted combination of these losses, or a fixed-reference prediction discrepancy. The current-task loss $\mathcal{L}$ measures adaptation progress.

\begin{assumption}[Restricted retention curvature]
\label{ass:ls_curvature}
The function $\mathcal{L}_{\mathrm{ret}}$ is twice continuously differentiable on a convex neighborhood $\mathcal{N}$ of the current $\boldsymbol{W}$. There exist $\kappa^l\geq0$ such that, for every $\widetilde{\boldsymbol{W}}\in\mathcal{N}$ and every perturbation $\Delta\boldsymbol{W}^l=\boldsymbol{A}_t^l\boldsymbol{C}^l$,
\begin{equation}
\label{eq:ls_majorizer}
\langle\Delta\boldsymbol{W},
\nabla^2\mathcal{L}_{\mathrm{ret}}(\widetilde{\boldsymbol{W}})
[\Delta\boldsymbol{W}]\rangle
\leq\sum_l\kappa^l\|\Delta\boldsymbol{W}^l\|_F^2.
\end{equation}
\end{assumption}

The effective-weight coordinates $\boldsymbol{C}^l$ satisfy $\|\Delta\boldsymbol{W}^l\|_F=\|\boldsymbol{C}^l\|_F$. Assumption~\ref{ass:ls_curvature} gives a block-diagonal upper majorizer of the full restricted Hessian, incorporating cross-layer interactions. For example, suppose symmetric constants $h^{lm}=h^{ml}\geq0$ uniformly bound its blocks so that
\begin{equation}
\label{eq:ls_cross_layer}
\begin{aligned}
&\big|\langle\boldsymbol{A}_t^l\boldsymbol{C}^l,
\nabla^2_{\boldsymbol{W}^l\boldsymbol{W}^m}
\mathcal{L}_{\mathrm{ret}}(\widetilde{\boldsymbol{W}})
[\boldsymbol{A}_t^m\boldsymbol{C}^m]\rangle\big|\\
&\leq h^{lm}\|\boldsymbol{C}^l\|_F\|\boldsymbol{C}^m\|_F.
\end{aligned}
\end{equation}
Then $\kappa^l=\sum_m h^{lm}$ is valid, since
\begin{equation}
\label{eq:ls_row_sum}
\begin{aligned}
&\langle\Delta\boldsymbol{W},
\nabla^2\mathcal{L}_{\mathrm{ret}}[\Delta\boldsymbol{W}]\rangle\\
&\leq\sum_{l,m}h^{lm}\|\boldsymbol{C}^l\|_F\|\boldsymbol{C}^m\|_F\\
&\leq\sum_l\Big(\sum_m h^{lm}\Big)\|\boldsymbol{C}^l\|_F^2\\
&=\sum_l\kappa^l\|\Delta\boldsymbol{W}^l\|_F^2.
\end{aligned}
\end{equation}
The last inequality uses $2ab\leq a^2+b^2$ and symmetry. This construction incorporates both within-layer and cross-layer curvature contributions into $\kappa^l$.

Define the surviving first-order retention sensitivity by
\begin{equation}
\label{eq:ls_first_order}
\rho^l=\|\boldsymbol{P}_{\boldsymbol{A}_t^l}
\nabla_{\boldsymbol{W}^l}\mathcal{L}_{\mathrm{ret}}(\boldsymbol{W})\|_F.
\end{equation}

\begin{lemma}[Local retention-risk bound]
\label{lem:ls_risk}
Under Assumption~\ref{ass:ls_curvature}, any perturbation in the fixed trainable spaces with $\boldsymbol{W}+\Delta\boldsymbol{W}\in\mathcal{N}$ satisfies
\begin{equation}
\label{eq:ls_general_risk}
\begin{aligned}
&\mathcal{L}_{\mathrm{ret}}(\boldsymbol{W}+\Delta\boldsymbol{W})
-\mathcal{L}_{\mathrm{ret}}(\boldsymbol{W})\\
&\leq\sum_l\langle
\nabla_{\boldsymbol{W}^l}\mathcal{L}_{\mathrm{ret}}(\boldsymbol{W}),
\Delta\boldsymbol{W}^l\rangle
+\frac12\sum_l\kappa^l\|\Delta\boldsymbol{W}^l\|_F^2\\
&\leq\sum_l\rho^l\|\Delta\boldsymbol{W}^l\|_F
+\frac12\sum_l\kappa^l\|\Delta\boldsymbol{W}^l\|_F^2.
\end{aligned}
\end{equation}
In particular, the update in Eq.~\ref{eq:ls_effective_step} obeys
\begin{equation}
\label{eq:ls_gd_risk}
\begin{aligned}
&\mathcal{L}_{\mathrm{ret}}(\boldsymbol{W}+\Delta\boldsymbol{W})
-\mathcal{L}_{\mathrm{ret}}(\boldsymbol{W})\\
&\leq-\sum_l c^l\langle
\nabla_{\boldsymbol{W}^l}\mathcal{L}_{\mathrm{ret}}(\boldsymbol{W}),
\boldsymbol{P}_{\boldsymbol{A}_t^l}\boldsymbol{G}_t^l\rangle
+\mathcal{C}(\{c^l\})\\
&\leq\sum_l\rho^l c^l
\|(\boldsymbol{A}_t^l)^\top\boldsymbol{G}_t^l\|_F
+\mathcal{C}(\{c^l\}),
\end{aligned}
\end{equation}
where $\mathcal{C}$ is the quadratic budget in Eq.~\ref{eq:ls_allocation}.
\end{lemma}

Proof of Lemma~\ref{lem:ls_risk}.
Apply the integral Taylor formula along $\boldsymbol{W}+u\Delta\boldsymbol{W}$, $0\leq u\leq1$. Convexity keeps this segment in $\mathcal{N}$. Assumption~\ref{ass:ls_curvature} bounds the second-order remainder by $\frac12\sum_l\kappa^l\|\Delta\boldsymbol{W}^l\|_F^2$, since $\int_0^1(1-u)\,du=1/2$. For the linear term, $\boldsymbol{P}_{\boldsymbol{A}_t^l}\Delta\boldsymbol{W}^l=\Delta\boldsymbol{W}^l$ implies
\begin{equation}
\begin{aligned}
\langle\nabla_{\boldsymbol{W}^l}\mathcal{L}_{\mathrm{ret}},
\Delta\boldsymbol{W}^l\rangle
&=\langle\boldsymbol{P}_{\boldsymbol{A}_t^l}
\nabla_{\boldsymbol{W}^l}\mathcal{L}_{\mathrm{ret}},
\Delta\boldsymbol{W}^l\rangle\\
&\leq\rho^l\|\Delta\boldsymbol{W}^l\|_F.
\end{aligned}
\end{equation}
This proves Eq.~\ref{eq:ls_general_risk}. Substituting Eq.~\ref{eq:ls_effective_step} and using Eq.~\ref{eq:ls_gradient_energy} gives Eq.~\ref{eq:ls_gd_risk}, completing the proof.

Lemma~\ref{lem:ls_risk} separates retention risk into projected first-order sensitivity and restricted second-order curvature. Layer scaling controls both contributions through $c^l$, while the fixed factors $\boldsymbol{A}_t^l$ specify the update directions.

\subsubsection{Matched Plasticity and Proof of Proposition~\ref{pro:layer_scaling}}
All allocation comparisons use the same checkpoint, frozen factors, and current gradient. Hereafter sums and weighted moments are over the active layers for which $\|(\boldsymbol{A}_t^l)^\top\boldsymbol{G}_t^l\|_F>0$, and at least one such layer is assumed to exist. By Eq.~\ref{eq:ls_gradient_energy}, the first-order current-task progress is
\begin{equation}
\label{eq:ls_progress}
\begin{aligned}
\delta&=-\langle\nabla\mathcal{L}(\boldsymbol{W}),
\Delta\boldsymbol{W}\rangle\\
&=\sum_l c^l\|(\boldsymbol{A}_t^l)^\top\boldsymbol{G}_t^l\|_F^2.
\end{aligned}
\end{equation}
Fixing $\delta>0$ separates layer allocation from changing the overall update magnitude. If $\nabla\mathcal{L}$ is $L_{\mathcal{L}}$-Lipschitz in the relevant neighborhood, then
\begin{equation}
\label{eq:ls_task_descent}
\begin{aligned}
&\mathcal{L}(\boldsymbol{W}+\Delta\boldsymbol{W})
-\mathcal{L}(\boldsymbol{W})\\
&\leq-\delta
+\frac{L_{\mathcal{L}}}{2}\sum_l(c^l)^2
\|(\boldsymbol{A}_t^l)^\top\boldsymbol{G}_t^l\|_F^2.
\end{aligned}
\end{equation}
The common value $\delta$ matches first-order plasticity, while Eq.~\ref{eq:ls_task_descent} quantifies the second-order contribution to the actual loss decrease.

Proof of Proposition~\ref{pro:layer_scaling}.
By Cauchy--Schwarz, applied to the factors $\sqrt{\kappa^l}c^l\|(\boldsymbol{A}_t^l)^\top\boldsymbol{G}_t^l\|_F$ and $\|(\boldsymbol{A}_t^l)^\top\boldsymbol{G}_t^l\|_F/\sqrt{\kappa^l}$,
\begin{equation}
\label{eq:ls_cauchy}
\delta^2\leq2\mathcal{C}(\{c^l\})
\sum_l\frac{\|(\boldsymbol{A}_t^l)^\top\boldsymbol{G}_t^l\|_F^2}{\kappa^l}.
\end{equation}
Equality holds exactly when $\kappa^l c^l$ is constant on the active layers. Enforcing Eq.~\ref{eq:ls_progress} gives the normalized allocation and its minimum budget:
\begin{equation}
\label{eq:ls_optimal_value}
\begin{aligned}
c^{l,\star}&=\frac{\delta/\kappa^l}
{\sum_j\|(\boldsymbol{A}_t^j)^\top\boldsymbol{G}_t^j\|_F^2/\kappa^j},\\
\mathcal{C}^{\star}&=\frac{\delta^2}
{2\sum_l\|(\boldsymbol{A}_t^l)^\top\boldsymbol{G}_t^l\|_F^2/\kappa^l}.
\end{aligned}
\end{equation}
Positive $\kappa^l$ and nonzero projected-gradient energy make the objective strictly convex on the active coordinates, ensuring uniqueness. The scale relation $\alpha^{l,\star}=r\sqrt{c^{l,\star}/\eta}$ follows from Eq.~\ref{eq:ls_effective_step}, establishing the proportionality relations in Eq.~\ref{eq:ls_optimal}. Taking $\delta$ sufficiently small keeps the updates in $\mathcal{N}$, where Lemma~\ref{lem:ls_risk} applies. The proof is completed.

To compare against uniform allocation, define gradient-energy weights
\begin{equation}
\label{eq:ls_weights}
\omega^l=\frac{\|(\boldsymbol{A}_t^l)^\top\boldsymbol{G}_t^l\|_F^2}
{\sum_j\|(\boldsymbol{A}_t^j)^\top\boldsymbol{G}_t^j\|_F^2}.
\end{equation}
For layerwise scalars $u^l$, write $\E_\omega[u]=\sum_l\omega^l u^l$, and use the corresponding weighted covariance and variance:
\begin{equation}
\label{eq:ls_moments}
\begin{aligned}
\operatorname{Cov}_\omega(u,v)
&=\E_\omega[uv]-\E_\omega[u]\E_\omega[v],\\
\operatorname{Var}_\omega(u)
&=\E_\omega[u^2]-(\E_\omega[u])^2.
\end{aligned}
\end{equation}
Uniform effective steps have $c^l_{\mathrm{unif}}=\delta/\sum_j\|(\boldsymbol{A}_t^j)^\top\boldsymbol{G}_t^j\|_F^2$, and their budget satisfies
\begin{equation}
\label{eq:ls_uniform_gap}
\begin{aligned}
\mathcal{C}_{\mathrm{unif}}
&=\frac{\delta^2\E_\omega[\kappa]}
{2\sum_l\|(\boldsymbol{A}_t^l)^\top\boldsymbol{G}_t^l\|_F^2},\\
\frac{\mathcal{C}_{\mathrm{unif}}}{\mathcal{C}^{\star}}
&=\E_\omega[\kappa]\E_\omega[\kappa^{-1}]\geq1.
\end{aligned}
\end{equation}
Equality holds if and only if $\kappa^l$ is constant on the active layers. The gap $\E_\omega[\kappa]\E_\omega[\kappa^{-1}]$ quantifies the benefit of adapting layer-wise steps to curvature heterogeneity. If an active layer has $\kappa^l=0$, assigning all progress to zero-curvature layers attains zero quadratic budget.

\subsubsection{Analysis of the Linear Schedule}
\label{appendix:ls_linear}
For $L\geq2$, retain Eq.~\ref{eq:scale} exactly and define its unit-base-scale profile by
\begin{equation}
\label{eq:ls_profile}
\begin{aligned}
d^l&=\frac{l-1}{L-1},\\
\bar\alpha^l(\gamma)&=\varphi(l,L,1,\gamma)
=\gamma+(1-\gamma)d^l.
\end{aligned}
\end{equation}
Thus $\alpha^l=\alpha_0\bar\alpha^l(\gamma)$, with the same shallowest and deepest scales $\gamma\alpha_0$ and $\alpha_0$ as in the main paper. To compare schedule shapes at matched $\delta$, normalize the effective steps as
\begin{equation}
\label{eq:ls_normalized_schedule}
c^l(\gamma)=\frac{\delta(\bar\alpha^l(\gamma))^2}
{\sum_j\|(\boldsymbol{A}_t^j)^\top\boldsymbol{G}_t^j\|_F^2
(\bar\alpha^j(\gamma))^2}.
\end{equation}
Equivalently, the base scale for this comparison satisfies
\begin{equation}
\label{eq:ls_base_normalization}
\alpha_0(\gamma)^2=\frac{r^2\delta}
{\eta\sum_j\|(\boldsymbol{A}_t^j)^\top\boldsymbol{G}_t^j\|_F^2
(\bar\alpha^j(\gamma))^2}.
\end{equation}
This normalization holds first-order progress fixed across schedule shapes. At a fixed $\alpha_0$, varying $\gamma$ changes both progress and allocation. The uniform schedule is $\gamma=1$, and reducing $\gamma$ shifts the normalized allocation toward deeper layers.

\begin{corollary}[Local benefit of shallow-to-deep scaling]
\label{cor:ls_linear}
For Eq.~\ref{eq:ls_normalized_schedule},
\begin{equation}
\label{eq:ls_covariance}
\left.\frac{d\mathcal{C}(\{c^l(\gamma)\})}{d\gamma}
\right|_{\gamma=1^-}
=-\frac{2\delta^2\operatorname{Cov}_\omega(\kappa,d)}
{\sum_l\|(\boldsymbol{A}_t^l)^\top\boldsymbol{G}_t^l\|_F^2}.
\end{equation}
If $\operatorname{Cov}_\omega(\kappa,d)<0$, there exists $\varepsilon\in(0,1)$ such that $\mathcal{C}(\{c^l(\gamma)\})<\mathcal{C}_{\mathrm{unif}}$ whenever $1-\varepsilon<\gamma<1$. When $\rho^l=0$ on active layers and the updates stay in $\mathcal{N}$, this improves the complete upper bound of Lemma~\ref{lem:ls_risk} at matched first-order plasticity.
\end{corollary}

Proof of Corollary~\ref{cor:ls_linear}.
Differentiating Eq.~\ref{eq:ls_normalized_schedule} at $\gamma=1$ yields
\begin{equation}
\label{eq:ls_step_derivative}
\begin{aligned}
c^l(1)&=\frac{\delta}
{\sum_j\|(\boldsymbol{A}_t^j)^\top\boldsymbol{G}_t^j\|_F^2},\\
\left.\frac{dc^l}{d\gamma}\right|_{\gamma=1^-}
&=-\frac{2\delta(d^l-\E_\omega[d])}
{\sum_j\|(\boldsymbol{A}_t^j)^\top\boldsymbol{G}_t^j\|_F^2}.
\end{aligned}
\end{equation}
Substitution into the derivative of Eq.~\ref{eq:ls_allocation} gives
\begin{equation}
\begin{aligned}
\left.\frac{d\mathcal{C}}{d\gamma}\right|_{\gamma=1^-}
&=\sum_l\kappa^l\|(\boldsymbol{A}_t^l)^\top\boldsymbol{G}_t^l\|_F^2
c^l(1)\left.\frac{dc^l}{d\gamma}\right|_{\gamma=1^-}\\
&=-\frac{2\delta^2\operatorname{Cov}_\omega(\kappa,d)}
{\sum_l\|(\boldsymbol{A}_t^l)^\top\boldsymbol{G}_t^l\|_F^2}.
\end{aligned}
\end{equation}
Negative covariance makes this left derivative positive, so a sufficiently small decrease of $\gamma$ below one strictly decreases the budget. The weighted covariance couples retention curvature with normalized depth through the current-task gradient energy. The proof is completed.

For a finite scaling strength, direct substitution of any positive profile $\bar\alpha^l$ into the normalized allocation gives
\begin{equation}
\label{eq:ls_finite_budget}
\mathcal{C}_\gamma
=\frac{\delta^2}
{2\sum_l\|(\boldsymbol{A}_t^l)^\top\boldsymbol{G}_t^l\|_F^2}
\frac{\E_\omega[\kappa\bar\alpha^4]}
{(\E_\omega[\bar\alpha^2])^2}.
\end{equation}
Hence the necessary and sufficient condition for improving the quadratic budget over uniform scaling at matched progress is
\begin{equation}
\label{eq:ls_finite_condition}
\mathcal{C}_\gamma<\mathcal{C}_{\mathrm{unif}}
\quad\Longleftrightarrow\quad
\E_\omega[\kappa\bar\alpha^4]
<\E_\omega[\kappa](\E_\omega[\bar\alpha^2])^2.
\end{equation}
Equivalently,
\begin{equation}
\operatorname{Cov}_\omega(\kappa,\bar\alpha^4)
+\E_\omega[\kappa]\operatorname{Var}_\omega(\bar\alpha^2)<0.
\end{equation}
The variance term quantifies the concentration cost of layer-wise allocation. Eq.~\ref{eq:ls_finite_condition} balances this cost against curvature-depth alignment at a finite scaling strength, including the experimental setting $\gamma=0.2$.

\subsubsection{First-Order Retention and Practical Scaling}
\textit{First-order retention.}
An illustrative retention objective with zero first-order term is a frozen-teacher prediction KL anchored at the same checkpoint used for the local comparison:
\begin{equation}
\mathcal{L}_{\mathrm{ret}}(\widetilde{\boldsymbol{W}})
=\E_{\boldsymbol{x}}D_{\mathrm{KL}}\!\left(
p_{\boldsymbol{W}}(\cdot\mid\boldsymbol{x})
\,\Vert\,
p_{\widetilde{\boldsymbol{W}}}(\cdot\mid\boldsymbol{x})\right).
\end{equation}
Under standard differentiability conditions its gradient vanishes at $\widetilde{\boldsymbol{W}}=\boldsymbol{W}$, giving $\rho^l=0$ and reducing the bound in Eq.~\ref{eq:ls_gd_risk} to $\mathcal{C}$. For general retention sensitivities $\rho^l$, define the full norm-based bound
\begin{equation}
\mathcal{B}(\{c^l\})
=\sum_l\rho^l c^l\|(\boldsymbol{A}_t^l)^\top\boldsymbol{G}_t^l\|_F
+\mathcal{C}(\{c^l\}).
\end{equation}
Along Eq.~\ref{eq:ls_normalized_schedule},
\begin{equation}
\label{eq:ls_nonzero_first_order}
\begin{aligned}
\left.\frac{d\mathcal{B}}{d\gamma}\right|_{\gamma=1^-}
&=-2\delta\operatorname{Cov}_\omega\!\left(
\frac{\rho^l}{\|(\boldsymbol{A}_t^l)^\top\boldsymbol{G}_t^l\|_F},d^l\right)\\
&\quad-\frac{2\delta^2\operatorname{Cov}_\omega(\kappa,d)}
{\sum_l\|(\boldsymbol{A}_t^l)^\top\boldsymbol{G}_t^l\|_F^2},
\end{aligned}
\end{equation}
where the covariance uses the layerwise values with weights from Eq.~\ref{eq:ls_weights}. A positive value of this derivative gives a local improvement of the full retention bound as $\gamma$ decreases from one. The two weighted covariances capture the contributions of first-order sensitivity and retention curvature to this improvement.

\textit{Effective updates.}
Eq.~\ref{eq:ls_effective_step} describes the relationship between the LoRA scale and the effective update under ordinary gradient descent. Adam~\cite{kingma2015adam} forms the update of $\boldsymbol{B}_t^l$ through adaptive moment estimates, while AdamW~\cite{loshchilov2018decoupled} additionally incorporates decoupled weight decay. The induced effective update $\Delta\boldsymbol{W}^l=(\alpha^l/r)\boldsymbol{A}_t^l\Delta\boldsymbol{B}_t^l$ stays in $\operatorname{span}(\boldsymbol{A}_t^l)$, so Eq.~\ref{eq:ls_general_risk} applies directly. Assumption~\ref{ass:ls_curvature} uses uniform neighborhood bounds on the full restricted Hessian. Normalized directional second differences along $\boldsymbol{P}_{\boldsymbol{A}_t^l}\boldsymbol{G}_t^l$ provide checkpoint-level curvature diagnostics.

\textit{Depth-based allocation.}
LiNeS~\cite{wang2025lines} motivates depth-based scaling through post-training differences, while Transformer Hessian analyses~\cite{zhang2024transformers} and layer-adaptive continual-learning regularization~\cite{moser2026layers} motivate heterogeneous layer sensitivities. Our analysis connects these observations with the projected-update geometry of KeepLoRA++: Proposition~\ref{pro:layer_scaling} allocates effective steps according to restricted retention curvature, Corollary~\ref{cor:ls_linear} relates the linear schedule to weighted depth alignment, and Eq.~\ref{eq:ls_finite_condition} balances alignment and concentration at a chosen scaling strength. Together, these results characterize how residual direction selection and layer-wise scaling coordinate adaptation progress and retention risk at a given checkpoint.

\subsection{Theoretical Analysis of Initialization of LoRA}
\label{appendix:b_init}

Proof of Proposition~\ref{pro:b_zero_equivalence}. At initialization ($k=0$), both parameterizations yield the same effective weight $\boldsymbol{W}$, satisfying
$\widetilde{\boldsymbol{B}}_t^{(0)}
=\boldsymbol{B}_t^{(0)}-\boldsymbol{B}_t^{(0)}
=\boldsymbol{0}$.

Suppose that at iteration $k$, the two parameterizations yield identical effective weights and satisfy
$\widetilde{\boldsymbol{B}}_t^{(k)}
=\boldsymbol{B}_t^{(k)}-\boldsymbol{B}_t^{(0)}$.
Since the effective weights are identical, they produce the same gradient $\boldsymbol{G}^{(k)}$ under the same loss and training data. With frozen $\boldsymbol{A}_t$, we have:
\begin{equation}
\frac{\partial\mathcal{L}}{\partial\boldsymbol{B}_t}
=
\frac{\partial\mathcal{L}}
{\partial\widetilde{\boldsymbol{B}}_t}
=s\boldsymbol{A}_t^\top\boldsymbol{G}^{(k)}.
\end{equation}
Consequently, gradient descent produces identical parameter updates:
\begin{equation}
\Delta\boldsymbol{B}_t^{(k)}
=\Delta\widetilde{\boldsymbol{B}}_t^{(k)}
=-\eta s\boldsymbol{A}_t^\top\boldsymbol{G}^{(k)}.
\end{equation}
Since $\boldsymbol{A}_t$ and $s$ remain fixed, the corresponding effective weight updates satisfy:
\begin{equation}
\begin{aligned}
\Delta\boldsymbol{W}_{\mathrm{c}}^{(k)}
&=s\boldsymbol{A}_t\Delta\boldsymbol{B}_t^{(k)},\\
\Delta\boldsymbol{W}_{\mathrm{z}}^{(k)}
&=s\boldsymbol{A}_t\Delta\widetilde{\boldsymbol{B}}_t^{(k)}
=\Delta\boldsymbol{W}_{\mathrm{c}}^{(k)}.
\end{aligned}
\end{equation}
Moreover, identical parameter updates preserve
$\widetilde{\boldsymbol{B}}_t^{(k+1)}
=\boldsymbol{B}_t^{(k+1)}-\boldsymbol{B}_t^{(0)}$.
Thus, both parameterizations yield identical effective weights at iteration $k+1$. By induction, the equivalence holds for all iterations, completing the proof.

\subsection{Supplementary Experiments}

\subsubsection{Additional Analyses of Layer-Wise Scaling and Robustness}
To further investigate the effects of layer-wise scaling, we conduct three complementary analyses on the MTIL.

Table~\ref{tab:layer_scale_comparison} compares the effects of applying the same shallow-to-deep scaling strategy to LoRA-FT and KeepLoRA. We set  rank $r=16$. For LoRA-FT, layer-wise scaling improves Last while reducing Average and leaving Transfer unchanged. In contrast, KeepLoRA++ improves all three aggregate metrics over KeepLoRA. These results suggest that layer-wise scaling alone does not consistently improve continual learning performance, whereas its combination with residual gradient adaptation provides complementary benefits.

\begin{table}[htbp]
\centering
\caption{Effects of layer-wise scaling on LoRA-FT and KeepLoRA in continual learning on MTIL (\%). The best results are in \textbf{bold}.}
\label{tab:layer_scale_comparison}
\footnotesize
\setlength{\tabcolsep}{5pt}
\renewcommand{\arraystretch}{1.12}
\begin{tabular}{lccc}
\toprule
Method & Transfer & Average & Last \\
\midrule
LoRA-FT & 65.4 & 69.6 & 71.6 \\
LoRA-FT + Layer-wise Scaling & 65.4 & 69.3 & 72.6 \\
\midrule
KeepLoRA & 69.0 & 77.5 & 86.1 \\
KeepLoRA++ & \textbf{69.5} & \textbf{78.0} & \textbf{86.6} \\
\bottomrule
\end{tabular}
\end{table}

Table~\ref{tab:uniform_scaling_comparison} investigates whether the improvement of KeepLoRA++ can be attributed simply to a reduced average scaling magnitude. We compare KeepLoRA++ with fixed uniform scaling and learnable scaling coefficients, both initialized with the same average scaling factor of $0.6\alpha_0$. Neither control reproduces the performance of KeepLoRA++, indicating that the improvement comes from the depth-dependent allocation of update magnitudes rather than merely reducing the overall scaling factor or introducing learnable scaling coefficients.

\begin{table}[htbp]
\centering
\caption{Uniform versus layer-wise scaling on MTIL (\%). The best results are in \textbf{bold}, and the second-best results are \underline{underlined}.}
\label{tab:uniform_scaling_comparison}
\footnotesize
\setlength{\tabcolsep}{5pt}
\renewcommand{\arraystretch}{1.12}
\begin{tabular}{lccc}
\toprule
Method & Transfer & Average & Last \\
\midrule
KeepLoRA & \underline{69.0} & \underline{77.5} & \underline{86.1} \\
\quad + Fixed Uniform Scaling ($0.6\alpha_0$) & 68.5 & 77.2 & 86.0 \\
\quad + Learnable Scaling (Init. $0.6\alpha_0$) & 68.3 & 76.3 & 83.0 \\
KeepLoRA++ ($\gamma=0.2$) & \textbf{69.5} & \textbf{78.0} & \textbf{86.6} \\
\bottomrule
\end{tabular}
\end{table}

Table~\ref{tab:multiseed_mtil} reports the results of KeepLoRA and KeepLoRA++ across three independent random seeds, together with their mean and sample standard deviation. KeepLoRA++ consistently outperforms KeepLoRA across all three aggregate metrics in each run, with relatively small variations across seeds. These results further support the robustness of the observed improvements.

\begin{table}[htbp]
\centering
\caption{Comparison of KeepLoRA and KeepLoRA++ across three random seeds on MTIL (\%). We report the mean and sample standard deviation. The best results are in \textbf{bold}.}
\label{tab:multiseed_mtil}
\footnotesize
\setlength{\tabcolsep}{3.5pt}
\renewcommand{\arraystretch}{1.12}
\begin{tabular}{ccccc}
\toprule
Method & Seed & Transfer & Average & Last \\
\midrule
\multirow{4}{*}{\rotatebox[origin=c]{90}{\scriptsize KeepLoRA}}
& 42 & 69.0 & 77.5 & 86.1 \\
& 0 & 68.8 & 77.4 & 86.1 \\
& 2026 & 68.7 & 77.3 & 85.9 \\
& Mean$\pm$Std. & 68.83$\pm$0.15 & 77.40$\pm$0.10 & 86.03$\pm$0.12 \\
\midrule
\multirow{4}{*}{\rotatebox[origin=c]{90}{\scriptsize KeepLoRA++}}
& 42 & 69.5 & 78.0 & 86.6 \\
& 0 & 69.6 & 77.9 & 86.5 \\
& 2026 & 69.6 & 78.0 & 86.7 \\
& Mean$\pm$Std. & \textbf{69.57$\pm$0.06} & \textbf{77.97$\pm$0.06} & \textbf{86.60$\pm$0.10} \\
\bottomrule
\end{tabular}
\end{table}

\subsubsection{Comparison on MTIL with order \MakeUppercase{\romannumeral 2}}
We compare different methods on MTIL in random order: StanfordCars, Food, MNIST, OxfordPet, Flowers, SUN397, Aircraft, Caltech101, DTD, EuroSAT and CIFAR100. As shown in Table~\ref{tab:MTIL_order2}, KeepLoRA++ consistently outperforms previous methods across all aggregate metrics.

\begin{table*}[t!]
\centering
\caption{\textnormal{\textbf{Comparison of different continual learning methods on MTIL for each task with order-\MakeUppercase{\romannumeral 2}} in terms of \emph{Transfer}, \emph{Average}, and \emph{Last} scores (\%). The best results are in \textbf{bold}, and the second-best results are \underline{underlined}.}}
\label{tab:MTIL_order2}

\begin{tabular}
{b{11em}@{\hspace{1pt}}b{1.0em}b{1.0em} *{11}{b{2.0em}}>{\centering\arraybackslash}m{1.8em}@{\hspace{0.8em}}}
\toprule
        \textbf{Method} & \rotatebox{45}{Arch. Kept} & \rotatebox{45}{w/o Extra Data} & \rotatebox{45}{Cars} &  \rotatebox{45}{Food} & \rotatebox{45}{MNIST} & \rotatebox{45}{OxfordPet} & \rotatebox{45}{Flowers} & \rotatebox{45}{Sun397} & \rotatebox{45}{Aircraft} & \rotatebox{45}{Caltech101} & \rotatebox{45}{DTD} & \rotatebox{45}{EuroSAT} & \rotatebox{45}{CIFAR100} & \textbf{Avg.}~~ \\
\midrule
\quad Zero-shot & \checkmark & \checkmark & 64.7 & 88.5 & 59.4 & 89.0 & 71.0 & 65.4 & 24.8 & 88.4 & 44.6 & 54.9 & 68.2 & \\
\hline
\multicolumn{15}{l}{\textbf{Transfer}}\\
\quad LwF {\cite{li2017learning}} & \checkmark & \ding{55} & ~~-- & 87.8 & \textbf{58.5} & 71.9 & 46.6 & 57.3 & 12.8 & 81.4 & 34.5 & 34.5 & 46.8 & 53.2 \\
\quad iCaRL {\cite{rebuffi2017icarl}} & \checkmark & \ding{55} & ~~-- & 86.1 & 51.8 & 67.6 & 50.4 & 57.9 & 11.0 & 72.3 & 31.2 & 32.7 & 48.1 & 50.9 \\
\quad LwF-VR {\cite{ding2022don}} & \checkmark & \ding{55} & ~~-- & 88.2 & 57.0 & 71.4 & 50.0 & 58.0 & 13.0 & 82.0 & 34.4 & 29.3 & 47.6 & 53.1 \\
\quad WiSE-FT {\cite{wortsman2022robust}} & \checkmark & \ding{55} & ~~-- & 87.2 & \underline{57.6} & 67.0 & 45.0 & 54.0 & 12.9 & 78.6 & 35.5 & 28.4 & 44.3 & 51.1 \\
\quad ZSCL {\cite{zheng2023preventing}} & \checkmark & \ding{55} & ~~-- & \underline{88.3} & 57.5 & 84.7 & 68.1 & 64.8 & 21.1 & \underline{88.2} &  \textbf{45.3} & \textbf{55.2} & \underline{68.2} &  \underline{64.1} \\
\quad O-LoRA {\cite{wang2023orthogonal}} & \checkmark & \checkmark & ~~-- & 87.8 & 56.7 & 90.1 & \underline{71.4} & 64.0 & 20.7 & 87.4 & \underline{43.9} & 46.3 & 65.9 & 63.4 \\
\quad InfLoRA {\cite{liang2024inflora}} & \checkmark & \checkmark & ~~-- & 88.2 & 56.7 & 90.2 &  71.3 &  \underline{65.0} &  \underline{22.2} & \underline{88.2} & 43.8 & 47.3 & 67.2 & 64.0 \\
\quad SD-LoRA {\cite{wu2025sd}} & \checkmark & \checkmark & ~~-- & 88.0 & 56.4 &  \underline{90.5} & 71.0 & 64.6 & 22.0 & 87.8 & 43.7 & 47.1 & 66.4 & 63.7 \\
\rowcolor{gray!15}
\quad KeepLoRA++ & \checkmark & \checkmark & ~~-- & \textbf{88.9} & 57.1 & \textbf{91.3} & \textbf{72.3} & \textbf{66.3} & \textbf{23.8} & \textbf{89.1} & \textbf{45.3} & \underline{49.1} & \textbf{70.0} & \textbf{65.3} \\
\cmidrule{2-15}
\quad L2P {\cite{wang2022learning}} & \ding{55} & \checkmark & ~~-- & 70.6 & 30.7 & 78.3 & 42.8 & 38.3 & 17.4 & 75.3 & 27.4 & 23.1 & 20.7 & 42.5 \\
\quad DualPrompt {\cite{wang2022dualprompt}} & \ding{55} & \checkmark & ~~-- & 79.9 & 46.9 & 85.2 & 51.3 & 45.1 & 9.3 & 82.7 & 29.9 & 42.9 & 47.2 & 52.1 \\
\quad S-Prompts {\cite{wang2022s}} & \ding{55} & \checkmark & ~~-- & 59.8 & 46.2 & 67.7 & 47.5 & 43.8 & 13.5 & 76.8 & 31.4 & 22.6 & 43.5 & 45.3 \\
\quad DIKI {\cite{tang2024mind}} & \ding{55} & \checkmark & ~~-- & 85.8 & \textbf{59.8} & \underline{89.1} & \underline{71.8} & 62.6 & \underline{24.3} & \underline{93.3} & 42.7 & 46.8 & 67.8 & 64.4 \\
\quad MoE-Adapters {\cite{yu2024boosting}} & \ding{55} & \ding{55} & ~~-- & \underline{88.8} & \underline{59.5} & \underline{89.1} & 69.9 & \underline{64.4} & 18.1 & 86.9 & 43.7 & \textbf{54.6} & 68.2 & 64.3 \\
\quad IAP {\cite{fu2026iap}} & \ding{55} & \checkmark & ~~-- & 85.7 & 59.4 & \underline{89.1} & 71.3 & 62.7 & \textbf{24.4} & \textbf{94.0} & \underline{43.8} & 49.0 & \underline{68.6} & \underline{64.9} \\
\rowcolor{gray!15}
\quad KeepLoRA++$_\text{cls}$ & \ding{55} & \checkmark & ~~-- & \textbf{89.3} & 59.0 & \textbf{90.7} & \textbf{72.5} & \textbf{66.2} & 24.0 & 88.9 & \textbf{44.1} & \underline{53.5} & \textbf{70.8} & \textbf{65.9} \\
\hline
\multicolumn{15}{l}{\textbf{Average}}\\
\quad LwF {\cite{li2017learning}} & \checkmark & \ding{55} & 49.0 & 77.0 & \textbf{92.1} & 85.9 & 66.5 & 67.2 & 20.9 & 84.7 & 44.6 & 45.5 & 50.5 & 62.2 \\
\quad iCaRL {\cite{rebuffi2017icarl}} & \checkmark & \ding{55} & 52.0 & 75.9 & 77.4 & 74.6 & 58.4 & 59.3 & 11.7 & 79.6 & 42.1 & 43.2 & 51.7 & 56.9 \\
\quad LwF-VR {\cite{ding2022don}} & \checkmark & \ding{55} & 44.9 & 75.8 & 91.8 & 85.3 & 63.5 & 67.6 & 16.9 & 84.9 & 44.0 & 40.6 & 51.3 & 60.6 \\
\quad WiSE-FT {\cite{wortsman2022robust}} & \checkmark & \ding{55} & 52.6 & 79.3 & \underline{91.9} & 83.9 & 63.4 & 65.2 & 23.3 & 83.7 & 45.4 & 40.0 & 48.2 & 61.5 \\
\quad ZSCL {\cite{zheng2023preventing}} & \checkmark & \ding{55} & 81.7 & 91.3 & \underline{91.9} & 91.0 & \underline{82.9} & 72.5 & 33.6 & 89.7 & \textbf{53.3} & \textbf{62.8} & \underline{69.9} & \underline{74.6} \\
\quad O-LoRA {\cite{wang2023orthogonal}} & \checkmark & \checkmark & 78.5 & 91.0 & 91.3 & 92.3 & 77.7 & 73.0 & 33.5 & 90.5 & 50.7 & 55.1 & 67.8 & 72.9 \\
\quad InfLoRA {\cite{liang2024inflora}} & \checkmark & \checkmark & \underline{84.0} & \underline{92.1} & 91.7 & \underline{93.2} & 81.6 & \underline{74.3} & \underline{34.3} & \underline{91.3} & 51.5 & 56.6 & 69.0 & 74.5 \\
\quad SD-LoRA {\cite{wu2025sd}} & \checkmark & \checkmark & 76.8 & 91.1 & 90.8 & 92.5 & 76.5 & 73.1 & 34.0 & 90.7 & 49.1 & 56.2 & 68.2 & 72.6 \\
\rowcolor{gray!15}
\quad KeepLoRA++ & \checkmark & \checkmark & \textbf{85.7} & \textbf{92.4} & \underline{91.9} & \textbf{93.6} & \textbf{84.7} & \textbf{75.0} & \textbf{36.4} & \textbf{91.9} & \underline{53.1} & \underline{58.0} & \textbf{71.6} & \textbf{75.9} \\
\cmidrule{2-15}
\quad L2P {\cite{wang2022learning}} & \ding{55} & \checkmark & 80.1 & 87.4 & 86.7 & 89.6 & 76.8 & 59.1 & 27.7 & 79.5 & 39.9 & 34.6 & 26.5 & 62.5 \\
\quad DualPrompt {\cite{wang2022dualprompt}} & \ding{55} & \checkmark & 78.6 & 88.4 & 89.7 & 91.7 & 80.0 & 62.4 & 23.2 & 85.0 & 41.3 & 51.6 & 50.7 & 67.5 \\
\quad S-Prompts {\cite{wang2022s}} & \ding{55} & \checkmark & 79.2 & 86.5 & 89.5 & 87.0 & 78.2 & 61.5 & 25.5 & 83.6 & 41.9 & 36.3 & 47.2 & 65.1 \\
\quad DIKI {\cite{tang2024mind}} & \ding{55} & \checkmark & 81.9 & 88.9 & \underline{92.1} & 92.8 & \underline{87.7} & 70.3 & \underline{34.3} & \underline{94.2} & 51.5 & 56.1 & 69.5 & 74.5 \\
\quad MoE-Adapters {\cite{yu2024boosting}} & \ding{55} & \ding{55} & \underline{84.9} & \underline{89.9} & 89.3 & 91.4 & 86.2 & \underline{72.2} & 33.4 & 89.4 & \textbf{53.3} & \underline{61.4} & 69.9 & 74.7 \\
\quad IAP {\cite{fu2026iap}} & \ding{55} & \checkmark & 82.5 & 89.2 & \textbf{92.3} & \underline{93.2} & \textbf{88.0} & 70.4 & \underline{34.3} & \textbf{94.4} & 52.4 & 57.9 & \underline{70.2} & \underline{75.1} \\
\rowcolor{gray!15}
\quad KeepLoRA++$_\text{cls}$ & \ding{55} & \checkmark & \textbf{88.1} & \textbf{92.4} & 92.0 & \textbf{93.9} & 87.5 & \textbf{75.5} & \textbf{38.8} & 91.9 & \underline{53.0} & \textbf{61.7} & \textbf{72.4} & \textbf{77.0} \\
\hline
\multicolumn{15}{l}{\textbf{Last}}\\
\quad LwF {\cite{li2017learning}} & \checkmark & \ding{55} & 34.6 & 69.6 & 99.3 & 88.7 & 61.1 & 72.5 & 32.5 & 88.1 & 65.6 & 90.9 & \underline{87.9} & 71.9 \\
\quad iCaRL {\cite{rebuffi2017icarl}} & \checkmark & \ding{55} & 46.0 & 81.5 & 91.3 & 82.8 & 66.5 & 72.2 & 16.3 & 91.6 & 68.1 & 83.2 & 87.8 & 71.6 \\
\quad LwF-VR {\cite{ding2022don}} & \checkmark & \ding{55} & 27.4 & 61.2 & \underline{99.4} & 86.3 & 60.6 & 70.7 & 23.4 & 88.0 & 61.3 & 84.3 & \textbf{88.1} & 68.2 \\
\quad WiSE-FT {\cite{wortsman2022robust}} & \checkmark & \ding{55} & 35.6 & 76.9 & \textbf{99.5} & 89.1 & 62.1 & 71.8 & 27.8 & 90.8 & 67.0 & 85.6 & 87.6 & 72.2 \\
\quad ZSCL {\cite{zheng2023preventing}} & \checkmark & \ding{55} & 78.2 & 91.1 & 97.6 & 92.5 & \underline{87.4} & 78.2 & 45.0 & 92.3 & \underline{72.7} & 96.2 & 86.3 & 83.4 \\
\quad O-LoRA {\cite{wang2023orthogonal}} & \checkmark & \checkmark & 70.3 & 89.8 & 97.8 & 92.9 & 73.8 & 79.8 & 44.4 & \underline{95.3} & 66.3 & 91.5 & 85.9 & 80.7 \\
\quad InfLoRA {\cite{liang2024inflora}} & \checkmark & \checkmark & \underline{82.4} & \underline{92.0} & 99.3 & \underline{93.9} & 85.4 & \underline{81.2} & \underline{46.1} & \textbf{96.5} & 70.0 & \underline{97.6} & 87.2 & \underline{84.7} \\
\quad SD-LoRA {\cite{wu2025sd}} & \checkmark & \checkmark & 72.3 & 89.7 & 97.3 & 92.4 & 76.1 & 78.9 & 45.3 & 95.2 & 61.6 & 96.9 & 86.1 & 81.1 \\
\rowcolor{gray!15}
\quad KeepLoRA++ & \checkmark & \checkmark & \textbf{84.3} & \textbf{92.4} & \textbf{99.5} & \textbf{94.0} & \textbf{90.6} & \textbf{81.6} & \textbf{49.3} & \textbf{96.5} & \textbf{72.9} & \textbf{97.9} & 87.3 & \textbf{86.0} \\
\cmidrule{2-15}
\quad L2P {\cite{wang2022learning}} & \ding{55} & \checkmark & 80.1 & 89.1 & 99.1 & 93.8 & 96.2 & 76.5 & 40.1 & 86.9 & 73.5 & 86.3 & 84.2 & 82.3 \\
\quad DualPrompt {\cite{wang2022dualprompt}} & \ding{55} & \checkmark & 78.6 & \underline{89.3} & \underline{99.2} & 94.1 & 96.5 & 76.8 & 39.8 & 89.0 & 71.6 & 90.7 & 84.9 & 82.8 \\
\quad S-Prompts {\cite{wang2022s}} & \ding{55} & \checkmark & 79.2 & 89.1 & 99.1 & 94.3 & 95.8 & 76.3 & 39.9 & 95.5 & 70.1 & 97.6 & 84.4 & 83.8 \\
\quad DIKI {\cite{tang2024mind}} & \ding{55} & \checkmark & 81.9 & 89.2 & \textbf{99.4} & 94.3 & \underline{96.8} & 76.7 & 46.3 & 95.9 & 74.8 & \textbf{98.3} & \underline{86.6} & 85.5 \\
\quad MoE-Adapters {\cite{yu2024boosting}} & \ding{55} & \ding{55} & \underline{84.1} & 88.5 & 94.0 & 91.8 & 94.1 & \underline{77.8} & \underline{50.4} & 93.3 & \textbf{77.1} & 87.7 & \underline{86.6} & 84.1 \\
\quad IAP {\cite{fu2026iap}} & \ding{55} & \checkmark & 82.5 & 88.6 & \textbf{99.4} & \underline{94.9} & \textbf{97.7} & 76.9 & 46.1 & \underline{96.1} & 74.7 & \underline{98.0} & \underline{86.6} & \underline{85.9} \\
\rowcolor{gray!15}
\quad KeepLoRA++$_\text{cls}$ & \ding{55} & \checkmark & \textbf{87.8} & \textbf{92.5} & \underline{99.2} & \textbf{95.0} & 96.0 & \textbf{83.0} & \textbf{56.0} & \textbf{97.3} & \underline{76.7} & \underline{98.0} & \textbf{87.8} & \textbf{88.1} \\
\bottomrule
\end{tabular}
\vspace{-0.1in}
\end{table*}

\subsubsection{Per-Training-Step Results}
We present the detailed per-training-step accuracies through all training steps in Table~\ref{tab:keeplorajjMTIL1}, \ref{tab:keeploraMTIL1},  \ref{tab:keeplorajjMTIL2}, \ref{tab:keeploraMTIL2}, \ref{tab:methods_dcl}, \ref{tab:methods_ucit}, and \ref{tab:methods_clvista}. These results demonstrate strong performance in terms of both learning plasticity and stability.

\begin{table*}[t!]
\centering
\caption{\textnormal{\textbf{Accuracy of KeepLoRA++ on the MTIL benchmark with order-\MakeUppercase{\romannumeral 1}.} Each row represents the performance on every dataset of the model trained after the corresponding task. \colorbox{transfer}{Transfer}, \colorbox{average}{Average}, and \colorbox{last}{Last} metrics (\%) are shown.}}
\label{tab:keeplorajjMTIL1}
\begin{tabular}{b{6em}*{11}{>{\centering\arraybackslash}m{2.0em}}>{\centering\arraybackslash}m{1.8em}}
\toprule
 & \rotatebox{45}{\textbf{Aircraft}} 
 & \rotatebox{45}{\textbf{Caltech101}} 
 & \rotatebox{45}{\textbf{CIFAR100}} 
 & \rotatebox{45}{\textbf{DTD}} 
 & \rotatebox{45}{\textbf{EuroSAT}} 
 & \rotatebox{45}{\textbf{Flowers}} 
 & \rotatebox{45}{\textbf{Food}} 
 & \rotatebox{45}{\textbf{MNIST}} 
 & \rotatebox{45}{\textbf{OxfordPet}} 
 & \rotatebox{45}{\textbf{Cars}} 
 & \rotatebox{45}{\textbf{Sun397}} 
 & \\
\midrule
\quad Transfer    &      & 85.5 & 69.5 & 45.8 & 55.1 & 70.7 & 88.1 & 64.9 & 90.6 & 60.0 & 65.1 & {\cellcolor{transfer}}69.5 \\
\midrule
\quad Aircraft    & {\cellcolor{diag}}57.7 & 85.5 & 68.8 & 45.6 & 53.8 & 72.0 & 89.1 & 62.7 & 91.3 & 61.0 & 65.1 & \\
\quad Caltech101  & 57.3 & {\cellcolor{diag}}97.1 & 70.1 & 45.1 & 52.6 & 71.4 & 89.1 & 62.2 & 91.5 & 60.8 & 66.1 \\
\quad CIFAR100    & 56.7 & 96.8 & {\cellcolor{diag}}87.6 & 46.6 & 56.7 & 69.9 & 87.9 & 66.9 & 90.2 & 59.8 & 65.6 & \\
\quad DTD         & 56.7 & 96.8 & 87.5 & {\cellcolor{diag}}74.4 & 57.5 & 70.2 & 87.5 & 65.4 & 90.4 & 59.7 & 65.4 & \\
\quad EuroSAT     & 56.6 & 96.8 & 87.4 & 74.7 & {\cellcolor{diag}}98.2 & 69.8 & 87.4 & 64.7 & 90.5 & 59.6 & 65.3 & \\
\quad Flowers     & 56.7 & 97.0 & 87.4 & 74.9 & 98.2 & {\cellcolor{diag}}93.7 & 87.4 & 64.6 & 90.7 & 59.8 & 65.2 & \\
\quad Food        & 56.1 & 97.0 & 87.1 & 74.3 & 98.3 & 92.7 & {\cellcolor{diag}}92.3 & 67.6 & 90.3 & 59.9 & 64.5 & \\
\quad MNIST       & 55.4 & 96.8 & 86.7 & 73.7 & 98.3 & 92.6 & 92.3 & {\cellcolor{diag}}99.5 & 90.1 & 59.9 & 64.7 & \\
\quad OxfordPet   & 55.2 & 96.8 & 86.8 & 73.9 & 98.3 & 92.6 & 92.4 & 99.5 & {\cellcolor{diag}}94.4 & 59.7 & 64.8 & \\
\quad Cars        & 55.0 & 96.7 & 86.7 & 73.9 & 98.2 & 92.4 & 92.3 & 99.5 & 94.5 & {\cellcolor{diag}}85.4 & 64.4 & \\
\quad SUN397      & 54.0 & 96.6 & 86.5 & 72.5 & 98.3 & 91.4 & 92.1 & 99.5 & 94.5 & 84.6 & {\cellcolor{diag}}82.2 & {\cellcolor{last}}86.6 \\
\midrule
\quad Average     & 56.1 & 95.8 & 83.9 & 66.3 & 82.6 & 82.6 & 90.0 & 77.5 & 91.7 & 64.6 & 66.7 & {\cellcolor{average}}78.0 \\
\bottomrule
\end{tabular}
\vspace{-0.1in}
\end{table*}

\begin{table*}[t!]
\centering
\caption{\textnormal{\textbf{Accuracy of KeepLoRA on the MTIL benchmark with order-\MakeUppercase{\romannumeral 1}.} Each row represents the performance on every dataset of the model trained after the corresponding task. \colorbox{transfer}{Transfer}, \colorbox{average}{Average}, and \colorbox{last}{Last} metrics (\%) are shown.}}
\label{tab:keeploraMTIL1}
\begin{tabular}{b{6em}*{11}{>{\centering\arraybackslash}m{2.0em}}>{\centering\arraybackslash}m{1.8em}}
\toprule
 & \rotatebox{45}{\textbf{Aircraft}} 
 & \rotatebox{45}{\textbf{Caltech101}} 
 & \rotatebox{45}{\textbf{CIFAR100}} 
 & \rotatebox{45}{\textbf{DTD}} 
 & \rotatebox{45}{\textbf{EuroSAT}} 
 & \rotatebox{45}{\textbf{Flowers}} 
 & \rotatebox{45}{\textbf{Food}} 
 & \rotatebox{45}{\textbf{MNIST}} 
 & \rotatebox{45}{\textbf{OxfordPet}} 
 & \rotatebox{45}{\textbf{Cars}} 
 & \rotatebox{45}{\textbf{Sun397}} 
 & \\
\midrule
\quad Transfer    &      & 84.6 & 68.7 & 45.9 & 54.3 & 70.1 & 87.7 & 64.8 & 90.3 & 59.5 & 64.1 & {\cellcolor{transfer}}69.0 \\
\midrule
\quad Aircraft    & {\cellcolor{diag}}59.0 & 84.6 & 68.4 & 45.4 & 52.2 & 71.9 & 89.0 & 63.8 & 91.1 & 60.6 & 63.6 & \\
\quad Caltech101  & 58.1 & {\cellcolor{diag}}97.0 & 69.1 & 45.4 & 50.8 & 71.1 & 88.7 & 61.8 & 91.1 & 60.1 & 64.8 & \\
\quad CIFAR100    & 56.0 & 96.8 & {\cellcolor{diag}}87.6 & 46.8 & 56.3 & 68.9 & 87.3 & 66.3 & 90.1 & 59.6 & 64.7 & \\
\quad DTD         & 55.9 & 96.7 & 87.5 & {\cellcolor{diag}}75.0 & 57.9 & 69.6 & 87.1 & 64.7 & 90.3 & 59.5 & 64.6 & \\
\quad EuroSAT     & 55.7 & 96.7 & 87.0 & 74.8 & {\cellcolor{diag}}98.4 & 69.3 & 87.0 & 65.2 & 90.2 & 59.1 & 64.6 & \\
\quad Flowers     & 55.6 & 97.0 & 86.9 & 74.4 & 98.4 & {\cellcolor{diag}}93.3 & 86.9 & 65.0 & 90.3 & 59.4 & 64.3 & \\
\quad Food        & 54.7 & 96.8 & 86.2 & 72.6 & 98.3 & 92.2 & {\cellcolor{diag}}91.8 & 66.7 & 89.8 & 59.0 & 63.8 & \\
\quad MNIST       & 54.3 & 96.7 & 85.8 & 72.4 & 98.1 & 91.8 & 91.8 & {\cellcolor{diag}}99.5 & 89.7 & 59.3 & 63.8 & \\
\quad OxfordPet   & 54.6 & 96.7 & 85.7 & 72.0 & 98.2 & 91.8 & 91.8 & 99.5 & {\cellcolor{diag}}94.7 & 59.2 & 63.8 & \\
\quad Cars        & 54.2 & 96.7 & 85.7 & 71.9 & 98.1 & 91.5 & 91.7 & 99.5 & 94.4 & {\cellcolor{diag}}84.3 & 63.7 & \\
\quad SUN397      & 53.2 & 96.8 & 85.7 & 71.4 & 98.1 & 90.8 & 91.4 & 99.6 & 94.5 & 83.1 & {\cellcolor{diag}}82.0 & {\cellcolor{last}}86.1 \\
\midrule
\quad Average     & 55.6 & 95.7 & 83.2 & 65.6 & 82.2 & 82.0 & 89.5 & 77.4 & 91.5 & 63.9 & 65.8 & {\cellcolor{average}}77.5 \\
\bottomrule
\end{tabular}
\vspace{-0.1in}
\end{table*}

\begin{table*}[t!]
\centering
\caption{\textnormal{\textbf{Accuracy of KeepLoRA++ on the MTIL benchmark with order-\MakeUppercase{\romannumeral 2}.} Each row represents the performance on every dataset of the model trained after the corresponding task. \colorbox{transfer}{Transfer}, \colorbox{average}{Average}, and \colorbox{last}{Last} metrics (\%) are shown.}}
\label{tab:keeplorajjMTIL2}
\begin{tabular}{b{6em}*{11}{>{\centering\arraybackslash}m{2.0em}}>{\centering\arraybackslash}m{1.8em}}
\toprule
 & \rotatebox{45}{\textbf{Cars}} 
 & \rotatebox{45}{\textbf{Food}} 
 & \rotatebox{45}{\textbf{MNIST}} 
 & \rotatebox{45}{\textbf{OxfordPet}} 
 & \rotatebox{45}{\textbf{Flowers}} 
 & \rotatebox{45}{\textbf{Sun397}} 
 & \rotatebox{45}{\textbf{Aircraft}} 
 & \rotatebox{45}{\textbf{Caltech101}} 
 & \rotatebox{45}{\textbf{DTD}} 
 & \rotatebox{45}{\textbf{EuroSAT}} 
 & \rotatebox{45}{\textbf{CIFAR100}} 
 & \\
\midrule
\quad Transfer   &      & 88.9 & 57.1 & 91.3 & 72.3 & 66.3 & 23.8 & 89.1 & 45.3 & 49.1 & 70.0 & {\cellcolor{transfer}}65.3 \\
\midrule
\quad Cars       & {\cellcolor{diag}}86.3 & 88.9 & 56.9 & 91.4 & 71.6 & 66.3 & 23.7 & 87.3 & 46.1 & 52.6 & 69.6 & \\
\quad Food       & 86.4 & {\cellcolor{diag}}92.9 & 57.4 & 91.3 & 72.6 & 66.4 & 23.6 & 88.5 & 45.8 & 50.2 & 70.7 & \\
\quad MNIST      & 86.3 & 92.9 & {\cellcolor{diag}}99.6 & 91.3 & 72.4 & 66.3 & 23.7 & 89.0 & 45.6 & 50.2 & 69.8 & \\
\quad OxfordPet  & 86.2 & 92.9 & 99.6 & {\cellcolor{diag}}94.6 & 72.8 & 66.3 & 23.8 & 89.4 & 45.5 & 48.6 & 69.8 & \\
\quad Flowers    & 86.1 & 92.9 & 99.6 & 94.7 & {\cellcolor{diag}}92.4 & 66.1 & 23.7 & 89.4 & 45.7 & 47.9 & 69.7 & \\
\quad Sun397     & 85.7 & 92.8 & 99.6 & 94.6 & 92.1 & {\cellcolor{diag}}82.6 & 24.2 & 90.2 & 45.2 & 47.1 & 70.1 & \\
\quad Aircraft   & 85.5 & 92.8 & 99.6 & 94.6 & 92.0 & 82.6 & {\cellcolor{diag}}52.6 & 90.2 & 44.6 & 46.8 & 70.1 & \\
\quad Caltech101 & 85.5 & 92.7 & 99.6 & 94.4 & 92.0 & 82.5 & 52.1 & {\cellcolor{diag}}96.8 & 44.0 & 49.2 & 70.2 & \\
\quad DTD        & 85.5 & 92.7 & 99.6 & 94.6 & 91.8 & 82.5 & 52.0 & 97.0 & {\cellcolor{diag}}74.5 & 49.6 & 70.2 & \\
\quad EuroSAT    & 85.3 & 92.7 & 99.6 & 94.5 & 91.6 & 82.3 & 51.8 & 97.1 & 74.0 & {\cellcolor{diag}}98.5 & 70.0 & \\
\quad CIFAR100   & 84.3 & 92.4 & 99.5 & 94.0 & 90.6 & 81.6 & 49.3 & 96.5 & 72.9 & 97.9 & {\cellcolor{diag}}87.3 & {\cellcolor{last}}86.0 \\
\midrule
\quad Average    & 85.7 & 92.4 & 91.9 & 93.6 & 84.7 & 75.0 & 36.4 & 91.9 & 53.1 & 58.0 & 71.6 & {\cellcolor{average}}75.9 \\
\midrule
\bottomrule
\end{tabular}
\vspace{-0.1in}
\end{table*}

\begin{table*}[t!]
\centering
\caption{\textnormal{\textbf{Accuracy of KeepLoRA on the MTIL benchmark with order-\MakeUppercase{\romannumeral 2}.} Each row represents the performance on every dataset of the model trained after the corresponding task. \colorbox{transfer}{Transfer}, \colorbox{average}{Average}, and \colorbox{last}{Last} metrics (\%) are shown.}}
\label{tab:keeploraMTIL2}
\begin{tabular}{b{6em}*{11}{>{\centering\arraybackslash}m{2.0em}}>{\centering\arraybackslash}m{1.8em}}
\toprule
 & \rotatebox{45}{\textbf{Cars}} 
 & \rotatebox{45}{\textbf{Food}} 
 & \rotatebox{45}{\textbf{MNIST}} 
 & \rotatebox{45}{\textbf{OxfordPet}} 
 & \rotatebox{45}{\textbf{Flowers}} 
 & \rotatebox{45}{\textbf{Sun397}} 
 & \rotatebox{45}{\textbf{Aircraft}} 
 & \rotatebox{45}{\textbf{Caltech101}} 
 & \rotatebox{45}{\textbf{DTD}} 
 & \rotatebox{45}{\textbf{EuroSAT}} 
 & \rotatebox{45}{\textbf{CIFAR100}} 
 & \\
\midrule
\quad Transfer   &      & 88.7 & 57.7 & 91.2 & 72.1 & 65.8 & 23.4 & 88.8 & 45.4 & 48.5 & 68.2 & {\cellcolor{transfer}}65.0 \\
\midrule
\quad Cars       & {\cellcolor{diag}}86.2 & 88.7 & 57.1 & 91.3 & 71.7 & 65.5 & 23.5 & 87.4 & 46.6 & 50.7 & 69.5 & \\
\quad Food       & 85.9 & {\cellcolor{diag}}92.9 & 58.3 & 91.1 & 72.3 & 66.0 & 23.9 & 88.3 & 45.3 & 49.8 & 70.5 & \\
\quad MNIST      & 85.8 & 92.8 & {\cellcolor{diag}}99.6 & 91.2 & 71.9 & 66.2 & 23.0 & 88.6 & 46.4 & 50.4 & 68.1 & \\
\quad OxfordPet  & 85.7 & 92.8 & 99.6 & {\cellcolor{diag}}94.8 & 72.4 & 65.9 & 23.0 & 89.3 & 46.0 & 48.2 & 67.8 & \\
\quad Flowers    & 85.6 & 92.8 & 99.6 & 94.8 & {\cellcolor{diag}}92.4 & 65.7 & 23.0 & 89.3 & 46.2 & 46.9 & 67.5 & \\
\quad Sun397     & 85.2 & 92.7 & 99.6 & 94.6 & 92.2 & {\cellcolor{diag}}82.7 & 24.0 & 89.6 & 44.2 & 47.0 & 68.0 & \\
\quad Aircraft   & 84.8 & 92.7 & 99.6 & 94.6 & 92.1 & 82.7 & {\cellcolor{diag}}51.6 & 89.3 & 44.2 & 46.3 & 68.0 & \\
\quad Caltech101 & 84.8 & 92.7 & 99.6 & 94.6 & 92.2 & 82.6 & 51.6 & {\cellcolor{diag}}97.1 & 44.5 & 48.7 & 68.3 & \\
\quad DTD        & 84.8 & 92.6 & 99.6 & 94.8 & 92.2 & 82.6 & 51.3 & 96.9 & {\cellcolor{diag}}74.5 & 48.2 & 68.2 & \\
\quad EuroSAT    & 84.6 & 92.7 & 99.6 & 94.6 & 92.1 & 82.2 & 51.1 & 97.0 & 74.5 & {\cellcolor{diag}}98.6 & 66.6 & \\
\quad CIFAR100   & 83.7 & 92.3 & 99.5 & 94.4 & 90.8 & 81.3 & 49.0 & 96.9 & 72.3 & 98.0 & {\cellcolor{diag}}87.3 & {\cellcolor{last}}85.9 \\
\midrule
\quad Average    & 85.2 & 92.3 & 92.0 & 93.7 & 84.8 & 74.8 & 35.9 & 91.8 & 53.1 & 57.5 & 70.0 & {\cellcolor{average}}75.6 \\
\bottomrule
\end{tabular}
\vspace{-0.1in}
\end{table*}

\begin{table*}[t!]
\centering
\footnotesize
\caption{\textnormal{\textbf{Accuracy of LoRA-FT, O-LoRA, CL-MoE, RegLoRA, KeepLoRA, and KeepLoRA++ on the MLLM-DCL benchmark.} Each row represents the performance on every dataset of the model trained after the corresponding task. \colorbox{transfer}{Transfer}, \colorbox{average}{Average}, and \colorbox{last}{Last} metrics (\%) are shown.}}
\label{tab:methods_dcl}

\vspace{0.5em}
\renewcommand{\arraystretch}{1.05}

\begin{minipage}[t]{0.48\textwidth}
\centering
\textbf{LoRA-FT}\\[0.25em]
\begin{tabular}{b{4em}*{5}{>{\centering\arraybackslash}m{2em}}>{\centering\arraybackslash}m{2em}}
\toprule
 & \rotatebox{45}{Sensing} & \rotatebox{45}{Medical} & \rotatebox{45}{Driving} & \rotatebox{45}{Science} & \rotatebox{45}{Finance} & \textbf{Avg.} \\
\midrule
Transfer & -- & 28.1 & 17.4 & 34.0 & 50.2 & {\cellcolor{transfer}}32.4 \\
\midrule
Sensing & {\cellcolor{diag}}78.8 & 28.1 & 17.3 & 34.8 & 55.6 &  \\
Medical & 75.5 & {\cellcolor{diag}}58.4 & 17.5 & 32.7 & 54.8 &  \\
Driving & 70.0 & 47.5 & {\cellcolor{diag}}52.3 & 34.6 & 40.9 &  \\
Science & 73.2 & 46.4 & 40.6 & {\cellcolor{diag}}50.4 & 49.5 &  \\
Finance & 69.3 & 44.3 & 29.1 & 41.4 & {\cellcolor{diag}}88.4 & {\cellcolor{last}}54.5 \\
\midrule
Average & 73.3 & 44.9 & 31.4 & 38.8 & 57.8 & {\cellcolor{average}}49.3 \\
\bottomrule
\end{tabular}
\end{minipage}
\hfill
\begin{minipage}[t]{0.48\textwidth}
\centering
\textbf{O-LoRA}\\[0.25em]
\begin{tabular}{b{4em}*{5}{>{\centering\arraybackslash}m{2em}}>{\centering\arraybackslash}m{2em}}
\toprule
 & \rotatebox{45}{Sensing} & \rotatebox{45}{Medical} & \rotatebox{45}{Driving} & \rotatebox{45}{Science} & \rotatebox{45}{Finance} & \textbf{Avg.} \\
\midrule
Transfer & -- & 28.4 & 18.4 & 33.7 & 52.5 & {\cellcolor{transfer}}33.3 \\
\midrule
Sensing & {\cellcolor{diag}}79.4 & 28.4 & 17.6 & 34.9 & 56.1 &  \\
Medical & 74.3 & {\cellcolor{diag}}58.5 & 19.2 & 33.2 & 56.0 &  \\
Driving & 74.7 & 48.3 & {\cellcolor{diag}}52.6 & 33.1 & 45.2 &  \\
Science & 74.6 & 46.5 & 42.2 & {\cellcolor{diag}}50.1 & 52.8 &  \\
Finance & 72.3 & 46.9 & 31.6 & 41.5 & {\cellcolor{diag}}88.1 & {\cellcolor{last}}56.1 \\
\midrule
Average & 75.0 & 45.7 & 32.6 & 38.5 & 59.6 & {\cellcolor{average}}50.3 \\
\bottomrule
\end{tabular}
\end{minipage}

\vspace{0.8em}

\begin{minipage}[t]{0.48\textwidth}
\centering
\textbf{CL-MoE}\\[0.25em]
\begin{tabular}{b{4em}*{5}{>{\centering\arraybackslash}m{2em}}>{\centering\arraybackslash}m{2em}}
\toprule
 & \rotatebox{45}{Sensing} & \rotatebox{45}{Medical} & \rotatebox{45}{Driving} & \rotatebox{45}{Science} & \rotatebox{45}{Finance} & \textbf{Avg.} \\
\midrule
Transfer & -- & 28.3 & 19.4 & 34.1 & 48.6 & {\cellcolor{transfer}}32.6 \\
\midrule
Sensing & {\cellcolor{diag}}79.4 & 28.3 & 18.7 & 35.2 & 56.4 &  \\
Medical & 74.8 & {\cellcolor{diag}}60.7 & 20.1 & 32.4 & 54.9 &  \\
Driving & 74.0 & 44.3 & {\cellcolor{diag}}52.1 & 34.7 & 39.6 &  \\
Science & 71.0 & 47.4 & 40.0 & {\cellcolor{diag}}50.7 & 43.3 &  \\
Finance & 71.8 & 47.4 & 29.5 & 41.5 & {\cellcolor{diag}}89.2 & {\cellcolor{last}}55.9 \\
\midrule
Average & 74.2 & 45.6 & 32.1 & 38.9 & 56.7 & {\cellcolor{average}}49.5 \\
\bottomrule
\end{tabular}
\end{minipage}
\hfill
\begin{minipage}[t]{0.48\textwidth}
\centering
\textbf{RegLoRA}\\[0.25em]
\begin{tabular}{b{4em}*{5}{>{\centering\arraybackslash}m{2em}}>{\centering\arraybackslash}m{2em}}
\toprule
 & \rotatebox{45}{Sensing} & \rotatebox{45}{Medical} & \rotatebox{45}{Driving} & \rotatebox{45}{Science} & \rotatebox{45}{Finance} & \textbf{Avg.} \\
\midrule
Transfer & -- & 28.1 & 19.6 & 33.9 & 52.4 & {\cellcolor{transfer}}33.5 \\
\midrule
Sensing & {\cellcolor{diag}}78.8 & 28.1 & 18.6 & 35.1 & 56.2 &  \\
Medical & 77.1 & {\cellcolor{diag}}59.5 & 20.7 & 33.0 & 55.7 &  \\
Driving & 77.8 & 51.6 & {\cellcolor{diag}}52.5 & 33.5 & 47.4 &  \\
Science & 77.9 & 48.4 & 44.7 & {\cellcolor{diag}}50.4 & 50.1 &  \\
Finance & 77.1 & 50.9 & 40.3 & 43.0 & {\cellcolor{diag}}88.4 & {\cellcolor{last}}59.9 \\
\midrule
Average & 77.7 & 47.7 & 35.4 & 39.0 & 59.6 & {\cellcolor{average}}51.9 \\
\bottomrule
\end{tabular}
\end{minipage}

\vspace{0.8em}

\begin{minipage}[t]{0.48\textwidth}
\centering
\textbf{KeepLoRA}\\[0.25em]
\begin{tabular}{b{4em}*{5}{>{\centering\arraybackslash}m{2em}}>{\centering\arraybackslash}m{2em}}
\toprule
 & \rotatebox{45}{Sensing} & \rotatebox{45}{Medical} & \rotatebox{45}{Driving} & \rotatebox{45}{Science} & \rotatebox{45}{Finance} & \textbf{Avg.} \\
\midrule
Transfer & -- & 28.5 & 16.6 & 34.1 & 55.6 & {\cellcolor{transfer}}33.7 \\
\midrule
Sensing & {\cellcolor{diag}}80.0 & 28.5 & 17.0 & 35.1 & 55.1 &  \\
Medical & 79.9 & {\cellcolor{diag}}58.6 & 16.3 & 33.7 & 55.6 &  \\
Driving & 79.8 & 57.7 & {\cellcolor{diag}}53.1 & 33.7 & 54.6 &  \\
Science & 79.2 & 54.9 & 51.1 & {\cellcolor{diag}}51.6 & 57.2 &  \\
Finance & 78.8 & 54.3 & 50.2 & 49.5 & {\cellcolor{diag}}89.3 & {\cellcolor{last}}64.4 \\
\midrule
Average & 79.6 & 50.8 & 37.5 & 40.7 & 62.4 & {\cellcolor{average}}54.2 \\
\bottomrule
\end{tabular}
\end{minipage}
\hfill
\begin{minipage}[t]{0.48\textwidth}
\centering
\textbf{KeepLoRA++}\\[0.25em]
\begin{tabular}{b{4em}*{5}{>{\centering\arraybackslash}m{2em}}>{\centering\arraybackslash}m{2em}}
\toprule
 & \rotatebox{45}{Sensing} & \rotatebox{45}{Medical} & \rotatebox{45}{Driving} & \rotatebox{45}{Science} & \rotatebox{45}{Finance} & \textbf{Avg.} \\
\midrule
Transfer & -- & 29.1 & 15.6 & 35.0 & 58.4 & {\cellcolor{transfer}}34.5 \\
\midrule
Sensing & {\cellcolor{diag}}77.9 & 29.1 & 14.3 & 35.2 & 58.5 &  \\
Medical & 79.6 & {\cellcolor{diag}}60.4 & 16.8 & 34.6 & 56.9 &  \\
Driving & 79.7 & 59.7 & {\cellcolor{diag}}53.4 & 35.1 & 60.2 &  \\
Science & 79.7 & 57.1 & 53.0 & {\cellcolor{diag}}50.5 & 58.0 &  \\
Finance & 79.8 & 57.7 & 51.9 & 50.0 & {\cellcolor{diag}}89.5 & {\cellcolor{last}}65.8 \\
\midrule
Average & 79.3 & 52.8 & 37.9 & 41.1 & 64.6 & {\cellcolor{average}}55.1 \\
\bottomrule
\end{tabular}
\end{minipage}

\vspace{-0.1in}
\end{table*}

\begin{table*}[t!]
\centering
\footnotesize
\caption{\textnormal{\textbf{Accuracy of LoRA-FT, O-LoRA, CL-MoE, RegLoRA, KeepLoRA, and KeepLoRA++ on the UCIT benchmark.} Each row represents the performance on every dataset of the model trained after the corresponding task. \colorbox{transfer}{Transfer}, \colorbox{average}{Average}, and \colorbox{last}{Last} metrics (\%) are shown.}}
\label{tab:methods_ucit}

\vspace{0.5em}
\renewcommand{\arraystretch}{1.05}

\begin{minipage}[t]{0.48\textwidth}
\centering
\textbf{LoRA-FT}\\[0.25em]
\begin{tabular}{b{4.5em}*{6}{>{\centering\arraybackslash}m{2em}}>{\centering\arraybackslash}m{2em}}
\toprule
 & \rotatebox{45}{ImgNet-R} & \rotatebox{45}{ArxivQA} & \rotatebox{45}{VizWiz} & \rotatebox{45}{IconQA} & \rotatebox{45}{CLEVR} & \rotatebox{45}{Flickr30k} & \textbf{Avg.} \\
\midrule
Transfer & -- & 52.6 & 18.3 & 6.0 & 17.0 & 40.3 & {\cellcolor{transfer}}26.8 \\
\midrule
ImgNet-R  & {\cellcolor{diag}}91.7 & 52.6 & 23.5 & 11.8 & 17.2 & 36.5 & \\
ArxivQA   & 90.5 & {\cellcolor{diag}}92.1 & 13.1 & 2.1 & 14.2 & 21.5 & \\
VizWiz    & 73.6 & 90.7 & {\cellcolor{diag}}61.0 & 4.2 & 19.0 & 49.7 & \\
IconQA    & 72.7 & 77.1 & 53.7 & {\cellcolor{diag}}79.7 & 17.4 & 47.8 & \\
CLEVR     & 68.8 & 77.4 & 52.3 & 67.8 & {\cellcolor{diag}}77.9 & 46.1 & \\
Flickr30k & 58.6 & 76.7 & 45.7 & 67.4 & 61.6 & {\cellcolor{diag}}58.0 & {\cellcolor{last}}61.4 \\
\midrule
Average & 76.0 & 77.8 & 41.6 & 38.8 & 34.6 & 43.3 & {\cellcolor{average}}52.0 \\
\bottomrule
\end{tabular}
\end{minipage}
\hfill
\begin{minipage}[t]{0.48\textwidth}
\centering
\textbf{O-LoRA}\\[0.25em]
\begin{tabular}{b{4.5em}*{6}{>{\centering\arraybackslash}m{2em}}>{\centering\arraybackslash}m{2em}}
\toprule
 & \rotatebox{45}{ImgNet-R} & \rotatebox{45}{ArxivQA} & \rotatebox{45}{VizWiz} & \rotatebox{45}{IconQA} & \rotatebox{45}{CLEVR} & \rotatebox{45}{Flickr30k} & \textbf{Avg.} \\
\midrule
Transfer & -- & 52.9 & 19.6 & 4.4 & 16.9 & 41.0 & {\cellcolor{transfer}}27.0 \\
\midrule
ImgNet-R  & {\cellcolor{diag}}91.5 & 52.9 & 24.7 & 13.3 & 17.3 & 36.5 & \\
ArxivQA   & 89.7 & {\cellcolor{diag}}94.2 & 14.5 & 0.0 & 12.9 & 25.0 & \\
VizWiz    & 80.9 & 91.7 & {\cellcolor{diag}}59.8 & 0.0 & 19.6 & 49.0 & \\
IconQA    & 80.2 & 80.3 & 54.5 & {\cellcolor{diag}}75.9 & 17.6 & 48.6 & \\
CLEVR     & 78.1 & 80.4 & 51.6 & 63.2 & {\cellcolor{diag}}72.4 & 46.0 & \\
Flickr30k & 74.2 & 80.9 & 45.3 & 62.9 & 63.8 & {\cellcolor{diag}}57.2 & {\cellcolor{last}}64.1 \\
\midrule
Average & 82.4 & 80.1 & 41.7 & 35.9 & 33.9 & 43.7 & {\cellcolor{average}}53.0 \\
\bottomrule
\end{tabular}
\end{minipage}

\vspace{0.8em}

\begin{minipage}[t]{0.48\textwidth}
\centering
\textbf{CL-MoE}\\[0.25em]
\begin{tabular}{b{4.5em}*{6}{>{\centering\arraybackslash}m{2em}}>{\centering\arraybackslash}m{2em}}
\toprule
 & \rotatebox{45}{ImgNet-R} & \rotatebox{45}{ArxivQA} & \rotatebox{45}{VizWiz} & \rotatebox{45}{IconQA} & \rotatebox{45}{CLEVR} & \rotatebox{45}{Flickr30k} & \textbf{Avg.} \\
\midrule
Transfer & -- & 52.0 & 19.3 & 7.4 & 17.8 & 41.3 & {\cellcolor{transfer}}27.6 \\
\midrule
ImgNet-R  & {\cellcolor{diag}}91.2 & 52.0 & 23.9 & 5.2 & 15.6 & 36.9 & \\
ArxivQA   & 89.2 & {\cellcolor{diag}}92.5 & 14.8 & 10.0 & 15.7 & 26.2 & \\
VizWiz    & 77.2 & 90.7 & {\cellcolor{diag}}60.4 & 6.9 & 20.6 & 49.5 & \\
IconQA    & 79.5 & 76.2 & 51.0 & {\cellcolor{diag}}54.7 & 19.4 & 47.9 & \\
CLEVR     & 76.7 & 75.4 & 48.1 & 52.6 & {\cellcolor{diag}}73.0 & 45.9 & \\
Flickr30k & 67.2 & 75.8 & 44.4 & 52.6 & 54.4 & {\cellcolor{diag}}57.3 & {\cellcolor{last}}58.6 \\
\midrule
Average & 80.2 & 77.1 & 40.4 & 30.3 & 33.1 & 44.0 & {\cellcolor{average}}50.9 \\
\bottomrule
\end{tabular}
\end{minipage}
\hfill
\begin{minipage}[t]{0.48\textwidth}
\centering
\textbf{RegLoRA}\\[0.25em]
\begin{tabular}{b{4.5em}*{6}{>{\centering\arraybackslash}m{2em}}>{\centering\arraybackslash}m{2em}}
\toprule
 & \rotatebox{45}{ImgNet-R} & \rotatebox{45}{ArxivQA} & \rotatebox{45}{VizWiz} & \rotatebox{45}{IconQA} & \rotatebox{45}{CLEVR} & \rotatebox{45}{Flickr30k} & \textbf{Avg.} \\
\midrule
Transfer & -- & 53.3 & 18.7 & 7.5 & 17.0 & 40.9 & {\cellcolor{transfer}}27.5 \\
\midrule
ImgNet-R  & {\cellcolor{diag}}91.6 & 53.3 & 23.7 & 12.1 & 16.9 & 36.4 & \\
ArxivQA   & 90.4 & {\cellcolor{diag}}92.8 & 13.7 & 5.0 & 16.4 & 21.1 & \\
VizWiz    & 83.6 & 89.3 & {\cellcolor{diag}}61.4 & 5.3 & 18.6 & 49.8 & \\
IconQA    & 84.3 & 78.1 & 57.4 & {\cellcolor{diag}}79.6 & 16.2 & 50.6 & \\
CLEVR     & 82.8 & 78.6 & 54.2 & 70.6 & {\cellcolor{diag}}75.0 & 46.5 & \\
Flickr30k & 80.2 & 79.1 & 47.1 & 69.4 & 65.7 & {\cellcolor{diag}}57.3 & {\cellcolor{last}}66.5 \\
\midrule
Average & 85.5 & 78.6 & 42.9 & 40.3 & 34.8 & 43.6 & {\cellcolor{average}}54.3 \\
\bottomrule
\end{tabular}
\end{minipage}

\vspace{0.8em}

\begin{center}
\begin{minipage}[t]{0.48\textwidth}
\centering
\textbf{KeepLoRA}\\[0.25em]
\begin{tabular}{b{4.5em}*{6}{>{\centering\arraybackslash}m{2em}}>{\centering\arraybackslash}m{2em}}
\toprule
 & \rotatebox{45}{ImgNet-R} & \rotatebox{45}{ArxivQA} & \rotatebox{45}{VizWiz} & \rotatebox{45}{IconQA} & \rotatebox{45}{CLEVR} & \rotatebox{45}{Flickr30k} & \textbf{Avg.} \\
\midrule
Transfer & -- & 52.8 & 20.4 & 9.2 & 18.1 & 41.5 & {\cellcolor{transfer}}28.4 \\
\midrule
ImgNet-R  & {\cellcolor{diag}}91.5 & 52.8 & 25.6 & 13.4 & 17.1 & 36.7 & \\
ArxivQA   & 90.4 & {\cellcolor{diag}}94.5 & 15.2 & 4.0 & 17.2 & 21.5 & \\
VizWiz    & 85.5 & 92.4 & {\cellcolor{diag}}61.5 & 10.1 & 21.0 & 50.6 & \\
IconQA    & 85.1 & 86.0 & 55.7 & {\cellcolor{diag}}76.9 & 17.1 & 50.9 & \\
CLEVR     & 84.1 & 89.3 & 51.5 & 68.3 & {\cellcolor{diag}}72.6 & 47.8 & \\
Flickr30k & 82.4 & 86.7 & 46.6 & 67.8 & 66.4 & {\cellcolor{diag}}57.2 & {\cellcolor{last}}67.8 \\
\midrule
Average & 86.5 & 83.6 & 42.7 & 40.1 & 35.2 & 44.1 & {\cellcolor{average}}55.4 \\
\bottomrule
\end{tabular}
\end{minipage}
\hfill
\begin{minipage}[t]{0.48\textwidth}
\centering
\textbf{KeepLoRA++}\\[0.25em]
\begin{tabular}{b{4.5em}*{6}{>{\centering\arraybackslash}m{2em}}>{\centering\arraybackslash}m{2em}}
\toprule
 & \rotatebox{45}{ImgNet-R} & \rotatebox{45}{ArxivQA} & \rotatebox{45}{VizWiz} & \rotatebox{45}{IconQA} & \rotatebox{45}{CLEVR} & \rotatebox{45}{Flickr30k} & \textbf{Avg.} \\
\midrule
Transfer & -- & 53.3 & 19.5 & 12.1 & 16.9 & 40.2 & {\cellcolor{transfer}}28.4 \\
\midrule
ImgNet-R  & {\cellcolor{diag}}91.5 & 53.3 & 24.3 & 19.2 & 17.9 & 32.0 & \\
ArxivQA   & 90.8 & {\cellcolor{diag}}93.3 & 14.7 & 6.3 & 14.9 & 20.8 & \\
VizWiz    & 84.8 & 93.3 & {\cellcolor{diag}}60.8 & 11.0 & 17.5 & 50.6 & \\
IconQA    & 84.1 & 91.7 & 56.8 & {\cellcolor{diag}}78.4 & 17.3 & 50.6 & \\
CLEVR     & 84.0 & 93.8 & 49.4 & 66.2 & {\cellcolor{diag}}75.0 & 47.1 & \\
Flickr30k & 81.2 & 94.4 & 46.1 & 66.5 & 69.6 & {\cellcolor{diag}}56.6 & {\cellcolor{last}}69.1 \\
\midrule
Average & 86.1 & 86.6 & 42.0 & 41.3 & 35.4 & 42.9 & {\cellcolor{average}}55.7 \\
\bottomrule
\end{tabular}
\end{minipage}
\end{center}

\vspace{-0.1in}
\end{table*}

\begin{table*}[t!]
\centering
\footnotesize
\caption{\textnormal{\textbf{Accuracy of KeepLoRA++ on the CL-VISTA benchmark.} Each row represents the performance on every dataset of the model trained after the corresponding task. \colorbox{transfer}{Transfer}, \colorbox{average}{Average}, and \colorbox{last}{Last} metrics (\%) are shown.}}
\label{tab:methods_clvista}

\vspace{0.5em}
\renewcommand{\arraystretch}{1.05}

\begin{minipage}[t]{\textwidth}
\centering
\textbf{KeepLoRA++}\\[0.25em]
\begin{tabular}{b{4.5em}*{8}{>{\centering\arraybackslash}m{2em}}>{\centering\arraybackslash}m{2em}}
\toprule
 & \rotatebox{45}{Count.} & \rotatebox{45}{Space} & \rotatebox{45}{Traffic} & \rotatebox{45}{Movie} & \rotatebox{45}{GUI} & \rotatebox{45}{Science} & \rotatebox{45}{Sports} & \rotatebox{45}{Reason.} & \textbf{Avg.} \\
\midrule
Transfer & -- & 42.32 & 51.46 & 37.56 & 64.97 & 61.60 & 52.14 & 55.97 &  {\cellcolor{transfer}}52.29 \\
\midrule
Count.    & {\cellcolor{diag}}59.04 & 42.32 & 53.24 & 37.58 & 63.25 & 60.23 & 53.10 & 58.12 \\
Space     & 57.08 & {\cellcolor{diag}}70.21 & 49.67 & 30.86 & 64.89 & 57.75 & 51.42 & 58.80 \\
Traffic   & 55.91 & 66.44 & {\cellcolor{diag}}71.14 & 44.23 & 65.25 & 63.93 & 48.55 & 58.12 \\
Movie    & 55.88 & 66.91 & 68.28 & {\cellcolor{diag}}84.92 & 66.47 & 61.50 & 48.69 & 52.98 \\
GUI      & 57.47 & 61.35 & 71.94 & 85.50 & {\cellcolor{diag}}78.52 & 64.59 & 48.36 & 55.98 \\
Science  & 57.20 & 62.58 & 65.82 & 85.36 & 77.35 & {\cellcolor{diag}}82.20 & 62.73 & 54.79 \\
Sports   & 57.31 & 61.25 & 63.68 & 85.33 & 76.13 & 81.43 & {\cellcolor{diag}}89.66 & 52.99 \\
Reason.  & 56.83 & 66.34 & 65.71 & 82.18 & 75.35 & 79.93 & 89.13 & {\cellcolor{diag}}86.75 & {\cellcolor{last}}75.28 \\
\midrule
Average & 57.09 & 62.18 & 63.69 & 67.00 & 70.90 & 68.95 & 61.46 & 59.82 &{\cellcolor{average}}63.88 \\

\bottomrule
\end{tabular}
\end{minipage}

\vspace{-0.1in}
\end{table*}

\vfill
\end{document}